%% file: main.tex
\documentclass{article}

\input{template/package}
\input{template/macro}

\usepackage[accepted]{template/icml2023}

\definecolor{pltgray}{RGB}{200, 200, 200}
\definecolor{pltred}{RGB}{255, 93, 28}
\definecolor{pltblue}{RGB}{84, 140, 188}

\icmltitlerunning{The SSL Interplay}

\begin{document}

\twocolumn[
\icmltitle{The SSL Interplay: Augmentations, Inductive Bias, and Generalization}

\begin{icmlauthorlist}
\icmlauthor{Vivien Cabannes}{meta}
\icmlauthor{Bobak T. Kiani}{mit}
\icmlauthor{Randall Balestriero}{meta}
\icmlauthor{Yann LeCun}{meta}
\icmlauthor{Alberto Bietti}{meta}
\end{icmlauthorlist}
\icmlaffiliation{mit}{MIT Department of Electrical Engineering and Computer Science, Cambridge, MA, USA}
\icmlaffiliation{meta}{Meta AI, New York, NY, USA}
\icmlcorrespondingauthor{Vivien Cabannes}{vivc@meta.com}
\icmlkeywords{Self-supervised learning, convergence rates, statistical learning theory}

\vskip 0.3in
]

\printAffiliationsAndNotice{} 

\begin{abstract}
\input{abstract}
\end{abstract}

\input{core}

\bibliography{main}

\bibliographystyle{template/icml2023}

\clearpage
\onecolumn
\appendix
\input{appendix/proof}

\input{appendix/downstream}

\input{appendix/upstream}

\input{appendix/examples}

\input{appendix/experiments}

\end{document}

%% file: template/package.tex
\usepackage[T1]{fontenc}
\usepackage[utf8]{inputenc}
\usepackage[english]{babel}

\usepackage{times}

\usepackage{mathtools} 
\usepackage{amsthm}
\usepackage{amsmath}

\usepackage{amsfonts,amssymb,stmaryrd}

\PassOptionsToPackage{hyphens}{url}
\usepackage[hyphens]{url}
\usepackage[hidelinks]{hyperref}

\usepackage{graphicx}
\usepackage{caption}
\usepackage{tikz}
\usetikzlibrary{hobby,decorations.markings}
\usepackage{pgfplots}
\pgfplotsset{compat=1.17}

\usepackage{array}
\usepackage{booktabs}
\usepackage{microtype}
\usepackage{xcolor}
\usepackage{enumitem}
\setitemize{itemsep=0pt,parsep=0pt,topsep=0pt}
\setenumerate{itemsep=0pt,parsep=0pt,topsep=0pt}

\usepackage[sectionbib,authoryear,round]{natbib}

\usepackage[normalem]{ulem}

%% file: template/macro.tex
\newcommand{\card}[1]{\left\vert #1 \right\vert}
\newcommand{\diff}{\mathop{}\!\mathrm{d}}
\newcommand{\ind}[1]{\mathbf{1}_{#1}}
\newcommand{\prob}[1]{\Delta_{#1}}
\DeclareMathOperator{\E}{\mathbb{E}}
\DeclareMathOperator{\Pbb}{\mathbb{P}}
\DeclareMathOperator*{\argmax}{arg\,max}
\DeclareMathOperator*{\argmin}{arg\,min}

\newcommand{\uniform}[1]{{\cal U}\paren{#1}}

\DeclareMathOperator{\Span}{Span}

\DeclareMathOperator{\diag}{diag}
\DeclareMathOperator{\diam}{diam}

\DeclareMathOperator{\ima}{im}
\DeclareMathOperator{\orb}{orb}
\DeclareMathOperator{\op}{op}

\DeclareMathOperator{\spec}{spec}
\DeclareMathOperator*{\supp}{supp}
\DeclareMathOperator{\trace}{Tr}

\newcommand{\C}{\mathbb{C}}
\newcommand{\N}{\mathbb{N}}
\newcommand{\Q}{\mathbb{Q}}
\newcommand{\R}{\mathbb{R}}
\newcommand{\Z}{\mathbb{Z}}

\renewcommand{\brace}[1]{\left\{ #1 \right\}}
\newcommand{\bracket}[1]{\left[ #1 \right]}

\newcommand{\paren}[1]{\left( #1 \right)}
\newcommand{\midvert}{\,\middle\vert\,}

\newcommand{\abs}[1]{\left| #1 \right|}

\newcommand{\norm}[1]{\left\| #1 \right\|}
\newcommand{\scap}[2]{\left\langle #1, #2 \right\rangle}

\newcommand{\Sfrak}{\mathfrak{S}}
\newcommand{\X}{\mathcal{X}}
\newcommand{\Y}{\mathcal{Y}}

\renewcommand{\epsilon}{\varepsilon}
\renewcommand{\phi}{\varphi}

\theoremstyle{plain}
\newtheorem{theorem}{Theorem}

\newtheorem{lemma}{Lemma}
\newtheorem{proposition}[lemma]{Proposition}
\newtheorem{definition}[lemma]{Definition}
\newtheorem{assumption}{Assumption}
\newtheorem{remark}[lemma]{Remark}
\newtheorem{example}{Example}



%% file: abstract.tex
Self-supervised learning (SSL) has emerged as a powerful framework to learn representations from raw data without supervision.
Yet in practice, engineers face issues such as instability in tuning optimizers and collapse of representations during training. 
Such challenges motivate the need for a theory to shed light on the complex interplay between the choice of data augmentation, network architecture, and training algorithm. 
We study such an interplay with a precise analysis of generalization performance on both pretraining and downstream tasks in a theory friendly setup, and highlight several insights for SSL practitioners that arise from our theory.

%% file: core.tex
\section{Introduction}

\begin{figure}[t]
    \centering
    \includegraphics[width=0.15\textwidth]{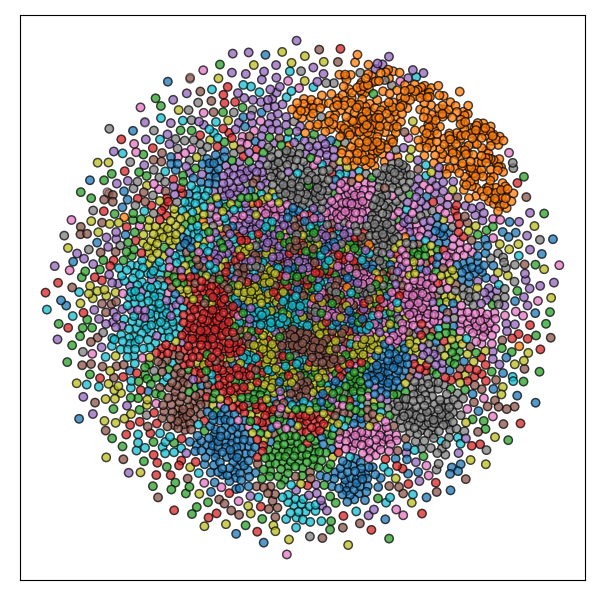}
    \includegraphics[width=0.15\textwidth]{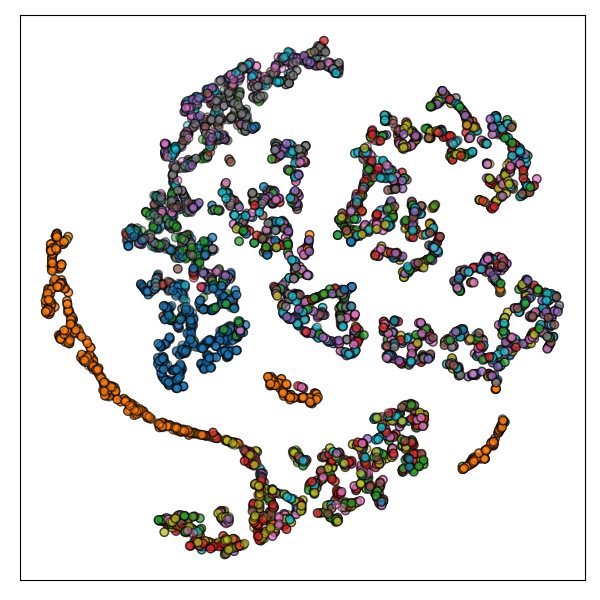}
	\includegraphics[width=0.15\textwidth]{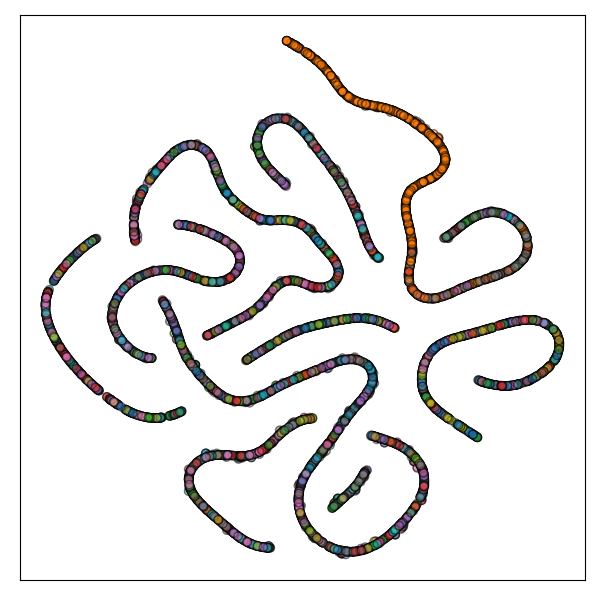}
	\vspace{-1em}
    \caption{\emph{Effect of augmentations and architecture.}
	  TSNE of representations learned on MNIST with no augmentations (left) or with rotations and an MLP (middle) or a CNN (right).
	  The representations depend on both the augmentations and the architecture.}
	\vspace{-1em}
    \label{fig:tsne}
\end{figure}

Self-supervised learning (SSL) aims to construct useful representations of data without the need for pre-constructed labels.
Due to the recent success and widespread applicability of SSL, established methods for training large neural networks now incorporate pre-training of models in an unsupervised manner over large amounts of data, before fine-tuning/probing them over downstream datasets~\citep{devlin2019bert,chen_simple_2020,brown2020language,radford2021learning}.
Self-supervised pretraining generally aims to render the model invariant to certain distorsions/views of the inputs, in order to capture useful features for downstream tasks~\citep[e.g.,][]{chen_simple_2020,caron2020unsupervised,grill2020bootstrap,caron2021emerging,bardes_vicreg_2022}.
Though very powerful, SSL methods can be challenging to implement properly.
They tend to suffer from various practical issues, such as instability and collapse during training and the need to carefully tune parameters related to the architecture, optimization algorithm, representation dimension, and form of augmentations.
These different aspects of pretraining can lead to widely different behaviors and representations, as illustrated for instance in Figure~\ref{fig:tsne}.
These challenges motivate new theoretical insights to better understand why such issues arise and how to better address them.

Our study focuses on the joint-embedding framework and characterizes learned representations for given choices of input distributions, data augmentations, and architecture.
To obtain a fine-grained picture, we study linear classes of functions endowed with a reproducing kernel, and analyze a theoretically friendly loss function that models both contrastive and non-contrastive methods.
Our work generalizes the discrete data setting of~\citet{haochen_provable_2021} and the finite dimensional setting of~\citet{saunshi_understanding_2022}, encompassing more expressive nonparametric models, potentially with universal approximation properties, and which can capture certain properties of architectures through their limiting kernel limits~\citep{jacot_ntk_2018}.

Our {\em contributions} are as follows:
\begin{enumerate}
  \item We unveil two central integral operators: an ``intrinsic'' one that depends on the input distribution and choice of augmentations and another capturing the inductive bias associated with the model of~computation.
  \item We provide new bounds on the downstream generalization error that are sharper than previous work, and which can handle distributions shift between data before and after performing augmentations. 
  \item We propose new generalization bounds on the pretraining excess risk via tools from convex analysis. This analysis yields novel insights, including an understanding of the benefits of using multiple augmentations per sample (e.g., ``multi-crop'').
  \item We detail several examples where optimal representations are found in closed form, illustrating the role of augmentations, architecture, and regularization in forming representations. 
  \item We discuss several practical insights for SSL practitioners that emerge from our theory, in particular on how design choices in pretraining may affect downstream performance, and on how to avoid collapse of representations.
\end{enumerate}

\paragraph{Related work.}
Foundations for theoretically analyzing SSL have emerged in the past few years.
Particularly relevant to our work, \citet{balestriero_contrastive_2022,kiani2022joint} provide theoretically friendly characterizations of many self-supervised learning settings, including closed-form solutions of representations in the kernel setting.
For contrastive learning, SSL was first theoretically analyzed by \citet{arora_theoretical_2019,tosh2021contrastive,tosh_contrastive_2021,tian_understanding_2021}.
Notably, \citet{haochen_provable_2021} recently leveraged tools in spectral graph theory to characterize guarantees on SSL performance under clustering assumptions.
These assumptions were deemed impractical by~\citet{saunshi_understanding_2022}, who highlighted the importance of incorporating inductive bias to obtain provable guarantees.
This line of work was extended to multi-modal SSL by \citet{Lee2021} where in essence the central symmetric operator $T$ is replaced by a non-symmetric one, and the eigen decomposition is replaced by the singular one.
The role of inductive bias has also been scrutinized through analysis of feature learning in training dynamics by \citet{wen2021toward} and \citet{tian2022understanding}.

\section{Setup }
\label{sec:setup}

Machine learning streamlines the task of creating algorithms for finding patterns in data.
An algorithm is conceptualized as a mapping $f$ from an input $x\in\X$ to an output $y\in\Y$.
To construct this mapping $f:\X\to\Y$, one can choose a measure of disagreement $\ell:\Y\times\Y\to\R$, and minimize the~risk
\begin{equation}
    \label{eq:down_obj}
    {\cal R}(f) = \E_{(X, Y)\sim\rho}[\ell(f(X), Y)],
\end{equation} 
for $\rho\in\prob{\X\times\Y}$ a distribution on I/O pairs.
We denote by $f^*\in\argmin{\cal R}$ an optimal I/O map according to the risk.
Mapping raw inputs (e.g., arrays of pixels), to outputs (e.g., classifying an animal in an image), is in general a challenging task.
An effective technique consists of first extracting (or engineering) meaningful features $\psi:\X\to\R^k$ from input data before using those features to search $f$ under the form $g\circ\psi$ for $g:\R^k\to\Y$ a simple function.\footnote{For convenience, several technicalities, such as measurability, have been deferred to the appendix.}

Though features $\psi$ can be hand-engineered, representation learning aims at improving such design via unsupervised learning procedures. 
On the one hand, {\em reconstruction based methods} mask or add noise to inputs via a mapping $Mx$ and aim to reconstruct the original input $x$ from the features $g\circ \psi$ using $g$, a simple prediction head.
Large language models largely rely on this paradigm, usually learning $\psi$ by completing sentences $Mx$ where word tokens are masked \citep[e.g.][]{devlin2019bert}.
On the other hand, {\em joint embedding methods} learn $\psi$ by leveraging invariance to small perturbations of the semantic information contained in inputs.
This is the paradigm we shall focus on.
Recently, joint embedding methods have relied heavily on the concept of data augmentation, such as small rotation, translation, color jittering of images.
In particular, contrastive methods learn $\psi$ by enforcing that if two augmentations $\xi$ and $\xi'$ come from the same data point, their representation $\psi(\xi)$ and $\psi(\xi')$ are close; while if they come from different data points, their representation are far away from one another \citep[e.g.,][]{chen_simple_2020}.
Non-contrastive methods only enforce similarities of augmented datapoints and avoid collapse by enforcing richness of the representation \citep[see, e.g.,][]{bardes_vicreg_2022}.
In the following, we focus on a theoretically friendly variant of VICReg \citep{balestriero_contrastive_2022} with parameter $\beta > 0$, defined for $\psi:\X\to\R^k$ by
\begin{align}
    {\cal L}(\psi) 
    &= \beta\E_X\E_{\xi, \xi'} \bracket{\norm{\psi(\xi) - \psi(\xi')}^2\midvert X} 
    \nonumber
    \\&\qquad+ \norm{\E_{\xi}[\psi(\xi)\psi(\xi)^\top] - I}^2_2,
    \label{eq:vic}
\end{align}
where pairs of inputs/augmentations $(X, \xi)$ follow a distribution $\mu\in\prob{\X\times\X}$, whose conditional $\paren{\xi\midvert X}$ arises from the choice of augmentation.
The first term in ${\cal L}$ enforces invariance of the representation $\psi$ to two augmentations $\xi$ and $\xi'$ of the same input $X$, while the second term lowers risk of collapse by pushing coordinates $\psi_i:\X\to\R$ of $\psi = (\psi_i)_{i\in[k]}$ to be orthogonal in $L^2$.

\begin{table}[t]
    \centering
	\begin{tabular}{|c|c|c|}
    \hline
        Practice & Theory & Quantity \\
    \hline
         Augmentation & Spectral embedding & $T$ \\
         Architecture & Space of functions & $K$ \\
         Optimization & Regularization & $\lambda$ \\
        \hline
         \multicolumn{2}{|c|}{Subtle Interplay} & $T_\lambda$ \\
    \hline
    \end{tabular}
	\caption{\emph{Analogy between practice and theory} that this paper proposes to help disentangle the various phenomena of SSL training.}
 \vspace{-1em}
    \label{tab:analogy}
\end{table}

\begin{remark}[Contrastive learning with ${\cal L}$]
    \label{lem:same_loss}
    When $\beta=1$, the population loss ${\cal L}$ is equivalent to the spectral contrastive loss studied in \citet{haochen_provable_2021} as a theoretically friendly proxy for SimCLR \citep{chen_simple_2020}.
    In other terms, ${\cal L}$ analyzes both contrastive and non-contrastive approaches to representation learning.
\end{remark}

Given a representation $\psi$, one can optimize for $f$ through linear probing by constructing $f = g\circ\psi$ where $g$ is a linear function.
$f$ is thereby in the class of functions
\begin{equation}
    \label{eq:lin_class}
    {\cal F} = \brace{x\mapsto w^\top \psi(x)\midvert w\in\R^k}.
\end{equation}
In practice, one might not know the optimal $\psi$, but can estimate it as $\hat\psi$ from empirical data, leading to an estimate~$\hat{\cal F}$ of this class of functions.

\section{Representation learning }
In this section, we study the representations induced by pretraining with specific augmentations and inductive biases.

\subsection{Closed form solution}
Equation \eqref{eq:vic} admits a closed form solution for $\psi$ upon noting that the invariant part is a quadratic form.

\begin{lemma}[Spectral embedding]
    \label{lem:close}
    There exists a linear positive symmetric operator $L$ in $L^2$ for which the operator $I-T$ is positive and
    \[
        \E_X\E_{\xi, \xi'} \bracket{\norm{\psi(\xi) - \psi(\xi')}^2\midvert X} = \sum_{i\in[k]}\psi_i^\top L\psi_i.
    \]
    To be consistent with previous literature, we will rather use $T = I - L/2$, which is also a linear positive symmetric operator, and is defined as, for $\psi_1, \psi_2 \in L^2$
    \[
        \psi_1^\top T \psi_2 = \E_X\E_{\xi, \xi'}\bracket{\psi_1(\xi)^\top \psi_2(\xi)\midvert X}
    \]
    As a consequence, if $(\lambda_i)$ are the eigenvalues of $T$ and $(f_i)$ are the corresponding eigenvectors, a minimizer of ${\cal L}$ is $\psi_i = \sqrt{\mu_i} f_i$ with $\mu_i = 1-\beta+\beta\lambda_i$.
\end{lemma}

Lemma \ref{lem:close} is closely tied to the guiding principle in unsupervised learning that a good representation of data should minimize variations over the manifold of the data \citep{cabannes_minimal_2022}, and techniques that learn such representations through spectral decomposition of a central operator \citep[see, e.g.,][]{Coifman2006}.

\subsection{Search within a linear class of functions}
\label{sec:kernel}

In this more technical section, we study solutions of~${\cal L}$ for~$\psi$ belonging to a linear class of functions. The coordinates of the mapping $\psi:\X\to\R^k$ are typically searched within a space of functions $\Psi\subset \R^\X$, leading to $\psi\in\Psi^k$.
In our theoretically friendly setup, we assume that $\Psi$ is a linear class of functions endowed with a Hilbertian topology such that the linear evaluations $\psi\mapsto\psi(x)$ are continuous for almost all $x\in\X$.
The theory of reproducing kernel Hilbert space \citep{Scholkopf2001} asserts that $\Psi$ can be parameterized by a Hilbert space ${\cal H}$ and a mapping $\phi:\X\to{\cal H}$ such that 
\begin{equation}
    \label{eq:rkhs}
    \Psi = \brace{x\mapsto f_\theta(x) \midvert f_\theta(x) = \scap{\theta}{\phi(x)}_{\cal H}, \theta\in{\cal H}}.
\end{equation}
This generalizes the setting of \citet{haochen_provable_2021} where~$\X$ is assumed to be finite and $\Psi$ is parameterized by ${\cal H}=\R^\X$ and $\phi(x) = \delta_x$, as well as the setting of \citet{saunshi_understanding_2022} where ${\cal H}$ is assumed to be finite dimensional.

To describe architectures such as neural networks with such a linear structure, it is common to linearize those models (e.g. \citet{jacot_ntk_2018}) as 
\[
    \psi_\theta(x) = \psi_{\theta_0}(x) + \scap{\nabla_{\theta_0}\psi_{\theta_0}(x)}{\theta-\theta_0} + o(\norm{\theta - \theta_0}),
\]
where $\theta$ are the network parameters, assumed close to their initialization~$\theta_0$, and $\psi_\theta$ is the neural network. In this case, we may take $\phi = \nabla_{\theta_0}\psi_{\theta_0}$, which arguably describes some regimes of wide neural networks~\citep{lee2019wide}.

To minimize ${\cal L}$ in practice and improve generalization, a regularization parameter is typically introduced.\footnote{While we study here the bias of Tikhonov regularization for simplicity, similar studies can be done for early stopped gradient descent or stochastic gradient descent when they are cast as spectral filters, as in~\citet{Lin2020}, see also literature related to optimization for matrix factorization problems \cite{chi_2019}, which has been applied to SSL by \citet{Simon2023}.}
The following lemma provides a closed form solution of the regularized variant of ${\cal L}$.

\begin{lemma}[Regularized population loss]
    \label{lem:close_rkhs}
    For $\Theta\in\R^k\otimes{\cal H}$, and a regularizer $\lambda > 0$, the regularized loss
    \(
        {\cal L}(S\Theta) + \lambda\norm{\Theta}^2_2
    \)
    can be minimized in closed form with the operator
    \begin{equation}
        T_\lambda = (1-\beta) I + \beta T - \lambda K^{-1}.
    \end{equation}
    where $K = SS^\top$ for $S:{\cal H}\to L^2(\mu_\Xi);\theta\mapsto f_\theta$ the embedding of ${\cal H}$ in $L^2$.
    Specifically, if $(\lambda_i)$ are the (decreasing) eigenvalues of $T_\lambda$ and $(f_i)$ the corresponding eigenfunctions, a minimizer is given by $\psi_i = \max\brace{\lambda_i, 0}f_i$.
\end{lemma}

\begin{figure}[t]
  \centering
  \includegraphics{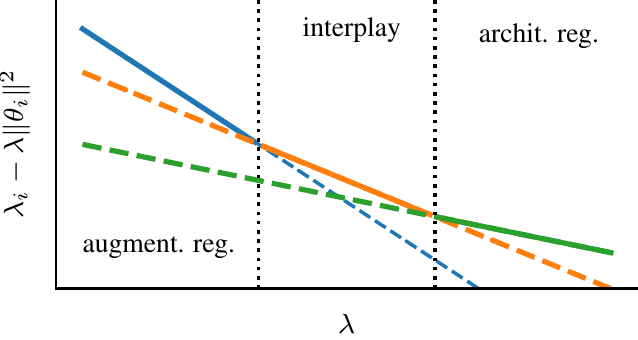}
  \vspace{-1em}
  \caption{
  \emph{Interplay between $T$ and $K$} as a function of $\lambda$.
  Illustration of Proposition \ref{prop:commute} in a setting where $(\lambda_i) = (.9, .75, .5)$ and $(\norm{\theta_i}^2) = (.4, .25, .125)$.
  The plot displays the eigenvalues associated with three different eigenfunctions as a function of $\lambda$, $\beta$ is set to one for convenience.
  When $\lambda=0$, the minimizer $\psi_*:\X\to\R$ of \eqref{eq:vic} is defined through $T$, here $\phi_*=f_1$ ($i=1$, shown in blue), when $\lambda$ is big $\psi_* = f_3$ (green) mainly depends on $K$. In the middle, there is an interplay between these two regimes leading to $\psi_* = f_2$ (orange).
  The three regimes are named the ``augmentation'', the ``architecture'' (or VCReg) and the ``interplay'' regime respectively.
  This abstract setting can be instantiated with a two-layer ReLU network and cropping as detailed in Figure~\ref{fig:interplay_crop}.}
  \label{fig:interplay}
\end{figure}

\subsection{The need for inductive bias}
Two different points of view motivate the introduction of the regularizer $\lambda\norm{\Theta}$ leading to the operator $T_\lambda$.

In the classical viewpoint of statistical learning theory, one would like to retrieve the eigenfunctions of $T$ to minimize ${\cal L}$ (Lemma \ref{lem:close}).
However, when solely accessing finitely many samples of data, eigenfunctions of $T$ should be searched within a space of finite capacity (i.e. $\{f\in\Psi \,|\, \|f\|_{\Psi}^2\leq\lambda^{-1}\}$).
Though fewer samples are needed for smaller models (e.g. the fewer neurons and layers in a deep network), such small models are unlikely to be expressive enough to represent the ideal solutions.
This echoes the classical trade-off between approximation and estimation error.
In the case of Laplacians, one can assume that the eigenfunctions of~$T$ are smooth thereby belong to a small space of functions that are well-approximated with a finite model of computation.
We refer the curious reader to \citet{cabannes_overcoming_2021} for results in this vein when $I-T$ is the real Laplacian in $L^2$. 

\begin{figure}[t]
    \centering
    \includegraphics{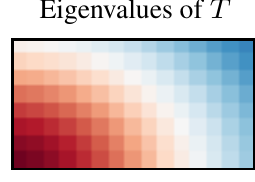}
    \includegraphics{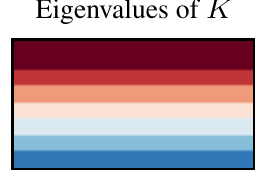}
    \includegraphics{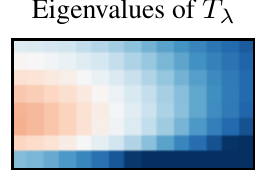}

	  \vspace{-1em}
    \caption{
    \emph{Trade-off on eigenvalues between~$T$ and~$K$.}
	Illustration of a harmonic setting where $T$ and $K$ are diagonalized in the same basis.
	This basis is parametrized by an ``invariance score'' ($x=m$ in \eqref{eq:spherical}) and a ``complexity score'' ($y=\card{S}$ in \eqref{eq:spherical}).
	The eigenvalues $\lambda_{x, y}(A)$ for $A\in\brace{T, K, T_\lambda}$ are represented with colors and displayed in a grid associated with $x\in[15]$ and $y\in[8]$.
	The sole use of the operator $T$ biases towards invariance (lower $x$) with high complexity (lower $y$), while the sole use of $K$ biases toward low complexity.
	The interplay between the two results in $T_\lambda$ whose biggest eigenfunctions have high invariance and low complexity, and corresponds to an ideal representation $\psi$. 
    }
    \label{fig:degree_vs_invariance}
\end{figure}

Another take was suggested by \citet{saunshi_understanding_2022}, which pointed out that eigenvalues of $T$ can have large multiplicity in realistic situations (in particular in the non-localized augmentations setting of Section \ref{sec:harmonics}), meaning that the space ${\cal F}$ is not uniquely defined from the loss ${\cal L}$.
As a consequence, defining the optimal solution solely from $T$ is somewhat ill-posed, whereas, when $K$ is properly chosen, $T_\lambda$ could define a ``more principled'' representation $\psi$.
Paradoxically, with this viewpoint, {\em bias could reduce the approximation error}.
Figure~\ref{fig:degree_vs_invariance} illustrates such an idea.
It leverages the following interpretation of the inductive bias in the friendly setting where~$T$ and~$K$~commute.

\begin{proposition}
    \label{prop:commute}
    If $T$ and $K$ commute, and if $(\lambda_i)$ are the eigenvalues of $T$ and $(f_i)$ its eigenfunctions, then there exists $(\theta_i)$ such that $f_i = f_{\theta_i}$ \eqref{eq:rkhs}. 
    Moreover, the optimal representation to minimize the regularized loss are the $f_i$ that maximize
    \(
        \beta\lambda_i - \lambda\norm{\theta_i}^2.
    \)
    In other terms, the regularization biases towards representations that have a small complexity with respect to the model of computation.
\end{proposition}

Lemma \ref{lem:close_rkhs} shows an interesting behavior of the VCReg loss ($\beta=0$, i.e. VICReg without invariance term).
In this setting, the optimal $\psi$ retrieve the largest eigenfunctions of $K$, recovering kernel PCA.
Learning downstream tasks with linear probing of the resulting $\psi$ is equivalent to linear regression with an eigenvalue cut-off, which is a powerful spectral filtering technique \citep[see, e.g.][]{Bun2017}.

\section{Illustrative examples}
\label{sec:example}

The analysis of this paper relies on two central operators: $T$, that is ``intrinsically'' defined from the data distribution and augmentations, and $K$, which relates to the model of computation (e.g. the network architecture).
Once those operators are chosen, Section \ref{sec:conv} provides a sharp analysis of convergence and generalization with SSL in the kernel regime.
In essence, Assumption \ref{ass:source} requires that the target function (downstream) align well with the learned representation (upstream) when given infinite data.
Here, the effect of $T$ and the inductive bias introduced by $\lambda K^{-1}$ on the learned representation can appear abstract. 
To provide intuition and outline important properties of these operators, this section lays out several concrete examples to help practitioners better understand the role of augmentations and their interplay with the inductive bias of the architecture.

Two different perspective have emerged to understand learned representation in SSL.
One intuition comes from the spectral clustering literature, and is the object of subsection \ref{sec:lap}.
The other intuitive way to understand SSL is based on harmonic analysis, and is the object of subsection \ref{sec:harmonics}. 
All in all, this section generalizes previous works by dropping out strong clustering assumptions in the data, showing that what really matter are the eigenfuntions of $T$, which eventually capture clustering structures when such clustering assumptions are invoked. 
It further uses harmonic analysis tools to better describe these eigenfunctions as suggested by \citet{saunshi_understanding_2022} and detailed in Table~\ref{tab:augmentations_hypercube}.

\subsection{Low-variation with localized augmentations}
\label{sec:lap}

When augmentations $\paren{\xi\midvert X}$ are localized around the input~$X$, optimizing the loss ${\cal L}$ \eqref{eq:vic} biases towards small gradients of $\psi$ along the directions of augmentations.
Formally for $\psi:\X\to\R$, using first order Taylor expansion,
\[
	\E\bracket{\norm{\psi(\xi) - \psi(\xi')}^2\midvert X} 
  	\simeq \E\bracket{\scap{\nabla\psi(X)}{\xi - \xi'}^2 \midvert X}.
\]
Under isotropic augmentations, the objective simplifies as
\[
	\E\bracket{\scap{\nabla\psi(X)}{\xi - \xi'}^2 \midvert X} \propto \norm{\nabla\psi(X)}^2,
\]
which enforces $\psi$ to have small variations on densely populated regions of the input space -- reminiscent of popular approaches to tackle representation and semi-supervised learning in the last two decades \citep{engelen_surver_2020}.
More generally, augmentations govern the important directions of invariance for $\psi$, recovering a finite-differences approach to the Tangent Prop algorithm \citep{Simard1991}.

Low variation methods are particularly useful when data display a clustering structure (c.f. Figure \ref{fig:var} for an illustration with neural networks).
If augmentations preserve the clustering structure, ${\cal L}$ is minimized by piecewise constant functions on each cluster, leading to useful features for downstream tasks that involve classifying different clusters \citep{haochen_provable_2021,Schiebinger2015}.
The inductive biases further deforms those top eigenfunctions to be regular in a sense defined by~$\Psi$ \eqref{eq:rkhs}, e.g., analytic if we use a radial basis function kernel \citep{Sun2008}.

\subsection{The role of augmentations}
\label{sec:harmonics}

\begin{table*}[t!]
    \small
    \centering
    \begin{tabular}{lcp{.1em}l}
	  \multicolumn{2}{c}{Augmentation example} && \multicolumn{1}{c}{Effect of the operator  $T$} \\ 
        \hline
	  Input (no augmentation) & \input{drawings/original} && \multicolumn{1}{c}{-} \\ 
         Random noise & \input{drawings/flip} && Attenuate higher order Fourier modes \\ 
Cropping & \input{drawings/crop} && Keep Fourier modes within cropping windows \\ 
Translations  & \input{drawings/translation} && Bias towards Fourier modes with cyclic invariance \\ 
Flipping & \input{drawings/index} && Equate eigenvectors of subsets related by flips \\ 
\hline
\multicolumn{4}{l}{ \textbf{Legend: } \input{drawings/legend} }
    \end{tabular}
    \caption{\emph{Effect of common augmentations} on the optimal representation $\psi$ through the operator $T$.
	Without augmentations, $\psi$ could match any Fourier basis function.
	Augmentations filter out some of those by attenuating their eigenvalues in~$T$, and the architecture will push $\psi$ to pick some specific frequencies among the remaining ones through the operator $K$.
        The table stylizes the effect of usual augmentations on parity functions over bit streams.
	We refer the reader to Appendix \ref{app:examples} for further details and derivations. 
  }
    \label{tab:augmentations_hypercube}
    \vspace{-1em}
\end{table*}

When augmentations are not localized, which is often the case in practice, harmonic analysis provides useful tools to study in-depth the role of augmentations, in particular when data are uniform on the sphere or the hypercube.
Our findings on the hypercube are summarized in Table \ref{tab:augmentations_hypercube}.
In such a setting, we show that common augmentations enforce smoothness, locality, or invariance to certain symmetries.
For example, crops push $\psi$ to focus on details that can appear within the crop size, filtering out long-range interactions between parts of the input that are likely to be spurious features. The following example formalizes this.

\begin{example}[Cropping] \label{ex:2d_cropping}
	Consider the hypercube setting where $\X = \brace{-1, 1}^n$ and $X$ is uniformly distributed.
	A basis of $L^2(\X,\R)$ is given by the parity functions $\chi_S:x\mapsto\prod_{i \in S} x_i$ for all the subsets $S\subseteq [n]$.
	Pre-training via cropping with window sizes $v \times w$ set $T\chi_S = 0$ for all $S$ whose support forms a window larger than the size $v \times w$.
    For all the other $S$, $T\chi_S = \lambda_S \chi_S$, where $\lambda_S$ decreases with the diameter of $S$.
    In other terms, pre-training with 2-D cropping eliminates the influence of functions which act globally outside of the cropping window. 
    This, in effect, imparts a locality to the induced representation $\psi$ which is often desirable for generalization.
\end{example}

This example suggests that the ideal crop size should match the desired scale of details for~$\psi$; e.g., on a dataset with fine-grained details such as iNaturalist, one should reduce the crop window size in comparison to a dataset such as ImageNet.
Appendix~\ref{app:examples} discusses further examples of augmentations, such as random noise or translations, and shows how they bias towards smooth or invariant~eigenfunctions.

\subsection{Interplay with the architecture}
\begin{figure}[t!]
  \centering
  \vspace{-.5em}
  \includegraphics{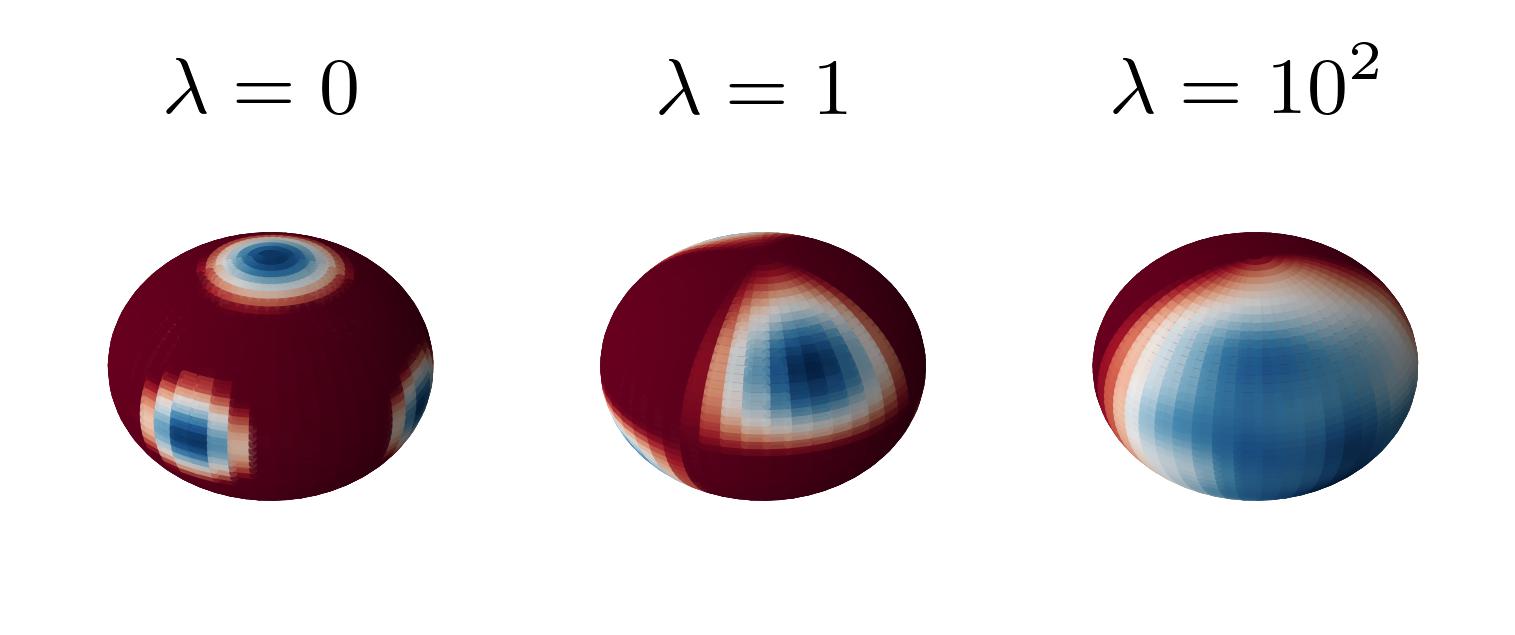}
  \vspace{-2em}
  \caption{
  \emph{Interplay on the sphere.}
  Level lines of the $7$-th eigenfunction of $T_\lambda$ for three different $\lambda$.
  Augmentations consist of translations of the $x,y,z$ coordinates together with Gaussian perturbations. 
  $K$ is the integral operator associated with the radial basis function. 
  Without regularization (left), the eigenfunction is highly localized at clusters corresponding to the action of the augmentations. 
  Increasing the regularization biases towards smoother harmonic eigenfunctions of $K$ (middle and right). }
  \label{fig:sphere_min}
  \vspace{-1em}
\end{figure}

While the design of augmentations and architecture can be done separately, changes to the architecture and optimization scheme play an important role in the resulting optimal~$\psi$. Generally, increasing the amount of inductive bias by increasing $\lambda$ shifts~$\psi$ towards smoother functions, in the sense captured by the $\cal H$ norm, which we illustrate in Figure \ref{fig:sphere_min}. 
In practice, the right amount (captured here by the parameter~$\lambda$) of inductive bias to enforce is often set by a mix of intuition, common knowledge and empirical evidence. 
For example, \citet{caron2021emerging} links the inductive bias of early stopping to beneficial outcomes noting that ``training longer~[...] has been leading to worse performance''.

\begin{example}[Dot-product kernel]
    On the Boolean hypercube setting of Example \ref{ex:2d_cropping}, many linear models \eqref{eq:rkhs} take the form $\phi(x)^\top \phi(y) = h(\scap{x}{y})$ (e.g., the classical NTK linearization of fully connected layer) leading to an integral operator~$K$ that is diagonalizable by parity functions. 
    More precisely, there exists $(\nu_i)\in\R^d$ such that $K\chi_S =\nu_{\card{S}}\chi_S$, where $\card{S}$ is the cardinality of $S$ and~$\nu_{\card{S}}$ decreases with~$\card{S}$.
    In the setting of crops, $T$ pushes towards representation on parity functions with small diameter ($\psi = (\chi_S)_S$ for $S$ with small diameters), while the inductive bias acts on the cardinality of the sets $S$, pushing towards the $\chi_S$ that maximize~$\nu_{\card{S}}$.
    Formal derivations are provided in Appendix~\ref{app:examples}.
\end{example}

\begin{figure}[t]
  \centering
  \includegraphics{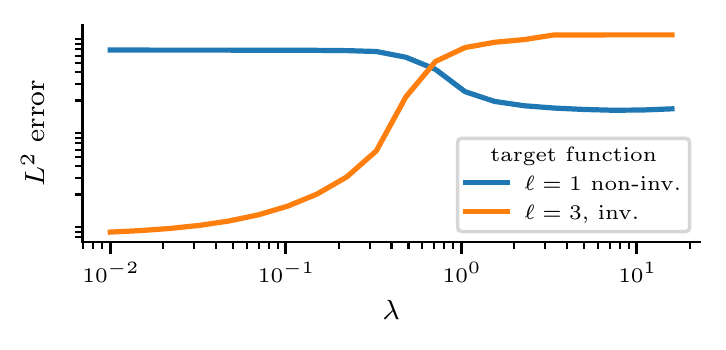}
  \vspace{-1em}
  \caption{
  \emph{Trade-off on downstream errors.}
  Effect of pretraining regularization~$\lambda$ on the empirical downstream error for two tasks on the sphere $\mathbb S^7$. 
  The targets~$f_\ell^*$ are polynomials of degree~$\ell\in \{1, 3\}$, with only~$f_3^*$ invariant to translations.
  $K$ is built from a dot-product kernel that acts as a regularizer on degrees, while $T$ is built from local translations.
  Designing $\psi$ from $T$ alone ($\lambda = 0$) is helpful to learn globally invariant polynomials in the downstream, while increasing the regularization $\lambda$ helps to learn polynomials of small degree. Experiment details in Appendix~\ref{app:downstream_error_sphere}, and Figure \ref{fig:nn_sphere} showcases a similar behavior for neural networks.
  }
  \label{fig:downstream_error}
  \vspace{-1em}
\end{figure}

Appendix~\ref{app:examples} provides additional examples. For instance, in the case of translations, there is a similar interplay between a low-degree bias in~$K$ versus~an invariance bias in~$T$. We also consider convolutional architectures, which can impose locality through constraints on~$\diam(S)$, on top of a low-degree bias.
Figure~\ref{fig:degree_vs_invariance} shows such trade-offs in eigenvalues, Figure~\ref{fig:sphere_min} visualizes how this interplay may affect leading eigenfunctions in a spherical setup, and Figure~\ref{fig:downstream_error} illustrates the resulting effect on different downstream tasks.

\section{Convergence analysis }
\label{sec:conv}

The following section analyzes guarantees on both the pretraining and downstream tasks.\footnote{The pretraining and downstream tasks refer to minimization of ${\cal L}$ and ${\cal R}$ respectively.}
For simplicity, we consider the mean-square loss $\ell(y, y') = \norm{y-y'}^2$ with $\Y=\R^{d_y}$.
The studies of many losses can be reduced to the least-squares case thanks to calibration inequalities \citep{bartlett_convexity_2006} or self-concordance properties \citep{ostrovskii_finite_2018}.
To precisely study convergence rates, we consider the kernel regime of Section \ref{sec:kernel}, where~${\cal F}$ is specified through~$\Theta_*$ of Lemma~\ref{lem:close_rkhs} as
\begin{equation}
    \tag{\ref{eq:lin_class}}
    {\cal F} = \brace{x\to w^\top \Theta_* \phi(x) \midvert w\in\R^k},
\end{equation}
and $\hat{\cal F}$ is defined similarly with $\hat\Theta$ as an estimate of $\Theta_*$.
In the following, $(f_i)$ denote the eigenfunctions of $T_\lambda$ ordered by decreasing eigenvalues, and~$\lambda$ is considered to be fixed throughout this section.

\subsection{Dealing with distribution shift}

Self-supervised learning algorithms often incorporate strong augmentations leading to potentially different marginal distributions over inputs and augmentations.
This discrepancy is often overlooked, many theoretical works implicitly assuming $\rho_\X = \mu_\Xi$.
In practice, the marginal distribution~$\rho_\X$ of inputs in the downstream task can be meaningfully different from the marginal distribution of augmentations~$\mu_\Xi$ on which we have imposed orthogonality of the representation~$\psi$ in the pretraining task.
However, the optimal representation $\psi$ is likely to be invariant to augmentations, meaning that ideally, $\psi(X)$ should have the same distribution when $X\sim\mu_\Xi$ or $X\sim\rho_\X$, which we write formally as $\psi_\#\mu_\Xi = \psi_\#\rho_\X$. 
Moreover, augmentations are likely to spread the input data distribution, leading to the dominance $\rho_\X \ll \mu_\Xi$.
This motivates the following assumptions and~definitions.

\begin{assumption}[Low expansion]
    \label{ass:interpolation_simple}
    There exists $c_r > 0$ such that for any function~$f$ in the original space of functions $\Psi$ defined in~\eqref{eq:rkhs},
    \[
        \norm{f}^2_{L^2(\rho_\X)} \leq c_r\norm{f}^2_{L^2(\mu_\X)},
    \]
\end{assumption}

\begin{assumption}
    \label{ass:robust_simple}
    For any $i$ smaller than the number of positive eigenvalues of $T_\lambda$, the projection of the target $f^*$ on $f_i$ in $L^2(\mu_\Xi)$ coincides with the projection on $f_i$ in $L^2(\rho_\X)$. 
\end{assumption}

To make those two concepts more concrete, we provide three examples below. 

\begin{example}
    \label{ex:interpolation}
    If $\rho_\X$ has a density against $\mu_\Xi$ which is bounded from below by $\delta \in (0, 1]$ on its support, i.e. $\mu_\Xi = \delta \rho_\X + (1-\delta)\mu_\perp$ with $\mu_\perp \in\prob{\X}$, then Assumption~\ref{ass:interpolation_simple} is met with $c_r = 1 / \delta$.
\end{example}

\begin{example}
\label{ex:inter_cov}
    Let $\Sigma_\tau = \E_{X\sim\tau}[\phi(X)\phi(X)^\top]$ be the covariance matrix of $\phi$ under the distribution~$\tau$.
    When there exists $c$ such that $\Sigma_{\rho_\X} \preceq c\Sigma_{\mu_\Xi}$ (i.e $c \Sigma_{\mu_\Xi} - \Sigma_{\rho_\X}$ is positive semi-definite), then Assumption \ref{ass:interpolation_simple} holds with $c_r=c$.
\end{example}

\begin{example}
    \label{ex:invariant}
    If $\psi_\sharp\mu_\Xi = \psi_\sharp\rho_\X$ holds for the optimal representation $\psi = (f_i)$, with $(f_i)$ the positive eigenfunctions of $T_\lambda$, and there exists a measurable function $g:\R^k\to\Y$ such that $f^* = g \circ \psi$, then Assumption \ref{ass:robust_simple} is verified.
\end{example}

In essence, Assumptions \ref{ass:interpolation_simple} and \ref{ass:robust_simple} allow for the incorporation of augmented data that does not resemble the original data as long as the model of computation (Assumption \ref{ass:interpolation_simple}) and training via the VICReg loss (Assumption \ref{ass:robust_simple}) do not bias too much towards this aberrant augmented data.
Example \ref{ex:interpolation} states that when the augmented data mostly looks like the original samples then one does not have to worry about bias introduced by the model of computation.
Example \ref{ex:inter_cov} gives a more relaxed guarantee based on second order moments.
Finally, Example \ref{ex:invariant} states that one need not worry about the idiosyncrasies of the augmented data if the learned representations confound augmented data with their original samples.

\subsection{Generalization on downstream tasks}

This section provides comprehensive generalization guarantees on supervised downstream tasks.
The following assumption ensures that the target function~$f^*(x) = \E_\rho[Y | X=x]$ of the downstream task is well represented by the pretraining problem.

\begin{assumption}[Source condition]
    \label{ass:source}
    $f^*$ belongs to the positive eigenspace of $T_\lambda$, i.e. $f^*\in \Span\brace{f_i\midvert \lambda_i > 0}$.
\end{assumption}
\begin{example}[Cluster assumption]
    \label{ex:cluster}
    If the support of the density $\mu_\Xi$ has $k$ connected components, $f^*$ is constant on those clusters, and $\lambda=0$, then Assumption \ref{ass:source} holds. 
\end{example}

We now give a simplified version of our downstream guarantee. See Theorem~\ref{thm:conv_master} in Appendix~\ref{app:theorem} for the full statement.

\begin{theorem}[Downstream error]
    \label{thm:conv_simple}
    Let $(X_i, Y_i)\sim\rho^{\otimes n}$ be $n$ samples drawn from the distribution for the downstream task and $\ell$ be the square loss.
    Define $k_\lambda < +\infty$ as the number of strictly positive eigenvalues of $T_\lambda$.
    Under Assumptions~\ref{ass:interpolation_simple},~\ref{ass:robust_simple}, and~\ref{ass:source}, after a transitory regime, the average excess risk of the optimally-regularized empirical risk minimizer $f_n$ is
    \begin{align}
        \nonumber
        &\E[{\cal R}(f_n) - {\cal R}(f^*)]
		\leq  \frac{2k_e\epsilon^2}{n} + \frac{\log(n)^{1.1}}{n} \norm{f^*}_{L^2(\rho)}
        \\&\qquad+ c_{f,T_\lambda}^2 \paren{{\cal L}_k(S\hat\Theta)- {\cal L}_k(S\Theta_{*})}
        +  c_{f,k}.
        \label{eq:conv}
    \end{align}
    where $\epsilon^2$ is the noise level of $Y$ (the supremum of conditional variances), $k_e \leq k$ is the effective dimension of the representation $\psi = \Theta\phi$ on the downstream task, $c_{f,k} \leq (k_\lambda - k)_+ \norm{f^*}^2_{L^2(\rho_\X)}$ is a constant relating to the concentration of the energy of $f^*$ the target function on the downstream task with respect to the eigenspaces of $T_\lambda$, $c_{f,T_\lambda} \leq \norm{T_\lambda^{-1}f^*}$ is a similar constant taking into account the decay of eigenvalues of $T_\lambda$, and the index $k$ in ${\cal L}_k$ indicates that we search over $\psi:\X\to\R^k$. 
\end{theorem}

\begin{figure}[t]
  \centering
  \includegraphics{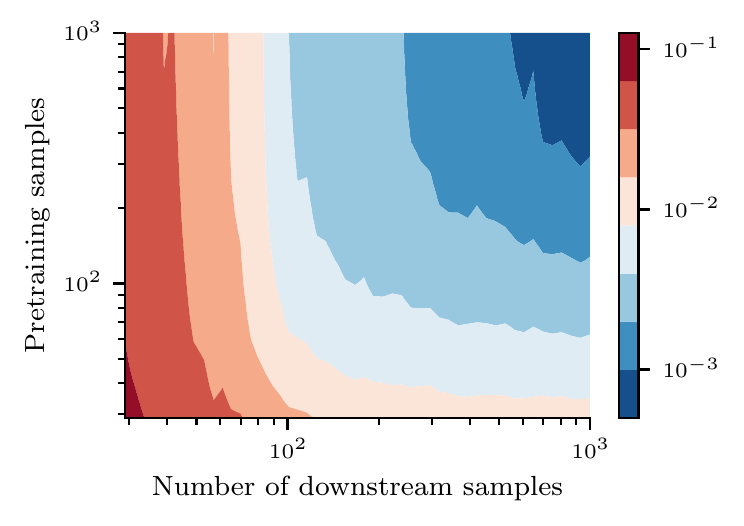}
  \vspace{-1em}
  \caption{
	\emph{Empirical downstream performance} on a simple task (detailed in Appendix \ref{app:examples}) depends on the number of both downstream samples ($x$-axis) and pretraining ($y$-axis) in log-scale.
  Along the left hand side of the plot, convergence rates of $n_{pre}^{-1/2}$ are observed with respect to the number of pretraining samples (Theorem \ref{thm:sgd}) while along the top, convergence rates of $n_{down}^{-1}$ are observed with respect to the number of downstream samples (Theorem \ref{thm:conv_simple}).
  At the bottom, a saturation phenomenon is observed where added downstream samples do not result in noticeable benefits as the excess risk stalls at ${\cal R}(\Pi_{\hat{\cal F}} f^*) - {\cal R}(f^*) > 0$.
}
  \label{fig:error_cl}
  \vspace{-1em}
\end{figure}

The results of Theorem \ref{thm:conv_simple} can be seen as a variance-bias decomposition. 
A variance term, due to misspecified linear regression, displays rates in $k\log(n)/n$. 
The $\log(n)$ factor is actually an artefact of our derivations, and could be removed with Theorem 1 of \citet{mourtada2022}.
A bias term relates to the approximation error. 
It captures both the hardness to learn $f^*$ with $T_\lambda$ through the constants~$c_{f,T_\lambda}$ and~$c_{f,k}$, and the error made on the pre-training task through~${\cal L} - {\cal L}^*$.
Note that the proof of Theorem \ref{thm:conv_simple} mindfully avoids bounding~$c_{f,T_\lambda}$ by $\norm{T_\lambda^{-1}}_{\op}\norm{f^*}$ which would introduce the inverse of the spectral gap of $T_\lambda$ in the bound, and would not characterize well situation where the target function $f^*$ is actually easy to learn with $T_\lambda$.
We also remark that for classification tasks, recent work shows that under mild assumptions on $\rho_\X$, and low noise conditions, it should be possible to convert the rates of Theorem~\ref{thm:conv_simple} into exponentially fast rates for the zero-one loss \citep{Cabannes2022rates}.
This is particularly the case under the cluster setting studied by~\citet{haochen_provable_2021}.\footnote{See also \citet{Rigollet2007} for fast rates in this setting.}
The theoretical convergence rates of Theorem \ref{thm:conv_simple} are validated experimentally in Figure~\ref{fig:error_cl}.

\subsection{Pretraining guarantees}

Theorem \ref{thm:conv_simple} above states that representations with small pretraining loss can solve downstream tasks that satisfy Assumption~\ref{ass:source} but do not address how difficult it is to find such representations.
This section aims to bridge that gap.
The following theorem details convergence rates of the empirical risk minimizer using Rademacher complexity arguments.

\begin{theorem}[Empirical risk minimizer]
    \label{thm:rad}
    Let $\Theta_n\in\R^k\otimes {\cal H}$ be the minimizer of the unbiased regularized empirical version of ${\cal L}$ based on a dataset ${\cal D}_n$. 
    Assume that ${\cal D}_n$ is built from $n$ input samples $(X_i) \sim \mu_\X^{\otimes n}$ and $m$ augmentation per samples $(\xi_{ij}) \sim\mu\vert_{X_i}^{\otimes m}$, then the average excess risk is bounded by
    \begin{equation}
        \E_{{\cal D}_n}[{\cal L}(S\Theta_n)] - {\cal L}(S\Theta) 
        \leq \frac{12\kappa^2 k}{\lambda\sqrt{n}} \paren{1 + \frac{\kappa^2 k}{\lambda}},
    \end{equation}
    where $\kappa$ is a bound on $\norm{\phi(X)}$.
\end{theorem}

Note that the proof of Theorem \ref{thm:rad} proceeds with a loose bound on the variance of the empirical risk, which is mainly due to the difficulty in dealing with non-exchangeability of the samples $(\xi_{ij})$.
In essence, the ease of minimizing~${\cal L}$ depends on both the variance of~${\cal L}$ when estimated with empirical data (or the variance of stochastic gradients when performing SGD), and the size of the space where we aim to find representations $\psi:\X\to\R^k$.
With stronger assumptions on the distribution of $\phi(\xi)$ (e.g., data are clustered, and the law of $\paren{\xi\midvert X}$ is invariant per cluster), one could show much better behavior of the excess risk with respect to the number of augmentations (e.g., replacing $n$ by the minimum number of points in one cluster multiplied by the number of views).
The following theorem states convergence rates with a stochastic gradient descent algorithm capturing such a potential situation.
Proofs and technicalities, based on convex optimization literature, are detailed in Appendix~\ref{app:upstream}.

\begin{theorem}[Sharper bounds]
    \label{thm:sgd}
    There exists an implementable algorithm that guarantees an average excess risk
    \begin{align}
        \nonumber
        &\E_{{\cal D}_n}[{\cal L}(S\Theta_n)] - {\cal L}(S\Theta)
        \\&\qquad\qquad\leq 3\kappa^2 c_\lambda c_\lambda'\sqrt{\frac{\sigma_X^2}{n} + \frac{\sigma_\xi^2}{nm}} + \frac{4\kappa^6 c_\lambda^2}{n}   
    \end{align}
    where $c_\lambda = 1 + \kappa^2 k_\lambda/\lambda$, $c_\lambda' = 1 + k_\lambda^2 / \lambda^2$, $k_\lambda$ is the number of positive eigenvalues of $T_\lambda$, $\kappa$ is a bound on $\norm{\phi}$, $\sigma_X$ relates to the variance of $\E\bracket{\psi(\xi)\midvert X}$, and $\sigma_\xi$ relates to the average variance of $\paren{\xi\midvert X}$.
    Moreover, when $K = SS^\top$ or the covariance of the $\phi(\xi)$ has a finite number of positive eigenvalues (e.g. $\X$ finite or $\cal H$ finite dimensional), with $c_K$ a constant that relates to the condition number of $K$, this bound can be tightened to
    \begin{equation}
        \E_{{\cal D}_n}[{\cal L}(S\Theta_n)] - {\cal L}(S\Theta)
        \leq \frac{4 c_K^2 c_\lambda^2}{n}.
    \end{equation}
\end{theorem}

In the setting studied in \citet{haochen_provable_2021}, we stress that Theorem \ref{thm:sgd} guarantees convergence rates of $O(n^{-1})$ rather than $O(n^{-1/2})$ on the upstream loss. 
In effect, we improve the rates of \citet[Theorem 4.3]{haochen_provable_2021} from $n^{-1/2}$ to $n^{-1}$ on both pretraining and downstream tasks.

\section{Practical insights}
In this section, we relate our theory to commonly observed phenomena when training SSL algorithms and offer best practices informed by our findings.

\subsection{The downstream problem }

Two regimes should be distinguished for the downstream problem.
When few downstream samples are available, {\em few-shot learning} requires a small effective dimension $k_e$ \eqref{eq:conv} to lower the estimation error and avoid fitting noise.
Limiting $k_e$ (or equivalently the capacity of $\hat{\cal F}$) can be done either by decreasing the representation dimension $k$ or applying regularization on downstream tasks.
This theoretical trade-off between effective dimension and number of downstream examples is illustrated empirically by \citet[Figure 6]{he_exploring_2022}.
On the contrary, when accessing a substantial amount of data for training downstream tasks, one could confidently augment the representation dimension $k$ to decrease the approximation error.
This was notably observed on large-scale datasets by \citet[Figure 1]{Garrido_rankme_2022}: as $k$ increases, the effective dimension $k_e$ converges to a limit, and the downstream performance keeps increasing until this limit is reached.
Remarkably, our theory explains this phenomenon: since $k_\lambda$ is finite, as $k$ increases, the effective dimension $k_e$ will be bounded by the limiting case where~$k=k_\lambda$.\footnote{Note that without regularization, $(1-\beta)I + \beta T$ is not trace-class so $k_e$ will not converge as $k$ increases.}

\subsection{The pretraining problem }
\emph{Usefulness of multiple augmentations per sample.}
Theorem~\ref{thm:sgd} shows how multiple augmentation such as multi-crops can result in faster convergence to an optimal representation~$\psi$.
There, the variance of the empirical risk depends on both~$\sigma_X$ due to variation over inputs, and~$\sigma_\xi$ due to variations over resulting views after augmentation.
With multiple augmentations per sample, one can reduce the latter variance and improve performance, which was observed with the introduction of multicrops in \citet{caron2020unsupervised}.
However, when the total amount $m \times n$ of pre-processed data is held fixed, it is generally better to process many inputs with two views $m=2$, rather than a few inputs with many augmentations. This finding matches the empirical observations of \citet{bardes_vicreg_2022} that if available, fresh samples are always better than more views.

\begin{figure}[t]
  \centering
  \includegraphics{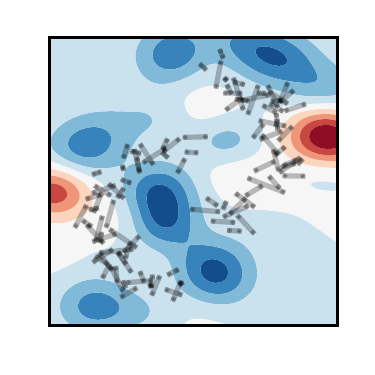}
  \includegraphics{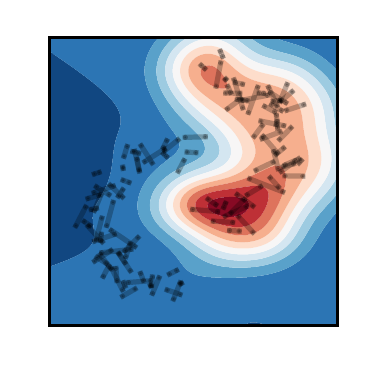}
  \vspace{-1em}
  \caption{
  \emph{Capacity trouble.}
  Level lines of the top eigenfunction of empirical estimate of $T_\lambda$ for negligible regularization (left) and small regularization $\lambda$ (right). 
  Experiments are done with a Gaussian kernel with scale about one tenth of the problem diameter, augmentations are represented as black dots, connected by a line when they provide from the same input $X$.
  When $\lambda$ is negligibly small, capacity troubles arise, infringing the recovery of the cluster structure on the right.}
  \label{fig:capacity}
  \vspace{-1em}
\end{figure}

\emph{Capacity trouble in pretraining.}
Theorems \ref{thm:rad} and \ref{thm:sgd} show that, without regularization restricting the capacity of the model of computation, one cannot expect to meaningfully solve the pretraining task.
This is captured by the quantity $c_\lambda$ that goes to infinity as $\lambda$ goes to zero.
Such issues related to the lack of regularization commonly arise in practice.
Given $n\times m$ upstream samples $(\xi_{ij})$, the empirical minimization of VICReg can be implemented by approximating $\mu$ with $\sum_{ij}\delta_{(i,\xi_{i,j})} / nm$.
In this setting, $T$ is the adjacency matrix of a graph with as many connected components as there are inputs $n$, as detailed in Appendix \ref{app:experiments}.
All the connected components define a maximal eigenvector of the empirical approximation to $T$, leading to a ``collapsed'' representation $\psi = \sum_{j} \delta_{\xi_{ij}} / m$.
Regularizing forces the optimizer to search for representation inside the space $\Psi$ which mixes those small clusters letting meaning eigenfunctions emerging (see Figure \ref{fig:capacity} for an illustration).

\subsection{Guidelines for practitioners}

Our theoretical study provides several insights that may be useful for SSL practitioners. We highlight a few~below.

\emph{Avoiding collapse.}
The common collapse phenomenon, where pretraining ends up fitting noise instead of learning useful features, may be addressed in several ways.
Our theory suggests to:
\begin{itemize}
    \item \emph{Reduce the model capacity}, through \emph{regularization} (e.g., early stopping), or \emph{simpler architectures} (e.g., a shallow CNN instead of an MLP). 
    As a consequence, $\Psi$ will have a lower effective dimension, $K$ will encourage ``simpler'' representations that can be learned with less data, even without any data augmentation. 
    \item \emph{Use stronger augmentations.} 
    $T$ will become more compact, reducing $k_\lambda$ the dimension of the ``positive eigenspace'' of $T_\lambda$.
    The ideal $\psi$ will exhibit more structure, thereby its search could be reduced to smaller spaces, making it harder to collapse.
\end{itemize}

\emph{Incorporating priors.}
Representations are typically used for solving downstream tasks, thus it is crucial to incorporate the right priors during pretraining.
Our theory showcases the important role of several factors.
\emph{(i) Augmentations} determine the nature of the invariance that is enforced (e.g., low variations, short-range dependencies, translation invariance); affects top eigenfunctions of~$T$.
\emph{(ii) Architecture} promotes ``simple'' representations (e.g., smoothness, locality); affects top eigenfunctions of~$K$.
\emph{(iii) Regularization} balances the interplay between augmentations and architecture; affects top eigenfunctions of~$T_\lambda$.
\emph{(iv) Pretraining data} impacts both~$T$ and~$K$ and their eigenfunctions, e.g., through clustering structure, or natural image statistics.

\section{Conclusion}

This paper presents a theoretical framework for studying self-supervised learning in the kernel regime.
It examines three key operators and their impact on convergence and generalization: $T$ linked with augmentations, $K$ linked with architecture choices, and $T_\lambda$ resulting from their interplay and tuned by the parameter $\lambda$.
Our analysis offers useful guarantees and practical guidelines for practitioners to improve the stability and performance of SSL algorithms. Looking beyond the kernel regime, future work can 
We left for future work the extension of our analysis beyond the kernel setting, in particular to understand non-linear training dynamics in finite width neural network and feature learning capabilities within layers.
Moreover, future studies could encompass more techniques that enhance performance in SSL, these include projecting representations before enforcing losses, batching the data, or applying different loss functions.

%% file: drawings/original.tex
\begin{tikzpicture}[main/.style = {draw, rectangle,outer sep = 0,minimum height=1em, minimum width=0.3cm,node distance=0.33cm}]
\node[main,fill=pltblue] (1) [] {};
\node[main,fill=pltblue] (2) [right of=1] {};
\node[main,fill=pltred] (3) [right of=2] {};
\node[main,fill=pltred] (4) [right of=3] {};
\node[main,fill=pltblue] (5) [right of=4] {};
\node[main,fill=pltblue] (6) [right of=5] {};
\node[main,fill=pltred] (7) [right of=6] {};
\node[main,fill=pltblue] (8) [right of=7] {};
\node[main,fill=pltred] (9) [right of=8] {};
\end{tikzpicture}  

%% file: drawings/flip.tex
\begin{tikzpicture}[main/.style = {draw, rectangle,outer sep = 0,minimum height=1em, minimum width=0.3cm,node distance=0.33cm}]
\node[main,fill=pltblue] (1) [] {};
\node[main,fill=pltred,ultra thick] (2) [right of=1] {};
\node[main,fill=pltblue,ultra thick] (3) [right of=2] {};
\node[main,fill=pltred] (4) [right of=3] {};
\node[main,fill=pltblue] (5) [right of=4] {};
\node[main,fill=pltblue] (6) [right of=5] {};
\node[main,fill=pltblue,ultra thick] (7) [right of=6] {};
\node[main,fill=pltblue] (8) [right of=7] {};
\node[main,fill=pltred] (9) [right of=8] {};
\end{tikzpicture} 

%% file: drawings/crop.tex
\begin{tikzpicture}[main/.style = {draw, rectangle,outer sep = 0,minimum height=1em, minimum width=0.3cm,node distance=0.33cm}]
\node[main,fill=pltgray] (1) [] {};
\node[main,fill=pltgray] (2) [right of=1] {};
\node[main,fill=pltred] (3) [right of=2] {};
\node[main,fill=pltred] (4) [right of=3] {};
\node[main,fill=pltblue] (5) [right of=4] {};
\node[main,fill=pltgray] (6) [right of=5] {};
\node[main,fill=pltgray] (7) [right of=6] {};
\node[main,fill=pltgray] (8) [right of=7] {};
\node[main,fill=pltgray] (9) [right of=8] {};
\end{tikzpicture} 

%% file: drawings/translation.tex
\begin{tikzpicture}[main/.style = {draw, rectangle,outer sep = 0,minimum height=1em, minimum width=0.3cm,node distance=0.33cm}]
\node[main,fill=pltred] (1) [] {};
\node[main,fill=pltblue] (2) [right of=1] {};
\node[main,fill=pltblue] (3) [right of=2] {};
\node[main,fill=pltblue] (4) [right of=3] {};
\node[main,fill=pltred] (5) [right of=4] {};
\node[main,fill=pltred] (6) [right of=5] {};
\node[main,fill=pltblue] (7) [right of=6] {};
\node[main,fill=pltblue] (8) [right of=7] {};
\node[main,fill=pltred] (9) [right of=8] {};
\end{tikzpicture}  

%% file: drawings/index.tex
\begin{tikzpicture}[main/.style = {draw, rectangle,outer sep = 0,minimum height=1em, minimum width=0.3cm,node distance=0.33cm}]
\node[main,fill=pltred] (1) [] {};
\node[main,fill=pltblue] (2) [right of=1] {};
\node[main,fill=pltred] (3) [right of=2] {};
\node[main,fill=pltblue] (4) [right of=3] {};
\node[main,fill=pltblue] (5) [right of=4] {};
\node[main,fill=pltred] (6) [right of=5] {};
\node[main,fill=pltred] (7) [right of=6] {};
\node[main,fill=pltblue] (8) [right of=7] {};
\node[main,fill=pltblue] (9) [right of=8] {};
\end{tikzpicture}  

%% file: drawings/legend.tex
\begin{tikzpicture}[main/.style = {draw, rectangle,outer sep = 0,minimum height=1em, minimum width=0.3cm,node distance=0.33cm}]
\node[main,fill=pltred] (1) [] {};
\end{tikzpicture} -1 bit,  $\; \; \; \; $ \begin{tikzpicture}[main/.style = {draw, rectangle,outer sep = 0,minimum height=1em, minimum width=0.3cm,node distance=0.33cm}]
\node[main,fill=pltblue] (1) [] {};
\end{tikzpicture} +1 bit, $\; \; \; \; $ \begin{tikzpicture}[main/.style = {draw, rectangle,outer sep = 0,minimum height=1em, minimum width=0.3cm,node distance=0.33cm}]
\node[main,ultra thick] (1) [] {};
\end{tikzpicture} flipped bit, $\; \; \; \; $ \begin{tikzpicture}[main/.style = {draw, rectangle,outer sep = 0,minimum height=1em, minimum width=0.3cm,node distance=0.33cm}]
\node[main,fill=pltgray] (1) [] {};
\end{tikzpicture} random bit

%% file: appendix/proof.tex
\section{Mathematical details and simple proofs}
\label{app:proof}

\subsection{Technical details}

Several technicalities were left implicit in the main text, we discuss it now.
In particular, we assumed that there exists a minimizer $f^*$ of the risk ${\cal R}$ which is true when $\Y$ is finite, or when $\ell$ is the least-square loss and $\paren{Y\midvert X}$ has a second order moment almost everywhere.
Moreover, in the proof for least-squares, we will assume for simplicity that $\Y=\R$.
The same derivations still holds true when $\Y=\R^k$ although it requires slight precautions such as working in $\Y\otimes{\cal H}$ rather than in ${\cal H}$ \citep[see][for example]{Cabannes2022rates}.

\paragraph{Integers.}
The representation dimension $k$ is an integer, and $[k]$ denotes the set $\brace{1, 2, \cdots, k}$.
For simplicity, we abuse notations and denote by $\N$ the set of strictly positive integers.

\paragraph{Geometry.}
The product $A\times B$ denotes the set of elements $(a, b)$ with $a\in A$ and $b\in B$.
The notation $a^\top$ denotes the adjoint of $a$ which depends on the Hilbertian metric space one consider $a$ to be part of ({\em e.g.} the adjoint in $L^2(\mu_\Xi)$ is not the same as the adjoint in $L^2(\rho_\X)$).
The notation $a^\top b$ denotes the scalar product $\scap{a}{b}$ with the Hilbertian metric space $a$ and $b$ are understood to be part of.
The Hilbertian norm on matrices or operators is denoted by $\norm{\cdot}_F$ (Frobenius), $\norm{\cdot}_2$ or $\norm{\cdot}_{HS}$ (Hibert-Schmidt).
The operator norm is denoted by $\norm{\cdot}_{\op}$.
Moreover, the identity is always denoted by $I$.

\paragraph{Distributions.}
In order to define probabilities, $\X$ and $\Y$ are assumed to be Polish spacesa endowed with the Borel topologies.
We used the simplex notation $\prob{A}$ to design the set of probability measure on $A$, and the tensor notations $\rho^{\otimes n}$ to denotes the measure of $n$ independent random variables all distributed according to $\rho$.
The notation $\phi_\# \rho$ denote the measure of $\phi(X)$ when $X$ is distributed according to the measure $\rho$.
The notation $\rho \ll \mu$ means that for any measurable set $X$ the fact that $\mu(X) = 0$ implies that $\rho(X) = 0$.
The notation $\delta_x$ denotes the Dirac distribution, which satisfies $\scap{f}{\delta_x} = f(x)$ using the duality bracket between functions and distributions.
For any distribution $p$, the space $L^2(p)$ is made of measurable functions that are square-integrable.

\paragraph{Functions.}
All functions such as $\ell$, $f$, $\psi$, $\phi$, and so on, are restricted to be measurable.
The notation $\circ$ denote the composition of functions $f\circ g(\cdot) = f(g(\cdot))$.
A function $f:\X\to\Y$ is understood as an element of $\Y^\X$, and we use some isomorphism such as $(\R^k)^\X = (\R^\X)^k$.
We use the notation $\R^k\otimes{\cal H}$ to denote linear bounded operator from ${\cal H}$ to $\R^k$.
This tensor product notation generalizes matrix notations with $\R^k\otimes \R^{d_h} = \R^{k\times d_h}$. In particular,
\[
    \Psi^k = \brace{x\to\Theta\phi(x)\midvert \Theta \in \R^k\otimes{\cal H}}.
\]
For $\Theta\in \R^k\otimes {\cal H}$, one can write $\Theta$ in row-style as an element of ${\cal H}^k$ as well as its adjoint $\Theta^\top \in {\cal H}\otimes \R^k$ in column-style which follows from the fact that ${\cal H}^k$ is self-adjoint when endowed with the $\ell^2$-product topology.

\subsection{Proof of Remark \ref{lem:same_loss}}
Let us characterize \eqref{eq:vic} in order to easily implement it with unbiased stochastic gradient.
We need to get the expectation outside the norm.
This can be done with the following derivations
\begin{align*}
    \norm{\E[\psi(\xi)\psi(\xi)^\top] - I}^2
    &= \trace\paren{(\E[\psi(\xi)\psi(\xi)^\top] - I)(\E[\psi(\xi')\psi(\xi')^\top] - I)}
    \\&= \E_{\xi, \xi'}\bracket{\trace\paren{\psi(\xi)\psi(\xi)^\top \psi(\xi')\psi(\xi')^\top}} - 2\E_\xi\bracket{\trace\paren{\psi(\xi)\psi(\xi)^\top}} + \trace(I)
    \\&= \E_{\xi, \xi'}\bracket{(\psi(\xi')^\top\psi(\xi))^2} - 2\E_\xi\bracket{\norm{\psi(\xi)}^2} + k.
\end{align*}
For the first part, we get
\[
    \E_X\E_{\xi,\xi'}\bracket{\norm{\psi(\xi) - \psi(\xi')}^2\midvert X}
    = 2\E_{\xi}[\norm{\psi(\xi)}^2] - 2\E_X\E_{\xi, \xi'}\bracket{\psi(\xi)^\top\psi(\xi')\midvert X}     
\]
As a consequence,
\begin{equation}
    \label{eq:vic_unbiased}
    {\cal L}(\psi; \beta) = 
    2(\beta - 1)\E_{\xi}[\norm{\psi(\xi)}^2] - 2\beta\E_X\E_{\xi, \xi'}\bracket{\psi(\xi)^\top\psi(\xi')\midvert X} 
    + \E_{\xi, \xi'}\bracket{(\psi(\xi')^\top\psi(\xi))^2} + k.
\end{equation}
In particular, when $\beta=1$, we retrieve the spectral contrastive loss introduced by \citet{haochen_provable_2021},
\[
    {\cal L}(\psi) 
    = - 2\E_X\E_{\xi,\xi'}\bracket{\psi(\xi)^\top \psi(\xi')\midvert X}
    + \E_{\xi,\xi'}[(\psi(\xi)^\top \psi(\xi'))^2] 
    + k.
\]

\subsection{Proof of Lemma \ref{lem:close}}
First, notice that if we define for $\psi:\X\to\R$, the mapping 
\(
    \omega(\psi) \to \E_X[\E_{\xi,\xi'}\bracket{\norm{\psi(\xi)-\psi(\xi)}^2\midvert X}]  
\), $\omega$ is a quadratic form on $L^2(\mu_\Xi)$.
As a consequence, it can be represented with a linear self-adjoint operator $L$ on $L^2(\mu_\Xi)$ such that 
\(
    \omega(\psi) = \scap{\psi}{L\psi}_{L^2(\mu_\Xi)}.
\)
Because $\omega(\psi) \geq 0$, we have $L \succeq 0$ (with $\succeq$ the Loewner order on symmetric operators, i.e. $A\succeq B$ if $A-B$ is positive).
The following lemma show that $L$ is bounded.

\begin{lemma}
    For any $\psi\in L^2(\mu_\Xi)$, $\omega(\psi) \leq 2\norm{\psi}^2_{L^2(\mu_\Xi)}$. As a consequence, $L \preceq 2I$.
\end{lemma}
\begin{proof}
    This follows from the fact that
    \begin{align*}
        \omega(\psi)
        &= \E_X[\E_{\xi,\xi'}\bracket{\norm{\psi(\xi) - \psi(\xi')}^2\midvert X}]
        \\&= \E_X[\E_{\xi,\xi'}\bracket{\norm{\psi(\xi) - \E\bracket{\psi(\xi)\midvert X} + \E\bracket{\psi(\xi)\midvert X} - \psi(\xi')}^2\midvert X}]
        \\&= \E_X[\E_{\xi,\xi'}\bracket{\norm{\psi(\xi) - \E\bracket{\psi(\xi)\midvert X}}^2 + \norm{\E\bracket{\psi(\xi)\midvert X} - \psi(\xi')}^2\midvert X}]
        \\&= 2\E_X[\E_{\xi}\bracket{\norm{\psi(\xi) - \E\bracket{\psi(\xi)\midvert X}}^2\midvert X}]
        \\&= 2\min_{\psi_0:\X\to\R}\E_X[\E_{\xi}\bracket{\norm{\psi(\xi) - \psi_0(X)}^2\midvert X}]
        \\&\leq 2\E_X[\E_{\xi}\bracket{\norm{\psi(\xi)}^2\midvert X}]
        = 2\E_{\xi}\bracket{\norm{\psi(\xi)}^2}
        = 2\norm{\psi}_{L^2(\mu_\Xi)}
    \end{align*}
    Hence for any $\psi$, with the $L^2(\mu_\Xi)$ geometry we have $\psi^\top L\psi \leq \psi^\top \psi$, which implies, since $L$ is self-adjoint, that $\norm{L}_{\op} \leq 2$. 
\end{proof}

Because $0 \preceq L \preceq 2I$, let us introduce $T = (2I-L) / 2$, we have $0 \preceq T \preceq 1$ and, with the $L^2(\mu_\Xi)$ geometry, for $\psi:\X\to\R^k$
\[ 
    {\cal L}(\psi) = \beta\sum_{i=1}^k \psi_i^\top L \psi_i + \norm{\E[\psi(\xi)\psi(\xi')^\top] - I}^2
    = 2\beta\sum_{i=1}^k \psi_i^\top (I-T) \psi_i + \norm{\E[\psi(\xi)\psi(\xi')^\top] - I}^2.
\]

In order to diagonalize all operators without relying on integral formulations of the spectral theorem, we introduce the following mild assumption.

\begin{assumption}
    \label{ass:diag}
    Assume that $T$ has a pure point spectrum.
\end{assumption}

\begin{example}
    When the distribution of augmentation have a density $p$ with respect to a any measure and $(x, \xi)\to p\paren{\xi\midvert x}/p(\xi)$ is in $L^2(\mu)$, or when $\X$ is finite, $T$ can be shown to be a compact operator, hence to have a pure point spectrum according to the spectral theorem.
\end{example}
\begin{proof}
    When $\X$ is finite, the $L^2$ spaces are finite dimensional, hence locally compact, which implies that all operators are compact.
    To prove the case with density, let us develop $T$ as an integral operator.
	We have, in $L^2(\mu_\Xi)$ geometry, for $f:\X\to\R$
	\begin{align*}
	  2f^\top (I - T) f 
	  &= \E_X\E\bracket{\norm{f(\xi) - f(\xi')}^2\midvert X}
	  = \E_X\E\bracket{\norm{f(\xi)}^2 + \norm{f(\xi')}^2 - 2\scap{f(\xi)}{f(\xi')}\midvert X}
	\\&= 2f^\top f - 2\E_X\E\bracket{\scap{f(\xi)}{f(\xi')}\midvert X}.
	\end{align*}
	This allow us to identify $T$ with the inner product, we have for $g:\X\to\R$ and $p$ the density of augmentations
	\begin{align*}
	  f^\top T g 
	  &= \E_X\E\bracket{\scap{f(\xi)}{g(\xi')}\midvert X}
	  = \int \scap{f(\xi)}{g(\xi')} p\paren{\xi \midvert x} p\paren{\xi'\midvert x} \diff \xi' \diff \xi \mu_\X(\diff x)
	\\&= \int \mu_\Xi(\diff \xi) \scap{f(\xi)}{\int \mu_\Xi(\diff \xi') g(\xi') \frac{\int \mu_\X(\diff x) p\paren{\xi \midvert x} p\paren{\xi'\midvert x}}{p(\xi)p(\xi')}}.
	\end{align*}
	As a consequence, one can consider $T$ as the integral operator in $L^2(\mu_\Xi)$ linked with the kernel 
	\[k(\xi, \xi') = \frac{\int \mu_\X(\diff x) p\paren{\xi \midvert x} p\paren{\xi'\midvert x}}{p(\xi)p(\xi')}.\]
	When this kernel is bounded, or simply when $\xi \to k(\xi, \xi)$ belongs to $L^2(\mu_\Xi)$, $T$ is trace-class hence compact.
\end{proof}

Let us now prove in order to minimize ${\cal L}$, one should take the eigenfunctions of the operator $(1-\beta)I + \beta T$ whose corresponding eigenvalues are the biggest positives ones.
It can be proven with simple geometry in a somewhat abstract space.
To do so, remark that $\psi:\X\to\R^k$ in $L^2(\mu_\Xi, \X, \R^k)$ can be represented as $\tilde\psi\in\R^k \otimes L^2(\mu_\Xi, \X, \R)$ with the linear map that associates $(\psi_i^\top \phi)$ to a function $\phi \in L^2(\mu_\Xi ,\X,\R)$.
Let denote $T_\beta = (1-\beta) I + \beta T$, the upstream loss can be characterized as
\begin{align*}
    {\cal L}(\psi) 
    &= 2\beta \sum_{i\in[k]} \scap{\psi_i}{(I - T)\psi_i}
    + \norm{\E_\xi[\psi(\xi)\psi(\xi)^\top] - I}^2
    \\&= 2\beta \sum_{i\in[k]} \scap{e_i^\top\psi}{(I - T)\psi^\top e_i}
    + \norm{\E_\xi[\sum_{i,j\in[k]}e_i^\top\psi(\xi)\psi(\xi)^\top e_j e_i e_j^\top] - I}^2
    \\&= 2\beta \sum_{i\in[k]} e_i \tilde\psi(I - T)\tilde\psi^\top e_i
    + \norm{\sum_{i, j\in[k]} e_i^\top\tilde\psi\tilde\psi^\top e_j e_i e_j^\top - I}^2
    \\&= 2\beta \trace\paren{\tilde\psi (I - T)\tilde\psi^\top} + \norm{\tilde\psi\tilde\psi^\top - I}^2
    = 2\beta \trace\paren{\tilde\psi (I - T)\tilde\psi^\top} + \trace\paren{\paren{\tilde\psi\tilde\psi^\top - I}^2}.
    \\&= \trace\paren{2\beta\tilde\psi (I - T)\tilde\psi^\top + \tilde\psi\tilde\psi^\top \tilde\psi\tilde\psi^\top - 2 \tilde\psi\tilde\psi^\top + I}.
    \\&= \trace_{L^2(\mu_\Xi)}\paren{\tilde\psi^\top \tilde\psi\tilde\psi^\top\tilde\psi + (2\beta (I - T) - 2 I)\tilde\psi^\top\tilde\psi} + k.
    \\&= \trace_{L^2(\mu_\Xi)}\paren{\tilde\psi^\top \tilde\psi\tilde\psi^\top\tilde\psi - 2 T_\beta\tilde\psi^\top\tilde\psi} + k.
    = \trace_{L^2(\mu_\Xi)}\paren{(\tilde\psi^\top \tilde\psi - T_\beta)^2 - T_\beta^2} + k.
\end{align*}

In order to find the minimizer of ${\cal L}$ with this new characterization, slight precautions are needed here since the two operators are not trace-class.
The following lemma takes those precautions in order to finish the proof.

\begin{lemma}
    \label{proof-charac}
    Let $A$ be a self-adjoint operator on $L^2(\mu_\Xi)$. Assume that there exists $c$ such that $A \preceq c I$ and that $A$ is pure-point spectrum. Then if $(\lambda_i, f_i)$ denote the eigen-decomposition of $A$ with $\lambda_i$ in decreasing order, the minimization of $\trace\paren{(B-A)^2 - B^2}$ under the constraint that $B$ is a self-adjoint positive operator of rank at most $k$, is reached for $B = \tilde\psi^\top\tilde\psi$ with $\psi:\X\to\R^k$ such that $\psi_i = \max(0, \lambda_i)^{1/2} f_i$.
\end{lemma}
\begin{proof}
Let us decompose $A$ into a positive part $A_+ \succeq 0$ and a negative part $A_- \succeq 0$ such that $A = A_+ - A_-$.
Using the fact that $B$ is positive self-adjoint, we get
\begin{align*}
    \trace\paren{(B-A)^2 - A^2} &= \trace{B^2 - 2B^{1/2}AB^{1/2}} = \trace\paren{B^2 - 2B^{1/2}A_+ B^{1/2}} + 2\trace\paren{B^{1/2}A_-B^{1/2}} 
    \\&\geq \trace\paren{B^2 - 2B^{1/2}A_+ B^{1/2}}.
\end{align*}
Let us decompose $B$ into $k$ symmetric operators of rank at most one as $B = \sum_{i=1}^k B_i$ such that $B_iB_j = 0$ for any $i \neq j\in[k]$.
Using the different properties of the operators introduced, we proceed with
\begin{align*}
    \trace\paren{(B-A)^2 - A^2} 
    &\geq \sum_{i=1}^k \trace\paren{B_i^2} - 2\trace\paren{B_iA_+}
    = \sum_{i=1}^k \norm{B_i}^2_{\op} - 2\norm{B_iA_+}_{\op}
    \\&\geq \sum_{i=1}^k \norm{B_i}^2_{\op} - 2\norm{B_i}_{\op}\norm{\Pi_{B_i} A_+}_{\op}
    \geq \sum_{i=1}^k \norm{B_i}^2_{\op} - 2\norm{B_i}_{\op}\big\|\prod_{j< i}(I - \Pi_{B_j}) A_+\big\|_{\op}
    \\&= \sum_{i=1}^k \paren{\norm{B_i}_{\op} - \big\|\prod_{j< i}(I - \Pi_{B_j}) A_+\big\|_{\op}}^2 - \big\|\prod_{j< i}(I - \Pi_{B_j}) A_+\big\|_{\op}^2
    \\&\geq -\sum_{i=1}^k \big\|\prod_{j< i}(I - \Pi_{B_j}) A_+\big\|_{\op}^2
    \geq -\sum_{i=1}^k \sigma_i(A_+)
\end{align*}
where $\Pi_B$ denote the orthogonal projector on the image of $B$, and $\sigma_i(A)$ the $i$-th singular value of $A$ (monotonically ordered with $\sigma_1(A)$ the biggest).
The last inequality is due to the Courant-Fisher min-max principle,
This inequality can be achieved with $\Pi_{B_i}$ the projection on the $i$-th eigenspace of $A$ and $\norm{B_i}_{\op} = \sigma_i(A)$.
In other terms, $B$ should match the first $k$ positive eigenvalues of $A$.
In the case where $A$ has less than $k$ positive eigenvalues, then $B$ should match all the positive eigenvalues and be null on the range of $A_-$.
\end{proof}

The following is a direct corollary of the proof above.
\begin{proposition}[Uniqueness of minimizers]
    The minimizers of ${\cal L}$ are unique up to orthogonal transformations and eigenfunction picking.
    More specifically, if $U\in\R^{k\times k}$ is orthogonal, i.e. $U^\top U = I$, then ${\cal L}(\psi) = {\cal L}(U\psi)$; and if $\lambda_k = \lambda_{k+1}$, one can choose different eigenfunctions as $f_k$ in the eigen-decomposition $(\lambda_i,f_i)$ of $T_\beta$.     
\end{proposition}

\subsection{Proof of Lemma \ref{lem:close_rkhs}}
Let us consider $\psi:\X\to\R^k$ with $\psi_i = f_{\theta_i}$ for $\theta_i \in {\cal H}$ and $S:{\cal H}\to L^2(\mu_\Xi); \theta \to \theta^\top \phi(\cdot)$.
We can use the tensor notations introduced earlier to parameterized $\psi = S\Theta^\top$ with $\Theta = (\theta_i)_{i\in[k]}$ seen as an element of $\R^k\otimes {\cal H}$.
The proof of Lemma \ref{lem:close_rkhs} follows from the fact that
\[
    \norm{\Theta}_2^2 
    = \norm{\Theta^\top}_2^2 
    = \scap{S\Theta^\top}{(S^\top S)^{-1}S\Theta^\top}_{L^2(\mu_\Xi)}
    = \sum_{i=1}^k S\theta_i^\top K^{-1} S\theta_i
    = \sum_{i=1}^k \psi_i^\top K^{-1} \psi_i.
\]
Since $\Psi = S({\cal H}) = \ima S = \ima K^{1/2} = K^{1/2}(L^2(\mu_\Xi))$, one can consider $K^{-1}$ as the inverse of $K$ such that for $\psi_i\in\ker K$, $\psi_i^\top K^{-1}\psi_i = +\infty$.
This is what we implicitly assumed in the main paper, which lead to the $(\psi_i)$ being all in $\Psi$.
Note that in many cases, $\Psi$ is dense in $L^2(\mu_\Xi)$ \citep{Micchelli2006}, and one does not need to take such a precaution since the $\ker K = \brace{0}$, and there is only one way to define $K^{-1}$ on $L^2(\mu_\Xi)$.

\subsection{Second proof of Lemma \ref{lem:close_rkhs} with covariance operators}
The proof given above of Lemma \ref{lem:close_rkhs} might seem quite abstract for the reader unfamiliar with reproducing kernel Hilbert space.
In this subsection, we provide a somewhat more accessible proof of this Lemma based on covariance operators.

Reusing the unbiased characterization of ${\cal L}$ we have
\begin{align*}
    {\cal L}(\psi; \beta) 
    &= 2(\beta - 1)\E_{\xi}[\norm{\psi(\xi)}^2] 
    - 2\beta\E_X\E_{\xi, \xi'}\bracket{\psi(\xi)^\top\psi(\xi')\midvert X} 
    + \E_{\xi, \xi'}\bracket{(\psi(\xi')^\top\psi(\xi))^2} + k.
    \\&= 2(\beta - 1)\trace\paren{\E_{\xi}[\psi(\xi)\psi(\xi)^\top]}
    - 2\beta\trace\paren{\E_X\E_{\xi, \xi'}\bracket{\psi(\xi))\psi(\xi')^\top\midvert X}}
    + \trace\paren{\E_{\xi}\bracket{\psi(\xi)\psi(\xi)^\top}^2} + k,
\end{align*}
where the last term provides from the fact that
\begin{align*}
    \E_{\xi, \xi'}\bracket{(\psi(\xi)^\top\psi(\xi'))^2} 
    &= \E_{\xi, \xi'}\bracket{\psi(\xi)^\top\psi(\xi')\psi(\xi')^\top\psi(\xi)}
    = \E_{\xi, \xi'}\bracket{\trace\paren{\psi(\xi')\psi(\xi')^\top\psi(\xi)\psi(\xi)^\top}}
    \\&= \trace\paren{\E_{\xi}[\psi(\xi)\psi(\xi)^\top]\E_{\xi'}[\psi(\xi')\psi(\xi')^\top]}
    = \trace\paren{\E_{\xi}[\psi(\xi)\psi(\xi)^\top]^2}.
\end{align*}

\subsubsection{Operator technicalities}
The search for $\psi$ will be done under the form $\Theta\phi$ for $\Theta\in\R^k\otimes{\cal H}$ and $\phi:\X\to{\cal H}$.
Let us discuss technicalities related to the infinite dimensional operators that will appear.
\begin{assumption}
    \label{ass:techn_H}
    The Hilbert space ${\cal H}$ is separable, and the mapping $\phi$ belongs to $L^2(\mu_\X)$ endowed with Borel topology on both $\X$ and ${\cal H}$.
\end{assumption}

\begin{remark}
    The operator $\Sigma = \E_\xi[\phi(\xi)\phi(\xi)^\top] \in {\cal H}\otimes{\cal H}$ is trace-class.
\end{remark}
\begin{proof}
    This follows from linearity of traces, expectations, together with the fact that $\trace{AB} = \trace{BA}$,
    \[
        \trace\Sigma = \E_\xi \trace{\phi(\xi)\phi(\xi)^\top} = \norm{\phi}_{L^2(\mu_\X)}^2 < +\infty.
    \]
    As a consequence, $\Sigma$ is compact, hence has a pure point spectrum, and since ${\cal H}$ is separable it can be diagonalized with its eigenvectors forming a basis of ${\cal H}$.
\end{proof}

We will see later that $\Sigma^{-1/2}\Sigma_X\Sigma^{-1/2}$ is indeed isometric to $T$.
Hence, under Assumption \ref{ass:diag}, $\Sigma^{-1/2}\Sigma_\X\Sigma^{-1/2}$ has a pure-point spectrum.
However, the following lemma shows that this operator is bounded without using the fact that $T \preceq I$.

\begin{remark}
    The operator $\Sigma_X  = \E_X\E_{\xi,\xi'}\bracket{\phi(\xi)\phi(\xi')^\top\midvert X} \in {\cal H}\otimes{\cal H}$ verifies $0 \preceq \Sigma_X  \preceq \Sigma$ with $\preceq$ the Loewner order ($A\preceq B$ if $B - A$ is semi-definite positive).
    As a consequence, $\Sigma_X $ is trace-class and $\Sigma^{-1/2}\Sigma_X \Sigma^{-1/2}$ is continuous.
\end{remark}
\begin{proof}
    This follows from Jensen inequality applies to $A\to AA^\top$, which can be proven using the positivity of covariance operator.
    \[
        0\preceq \E[(A - \E[A])(A - \E[A])^\top],
        \qquad\Rightarrow\qquad
        \E[A]\E[A]^\top\preceq\E[AA^\top].
    \]
    As a consequence,
    \begin{align*}
        \E_{\xi,\xi'}\bracket{\phi(\xi)\phi(\xi')^\top\midvert X=x} \preceq \E_\xi\bracket{\phi(\xi)\phi(\xi)^\top \midvert X=x},
    \end{align*}
    which implies that
    \[
        \Sigma_X  \preceq \Sigma.
    \]
    As a consequence, $\trace{\Sigma_X } \leq \trace\Sigma < +\infty$ and $\Sigma^{-1/2}\Sigma_X \Sigma^{-1/2} \preceq I$, and $\norm{\Sigma^{-1/2}\Sigma_X \Sigma^{-1/2}}_{\op} \leq 1$.
    The positivity follows from the fact that $\Sigma_X $ is a covariance operator
    \(
        \Sigma_X  = \E_X\bracket{\E_\xi\bracket{\phi(\xi)\midvert X}\E_\xi\bracket{\phi(\xi)\midvert X}^\top}.
    \)
\end{proof}

\subsubsection{Operator formulation}

Let us begin by proving a variant of the lemma where everything is expressed in ${\cal H}$.
We expand later on the isometry between ${\cal H}$ and $L^2(\mu_\Xi)$ (due to the isometry between $S$ and $\Sigma^{1/2}$) that allows us to transfer it to the lemma written in the paper.

\begin{lemma} \label{lem:optimal_in_RKHS}
    For $(\theta_i)\in {\cal H}^k$ and $f_\theta:x\to \scap{\phi(x)}{\theta}$, and a regularizer $\lambda \in \R$
    \[
        {\cal L}((f_{\theta_i})_{i\in[k]}) + \lambda \sum_{i\in[k]}\norm{\theta_i}^2_2 = \trace\paren{\Big(\Sigma^{1/2}(\sum_{i\in[k]}\theta_i\theta_i^\top)\Sigma^{1/2} - A)^2 - A^2} + k, 
    \]
    with $A$ and $\Sigma$ being operator on ${\cal H}$ defined as
    \begin{align*}
        A = \Sigma^{-1/2}((1-\beta)\Sigma + \beta\Sigma_X - \lambda I)\Sigma^{-1/2},\quad
        \Sigma = \E_{\xi}\bracket{\phi(\xi)\phi(\xi)^\top},\quad
        \Sigma_X = \E_X[\E_{\xi, \xi'}\bracket{\phi(\xi)\phi(\xi')^\top \midvert X}].
    \end{align*}
    As a consequence, a minimizer~$\Theta_*$ of ${\cal L}$ is such that $\Theta_*$ matches the eigenvalue decomposition of $A$ on positive eigenvalues up to the $k$-th.
    Formally, if $A = \sum_{i\in\N} \lambda_i u_i\otimes u_i$ with $u_i\in{\cal H}$ and $(\lambda_i)$ in decreasing order, 
    \begin{equation*}
        \Theta_* = (\theta_i)_{i\in[k]},\,\,\text{with}\,\, \theta_i = \sqrt{\max(\lambda_i, 0)} \Sigma^{-1/2}u_i.
    \end{equation*}
    Moreover, $(f_{\theta_i})$ are orthogonal in $L^2(\mu_{\Xi})$, where $\mu_{\Xi}$ denotes the marginal distribution over augmentations.
\end{lemma}
\begin{proof}
Let us now rewrite the different quantities appearing in ${\cal L}$ based on the parameterization $\psi = \Theta\phi$.
We have
\[
    \trace\paren{\E[\psi(\xi)\psi(\xi)^\top]}
    = \trace\paren{\E[\Theta\phi(\xi)\phi(\xi)^\top\Theta^\top ]}
    = \trace\paren{\Theta\E[\phi(\xi)\phi(\xi)^\top]\Theta^\top }
    = \trace\paren{\Theta\Sigma\Theta^\top }
    = \trace\paren{\Sigma^{1/2}\Theta^\top \Theta\Sigma^{1/2}}.
\]
The adjoint $\Theta^\top $ is taken with respect to the canonical topology on ${\cal H}$ and $\R^k$. 
Similarly,
\[
    \trace\paren{\E_X\E\bracket{\psi(\xi)\psi(\xi')^\top\midvert X}}
    = \trace\paren{\Theta\Sigma_X \Theta^\top }
    = \trace\paren{\Sigma^{-1/2}\Sigma_X \Sigma^{-1/2}\Sigma^{1/2}\Theta^\top \Theta\Sigma^{1/2}}.
\]
For the last term, we get
\[
    \trace\paren{\E[\psi(\xi)\psi(\xi)^\top]^2}
    = \trace\paren{(\Theta \Sigma\Theta^\top )^2}
    = \trace\paren{\Theta \Sigma\Theta^\top \Theta \Sigma \Theta^\top }
    = \trace\paren{\Sigma^{1/2} \Theta^\top  \Theta \Sigma \Theta^\top \Theta \Sigma^{1/2}}
\]
Collecting the different terms, we get
\begin{align*}
    &{\cal L}(\Theta\phi) + 2\lambda\trace(\Theta^\top \Theta)- k
    \\&= 
  \trace\big(2(\beta - 1) \Sigma^{1/2}\Theta^\top \Theta\Sigma^{1/2} - 2\beta\Sigma^{-1/2}\Sigma_X \Sigma^{-1/2}\Sigma^{1/2}\Theta^\top \Theta\Sigma^{1/2} \\&\qquad\qquad\qquad\qquad+ \Sigma^{1/2}\Theta^\top \Theta\Sigma\Theta^\top \Theta\Sigma^{1/2} + 2\lambda\Sigma^{-1}\Sigma^{1/2}\Theta^\top \Theta\Sigma^{1/2}\big)
    \\&= 
    \trace\paren{\paren{\Sigma^{1/2}\Theta^\top \Theta\Sigma^{1/2} + (\beta-1)I - \beta\Sigma^{-1/2}\Sigma_X \Sigma^{-1/2} + \lambda\Sigma^{-1}}^2 - \paren{(\beta-1)I - \beta\Sigma^{-1/2}\Sigma_X \Sigma^{-1/2} + \lambda\Sigma^{-1}}^2}
    \\&= 
    \trace\paren{\paren{\Sigma^{1/2}\Theta^\top \Theta\Sigma^{1/2} - \Sigma^{-1/2}((1-\beta)\Sigma + \beta \Sigma_X  - \lambda)\Sigma^{-1/2}}^2 - \paren{\Sigma^{-1/2}((1-\beta)\Sigma + \beta \Sigma_X  - \lambda)\Sigma^{-1/2}}^2}.
\end{align*}
This proves the first part of the lemma.
Remark that the expression of the lemma is slightly different from the generalization to continuous $\X$ suggested by \citet{haochen_provable_2021} in their Appendix F, that would reuse the work of \citet{Schiebinger2015} considering the covariance operator with feature $\bar\phi(x) = q^{-1/2}(x)\E\bracket{\phi(\xi)\midvert X=x}$ where $q:x\to\E_{X\sim\mu_\Xi}[k(x, X)]$ rather than $\Sigma^{-1/2}\Sigma_X\Sigma^{-1/2}$.

Finally, let us prove that the $f_{\theta_i}$ are orthogonal in $L^2$, we have
\begin{align*}
    \scap{f_{\theta_i}}{f_{\theta_j}}_{L^2(\mu_\Xi)}
    &= \sqrt{\max(\lambda_i, 0)\max(\lambda_j, 0)} \E[\scap{\Sigma^{-1/2}u_i}{\phi(\xi)}\scap{\Sigma^{-1/2}u_i}{\phi(\xi)}]
    \\&= \sqrt{\max(\lambda_i, 0)\max(\lambda_j, 0)} \E[u_i^\top\Sigma^{-1/2}\phi(\xi)\phi(\xi)^\top\Sigma^{-1/2}u_j]
    \\&= \sqrt{\max(\lambda_i, 0)\max(\lambda_j, 0)} u_i^\top\Sigma^{-1/2}\E[\phi(\xi)\phi(\xi)^\top]\Sigma^{-1/2}u_j
    \\&= \sqrt{\max(\lambda_i, 0)\max(\lambda_j, 0)} u_i^\top\Sigma^{-1/2}\Sigma\Sigma^{-1/2}u_j
    \\&= \sqrt{\max(\lambda_i, 0)\max(\lambda_j, 0)} u_i^\top u_j
    = \sqrt{\max(\lambda_i, 0)\max(\lambda_j, 0)} \delta_{ij}.
\end{align*}
This proves the orthogonality of the $f_{\theta_i}$ in $L^2(\mu_\Xi)$.
\end{proof}

\subsubsection{Isometric formulation}
Let us consider $S:{\cal H}\to L^2(\mu_\Xi); \theta \to f_\theta$ the embedding of the RKHS in $L^2(\mu_\Xi)$.

\begin{lemma}
    \label{lem:iso}
    $S$ is isometric to $\Sigma^{1/2}$, and $K = S^\top S$ is an integral operator that maps $f\in L^2(\mu_\Xi)$ to $Kf \in L^2(\mu_\Xi)$ defined for $\xi\in\X$ as
    \begin{equation}
        Kf(\xi) = \E_{\xi'}\bracket{\phi(\xi)^\top\phi(\xi')f(\xi')}.
    \end{equation}
\end{lemma}
\begin{proof}
    This follows from the fact that both $S$ and $\Sigma^{1/2}$ are a square root of $\Sigma$.
    Indeed, $\Sigma = S^\top S$, since for $\theta\in{\cal H}$,
    \begin{align*}
        \scap{\theta}{S^\top S\theta}_{\cal H}
        &= \scap{S\theta}{S\theta}_{L^2(\mu_\X)}
        = \E_\xi[S\theta(\xi)^2]
        \\&= \E_\xi[\scap{\theta}{\phi(\xi)}^2]
        = \E_\xi[\scap{\theta}{\phi(\xi)\otimes\phi(\xi)\theta}]
        \\&= \scap{\theta}{\E[\phi(\xi)\otimes\phi(\xi)]\theta}
        = \scap{\theta}{\Sigma\theta}.
    \end{align*}
    As a consequence, $S$ is isometric to $\Sigma^{1/2}$ (if we write the singular value decomposition of $S$ as $USV^\top$, then $\Sigma^{1/2} = USU^\top$).
    Regarding the part in $K$, one can check with the same derivation that $S^\top f = \E[f(\xi)\phi(\xi)] \in {\cal H}$ hence the value of $(Kf)(\xi)= (S^\top f)^\top \phi(\xi)= \E_{\xi'}[f(\xi')\phi(\xi')^\top \phi(\xi)]$.
\end{proof}

Using the isometry one can replace $\norm{S\theta} = \norm{\Sigma^{1/2}\theta}$ with the Hilbertian norm on ${\cal H}$ and $L^2(\mu_\Xi)$, so that for $C$ operating in ${\cal H}$, $\trace{SCS^\top} = \trace{\Sigma^{1/2}C\Sigma^{1/2}}$.
Going back to the proof in ${\cal H}$, one can replace all the $\Sigma^{1/2}$ by $S$ or its adjoint at the right places to get the following statement.

\begin{lemma}
    \label{lem:charac}
    For $\Theta\in\R^k\otimes{\cal H}$, and a regularized $\lambda \in \R$
    \[
        {\cal L}(S\Theta) + \lambda\norm{\Theta}^2_2 = \trace((S\Theta^\top \Theta S^\top-T_\lambda)^2 - T_\lambda^2) + k
    \] 
    where
    \[
        T = S^{-\top}\Sigma_X S^{-1}, \qquad T_\lambda = (1-\beta)I + \beta T - \lambda K^{-1}, \qquad K = SS^\top,
    \]
    with $S:{\cal H}\to L^2(\mu_\Xi); \theta \to f_\theta$ the embedding of ${\cal H}$ in $L^2(\mu_\Xi)$, where $\mu_{\Xi}$ denotes the marginal distribution over augmentations.
    As a consequence, a minimizer~$\Theta_*$ of ${\cal L;\lambda}$ is such that $S\Theta_*^\top$ matches the eigenvalue decomposition of $T_\lambda$ on positive eigenvalues up to the $k$-th.
\end{lemma}
\begin{proof}
    This lemma follows from the previous discussion. The fact that $S^{-\top}\Sigma S^{-1}$ equates to $T$ on the $L^2(\mu_\Xi)$-closure of $\Psi$ is due to the characterization in Lemma \ref{lem:close_rkhs}.
	We can nonetheless prove it in a more direct fashion, by adapting Lemma B.9 of \citet{saunshi_understanding_2022} to our case.
\end{proof}

\subsection{Proof of Proposition \ref{prop:commute}}

Proposition \ref{prop:commute} relies on the fact that when two operators commutes, they can be diagonalized in the same basis.

\begin{lemma}
When $K$ and $T$ commute, $K$ and $T$ can be diagonalized by the same eigenfunctions $(f_i)$.
\end{lemma}
\begin{proof}
    When the operators commute, if $f$ is an eigenfunction of $T$ with $Tf = \lambda f$, then $TKf = KTf = \lambda Kf$.
    This means that the eigenspace of $T$, i.e. $\ker(T-\lambda I)$ are stable by $K$.
    As a consequence, $K$ can be decomposed with respect to the summation $L^2 = \oplus_{\lambda\in\spec(T)} \ker(T-\lambda I)$.
    By diagonalizing the restrictions of $K$ on each of those spaces, there exists a basis that diagonalizes both $K$ and $T$.
\end{proof}

While we did not discuss it in the main text, one should not consider any eigenvalue decomposition of $T$ but only the eigenfunctions that jointly diagonalize $T$ and $K$.
However, note that to find those eigenfunctions, based on Courant-Fisher principle, one can take, recursively on $i\in\N$, $f_i = f_{\theta_i}$ an eigenfunction in $\ker(T-\lambda_i I)$ that maximizes or minimizes $\norm{\theta_i}$.
Those eigenfunctions $(f_i)$ will diagonalize $T_\lambda$, and the optimal representation will pick the ones that maximize $f_i^\top T_\lambda f_i$ as long as this quantity is positive.

If $f_i$ diagonalize $K$ then $f_i \in \ima K^{1/2} = \Psi = \ima S$, hence there exists a $\theta_i\in{\cal H}$ such that $f_i = S\theta_i$.
As a consequence, with the $L^2(\mu_\Xi)$ geometry, $f_i^\top K^{-1} f_i = (S\theta_i)^\top (S^\top S)^{-1} S\theta_i = \norm{\theta_i}^2_2$.
We use this to derive that
\begin{align*}
    f_i T_\lambda f_i
    &= (1-\beta)f_i^\top f_i + \beta f_i^\top T f_i - \lambda f_i^\top K^{-1} f_i
    = 1-\beta + \beta\lambda_i - \lambda \norm{\theta_i}^2_2.
\end{align*}
In other term the maximal eigenvalues of $T_\lambda$ are found by maximizing $\beta\lambda_i - \lambda\norm{\theta_i}^2$.

\begin{remark}
\label{rmk:haochen}
Recently, \citet{haochen2022theoretical} have taken this second perspective on inductive bias perspective by looking at the ``barrier'' case where one can only match eigenfunctions that belongs to the function space $\Psi$. 
In the kernel regime, this is deceptive since, for example, when considering the Gaussian kernel $\phi(x)^\top \phi(x') = -\exp(\norm{x-x'}^2)$, $\Psi$ is made of analytic functions \citep{Sun2008}, hence cannot parameterize any indicator functions without being one everywhere, therefore their approach would fail to explain how the Gaussian kernel could learn fast under the cluster assumption.
\end{remark}

\subsection{Remark about VCReg}
When ${\cal L}=0$, finding $\psi$ correspond in finding $k$ functions $(f_{\theta_i})_i$ that are orthogonal in $L^2(\mu_\Xi)$ and maximize $1-\lambda\norm{\theta}^2 = 1-\lambda f_\theta^\top K^{-1} f_\theta$ before multiplying them by $(1-\lambda\norm{\theta_i}^2)_+$.
Using Courant-Fisher min-max principle, the function $(f_{\theta_i})_i$ are given by the $k$ biggest eigenfunctions of $K$.

\subsection{Proof of Example \ref{ex:interpolation}}
If $\mu_\Xi = \delta \rho_\X + (1- \delta)\mu_\perp$, then for any measurable function $f$
\[
    \norm{f}^2_{L^2(\mu_\Xi)}
    = \delta \int_\X f(x)^2 \rho_\X(\diff x) 
    + (1-\delta) \int_\X f(x)^2 \mu_\perp(\diff x) 
    \geq \delta \norm{f(X)}_{L^2(\rho_\X)}^2.
\]

\subsection{Proof of Example \ref{ex:inter_cov}}
This follows from the embedding $S_\rho$ and $S_\mu$ of ${\cal H}$ in $L^2(\rho_\X)$ and $L^2(\mu_\Xi)$ respectively.
We have seen earlier that $S_\mu^\top S_\mu = \Sigma_{\mu_\Xi}$ and $S_X^\top S_X = \Sigma_{\rho_\X}$.
Let $f\in\Psi$, there exists $\theta\in{\cal H}$ such that $f = f_\theta$, hence, using the isometry between $S$ and $\Sigma^{1/2}$,
\begin{align*}
    \norm{f_\theta}_{L^2(\rho_\X)}^2
    &= \norm{S_X\theta}_{L^2(\rho_\X)}^2
    = \norm{\Sigma_{\rho_\X}^{1/2}\theta}_{\cal H}^2
    = \norm{\Sigma_{\rho_\X}^{1/2}\Sigma_{\mu_\Xi}^{-1/2} \Sigma_{\mu_\Xi}^{1/2}\theta}_{L^2(\rho_\X)}^2
    \\&\leq \norm{\Sigma_{\mu_\Xi}^{-1/2}\Sigma_{\rho_\X}\Sigma_{\mu_\Xi}^{-1/2}}_{\op} \norm{\Sigma_{\mu_\Xi}^{-1/2}\theta}_{\cal H}^2
    = \norm{\Sigma_{\mu_\Xi}^{-1/2}\Sigma_{\rho_\X}\Sigma_{\mu_\Xi}^{-1/2}}_{\op} \norm{f_\theta}_{L^2(\mu_\Xi)}^2.
\end{align*}
We conclude by using the equivalence the fact that $A \preceq cB$ implies that $B^{-1/2} A B^{-1/2}\preceq c\cdot I$.

\subsection{Proof of Example \ref{ex:invariant}}
This follows from the definition of the different objects, 
\[
    \Pi_{f_i}^{(\rho_\X)}(f) = w f_i,
    \qquad\text{with}\qquad
    w = \argmin_{w\in\R} \E_{X\sim\rho_\X}[\norm{f(X) - w f_i(X)}^2].
\]
We develop this last objective as
\begin{align*}
    \E_{X\sim\rho_\X}[\norm{f(X) - wf_i(X)}^2]
    &= \E_{X\sim\rho_\X}[\norm{g(\psi(X)) - wf_i(X)}^2]
    = \E_{Z\sim\psi_\#\rho_\X}[\norm{g(Z) - w^\top Z}^2]
    \\&= \E_{Z\sim\mu_\Xi}[\norm{f(X) - w^\top f_i(X)}^2].
\end{align*}
Hence, the equality of the $w$ and of the projections.

\subsection{Proof of Example \ref{ex:cluster}}
If $\mu_\Xi$ has $k$ connected components, then the indicators of those components will be orthogonal in $L^2(\mu_\Xi)$ while minimizing the invariant term $\E_x\E\bracket{\norm{\phi(\xi) - \phi(\xi')}^2\midvert X}$.
As a consequence, $f^*$ belongs to the space of the $(f_i)_{i\leq k}$.

%% file: appendix/downstream.tex
\section{Control of the downstream convergence}
\label{app:theorem}

This section is devoted to the proof of Theorem \ref{thm:conv_simple}.
In all the following, $k_\lambda$ designs the number of positive eigenvalues of $T_\lambda$ (including multiplicity) as an operator on $L^2(\mu_\Xi)$.
We fix $k \leq k_\lambda$, and design by ${\cal F}$ the span of the $(f_i)_{i\in[k]}$.
In the kernel regime, the space ${\cal F}$ can also be written as ${\cal F} = \brace{w^\top \Theta_* \phi\midvert w\in\R^k}$ for $\Theta_*$ the minimizer defined in Lemma \ref{lem:charac}.
We denote by $\hat{\cal F}$ the space defined similarly from an estimate $\hat\Theta$ of $\Theta_*$.

The error on the downstream task could be decomposed into three quantities: the error on the downstream task linked with the capacity of $\hat{\cal F}$ \eqref{eq:capac}; the error on the upstream task linked to approximation error between ${\cal F}$ and $\hat{\cal F}$ \eqref{eq:opera},
the error due to the fact that the downstream task might not be effectively solved within ${\cal F}$ \eqref{eq:source}.

\begin{lemma}[Decomposition intuition]
    \label{lem:decomp}
    Let ${\cal F}$ and $\hat{\cal F}$ be two closed convex sets of $L^2(\rho_\X)$, and $\Pi_{\cal F}$ design the orthogonal projection on the space ${\cal F}$ according to $L^2(\rho_\X)$ geometry.
    For any function $f:\X\to\Y$ in $\hat{\cal F}$, the excess of risk \eqref{eq:down_obj} can be decomposed as
    \begin{align}
        \label{eq:capac}
        {\cal R}(f) - {\cal R}(f^*) 
        &\leq \norm{f - \Pi_{\hat{\cal F}}f^*}^2_{L^2(\rho_\X)} 
        \\&+ 2\norm{(I - \Pi_{\hat{\cal F}})\Pi_{\cal F} f^*}^2_{L^2(\rho_\X)}
        \label{eq:opera}
        \\&+ \norm{(I - \Pi_{\cal F}) f^*}^2_{L^2(\rho_\X)},
        \label{eq:source}
    \end{align}
\end{lemma}
\begin{proof}
    The proof of the lemma follows from classical characterization of the mean square error and a triangular inequality.
    Introduce the following technical assumption.
    \begin{assumption}
    Assume $(X, Y) \to Y$ to belong to $L^2(\rho)$.
    \end{assumption}
    When $\ell(y, y') = \norm{y - y'}^2$, using the fact that $(X, Y) \to Y - \E\bracket{Y\midvert X}$ is orthogonal to any measurable function that does not depend on $Y$ in $L^2(\rho)$,
    \[
        {\cal R}(f) = \E[\norm{f(X) - Y}^2] = 
        \E[\norm{f(X) - \E\bracket{Y\midvert X}}^2]
        + \E[\norm{\E\bracket{Y\midvert X} - Y}^2].
    \]
    As a consequence, $f^*(x) = \E\bracket{Y\midvert X=x}$ and 
    \[
        {\cal R}(f) - {\cal R}(f^*) = \norm{f - f^*}^2_{L^2(\rho_\X)}.
    \]
    Let us decompose the excess of risk with the orthogonal projection of $\hat{\cal F}$, we have
    \begin{align*}
        {\cal R}(f) - {\cal R}(f^*)
        &= \norm{f - f^*}^2_{L^2(\rho_\X}
        = \norm{f - \Pi_{\hat{\cal F}}f^*}^2_{L^2(\rho_\X}
        +\norm{(I - \Pi_{\hat{\cal F}})f^*}^2_{L^2(\rho_\X}
    \end{align*}
    The second term is worked out as
    \begin{align*}
        \norm{(I - \Pi_{\hat{\cal F}})f^*}^2_{L^2(\rho_\X)}
        & =\norm{(I - \Pi_{\hat{\cal F}})\Pi_{\cal F}f^* + (I - \Pi_{\hat{\cal F}})(I-\Pi_{\cal F})f^*}^2_{L^2(\rho_\X)}
        \\& \leq 2\norm{(I - \Pi_{\hat{\cal F}})\Pi_{\cal F}f^*}^2_{L^2(\rho_\X)} + \norm{(I - \Pi_{\hat{\cal F}})(I-\Pi_{\cal F})f^*}^2_{L^2(\rho_\X)}
        \\& \leq 2\norm{(I - \Pi_{\hat{\cal F}})\Pi_{\cal F}f^*}^2_{L^2(\rho_\X)} + \norm{(I-\Pi_{\cal F})f^*}^2_{L^2(\rho_\X)}
    \end{align*}
    where the last inequality is due to the fact that projections contract distances.
\end{proof}

For linear probing \eqref{eq:lin_class}, when the downstream task is learned with $n$ data points and a noise level $\epsilon$, \eqref{eq:capac} is expected to behave as $\epsilon^2 k / n$ \citep{mourtada2022exact}.
In this linear setting, \eqref{eq:opera} should be seen as a measure of angle between ${\cal F}$ and $\hat{\cal F}$ seen through the eyes of $f^*$ \citep{davis_rotation_1970,kato_perturbation_1995}.

\subsection{Controlling \texorpdfstring{\eqref{eq:capac}}{Capacity}}

The downstream task error relates to the generalization error of mis-specified linear model.
To bound it, we will use the convergence rates analysis through concentration of integral operators of \citet{smale_learning_2007} and \citet{caponnetto_optimal_2007}.
It requires reworking slightly the previous decomposition.

\begin{lemma}[Warm-up]
    Let $\hat{\cal F}$ be the span of the $(\psi_i)_{i\in[k]}$, with $S_\psi:\R^k \to L^2$ defined as $S_\psi w = w^\top \psi$, then
    \begin{equation}
        \Pi_{\hat{\cal F}}f^* = S_\psi \E[\psi(X)\psi(X)^\top]^{-1} \E[Y\psi(X)].
    \end{equation}
    Based on data $(X_i, Y_i)$, one can define the empirical risk minimizer $f_n= S_\psi w_n$, where $w_n$ is the minimizer of
    \begin{equation}
        w_n \in \argmin_{w\in\R^k} \sum_{i=1}^n \norm{w^\top \psi(X_i) - Y_i}^2
        = [\frac{1}{n}\sum_{i=1}^n\phi(X_i)\phi(X_i)^\top]^{-1} \frac{1}{n} \sum_{j=1}^n Y_i\phi(X_i).
    \end{equation}
\end{lemma}
\begin{proof}
    The two formula can be proven at once by remarking that if $\Pi_{\hat{\cal F}}f^*$ is defined as $S_\psi w$ for $w$ minimizing
    \[
        \E[\norm{w^\top \phi(X) - Y}^2]
        = w^\top \E[\phi(X)\phi(X)^\top]w - 2w^\top \E[Y\phi(X)] + \E[\norm{Y}^2].
    \]
    Minimizing this quadratic form leads to the first results.
    The second result is proven in the same way after substituting the distribution over $(X, Y)$ by the empirical one $n^{-1} \sum_{i\in[n]} \delta_{(X_i, Y_i)}$.
\end{proof}

As a consequence of this warm-up lemma, let us introduce some notations, for $\psi:\X\to\R^k$ and some data $(X_i)$, define
\begin{equation}
    S_\psi:\R^k \to L^2(\rho_\X); w\to w^\top \psi,
    \qquad
    \hat S_\psi:\R^k \to \ell^2(n); w \to (w^\top \psi(X_i))_{i\in[n]},
\end{equation}
where $\ell^2(n)$ is endowed with normalized (i.e. probability-like) scalar product $\scap{a}{b} = n^{-1} \sum_{i\in[n]} a_ib_i$.
Similarly to Lemma~\ref{lem:iso}, one can show that the adjoint of $S_\psi$ and $\hat S_\psi$, and the covariance operators are
\begin{align}
    \nonumber
    &S_\psi:L^2(\rho_\X)\to \R^k; f \to \E_{\rho_\X}[f(X)\psi(X)],
    \qquad
    \hat S_\psi:\ell^2(n)\to \R^k; (Y_i)_{i\in n} \to \frac{1}{n}\sum_{i\in[n]} Y_i\psi(X_i).
    \\&\Sigma_\psi = S_\psi S_\psi^\top = \E_{\rho_\X}[\psi(X)\psi(X)^\top],
    \qquad
    \hat \Sigma_\psi = \hat S_\psi \hat S_\psi^\top = \frac{1}{n}[\psi(X)\psi(X)^\top],
\end{align}
In this subsection, we will only consider $S$ and $\Sigma$ associated with $\psi$ and we remove the indices for convenience.
To simplify notation when $f\in L^2(\rho_\X)$ we will write $\hat S^\top f$ for $\hat S^\top (f(X_i))_{i\in[n]}$.

\begin{assumption}[Homoskedastic noise]
    \label{ass:noise}
    There exists $\epsilon > 0$ such that for $\rho_\X$-almost all $x$, the variance of $\paren{Y\midvert X=x}$ is bounded by $\epsilon^2$.
\end{assumption}

\begin{lemma}[Bias-Variance decomposition]
    Based on data $(X_i, Y_i)$, one can define the regularized empirical risk minimizer $f_n= S_\psi w_n$ with a regularization parameter $\gamma > 0$ as 
    \begin{equation}
        \label{eq:erm}
        w_n \in \argmin_{w\in\R^k} \sum_{i=1}^n \norm{w^\top \psi(X_i) - Y_i}^2 + \gamma \norm{w}^2.
    \end{equation}
    When doing so, under Assumption \ref{ass:noise}, the average excess of risk can be decomposed as, with $M = \sup \norm{\psi(X)}$,
    \begin{align}
        \E_{(X_i, Y_i)}[\norm{f_n- \Pi_{\hat{\cal F}}f^*}^2_{L^2(\rho_\X)}]
        \nonumber
        &\leq  \frac{\epsilon^2}{n} \paren{1 + \frac{M^2}{\gamma n}} \trace\paren{(\Sigma+ \gamma)^{-1}\Sigma}
        \nonumber
        \\&+ 2\gamma\paren{1 + \frac{M^2}{\gamma n}}^2\scap{\Pi_{\hat{\cal F}}f^*}{\Sigma(\Sigma + \gamma)^{-1}\Pi_{\hat{\cal F}}f^*}_{L^2(\rho_\X)}
        \nonumber
        \\&+ 2\E_{(X_i)}\norm{S(\hat\Sigma + \gamma)^{-1}\hat S^\top (I - \Pi_{\hat{\cal F}}) f^*}^2_{L^2(\rho_\X)}.
        \label{eq:to_work_out}
    \end{align}
\end{lemma}
\begin{proof}
    Retaking the warm-up lemma, one can show that
    \[
        w_n = (\hat\Sigma + \gamma)^{-1}\hat S (Y_i)_{i\in[n]}.
    \]
    As a consequence, using the usual bias-variance decomposition, and the fact that $f^* = \E_\rho\bracket{Y\midvert X=\cdot}$, we develop
    \begin{align*}
        &\E_{\paren{Y_i\midvert X=X_i}}[\norm{f_n - \Pi_{\hat{\cal F}}f^*}^2]
        = \E_{\paren{Y_i\midvert X=X_i}}[\norm{S(\hat\Sigma + \gamma)^{-1}\hat S^\top (Y_i)_{i\in[n]} - \Pi_{\hat{\cal F}} f^*}^2]
        \\&= \E_{\paren{Y_i\midvert X=X_i}}\bracket{\norm{S(\hat\Sigma + \gamma)^{-1}\hat S^\top (Y_i - \E\bracket{Y\midvert X=X_i})_{i\in[n]}}^2}
        + \norm{S(\hat\Sigma + \gamma)^{-1}\hat S^\top f^* - \Pi_{\hat{\cal F}}f^*}^2.
    \end{align*}
    The first term can be worked out with \citet{mourtada2022exact} techniques as
    \[
        \E_{(X_i, Y_i)}\bracket{\norm{S(\hat\Sigma + \gamma)^{-1}\hat S^\top (Y_i - \E\bracket{Y\midvert X=X_i})_{i\in[n]}}^2}
        \leq \frac{\epsilon^2}{n} \paren{1 + \frac{R^2}{\gamma n}} \trace\paren{(\Sigma+ \gamma)^{-1}\Sigma}
    \]
    under the assumption that the variance of $\paren{Y\midvert X}$ is bounded by $\epsilon^2$. 

    We work out the second term with
    \[
        \norm{S(\hat\Sigma + \gamma)^{-1}\hat S^\top f^* - \Pi_{\hat{\cal F}}f^*}
        \leq
        \norm{S(\hat\Sigma + \gamma)^{-1}\hat S^\top (f^* - \Pi_{\hat{\cal F}} f^*)}
        + \norm{S(\hat\Sigma + \gamma)^{-1}\hat S^\top \Pi_{\hat{\cal F}}f^* - \Pi_{\hat{\cal F}}f^*}.
    \]
    Once again, the last part can be worked out with techniques of \citet{mourtada2022exact} to get
    \[
        \E_{(X_i, Y_i)}[\norm{S(\hat\Sigma + \gamma)^{-1}\hat S^\top \Pi_{\hat{\cal F}}f^* - \Pi_{\hat{\cal F}}f^*}^2] \leq 
        \gamma\paren{1 + \frac{R^2}{\gamma n}}^2\scap{\Pi_{\hat{\cal F}}f^*}{\Sigma(\Sigma + \gamma)^{-1}\Pi_{\hat{\cal F}}f^*}_{L^2(\rho_\X)}.
    \]
    This provides the decomposition of the lemma.
\end{proof}

Let us work out the last term in \eqref{eq:to_work_out}.
\begin{lemma}
    For $t = \norm{(\Sigma+\gamma)^{-1/2}(\Sigma - \hat \Sigma) (\Sigma+\gamma)}_{\op}$ and $M$ such that $\norm{\psi(X)}\leq M$ almost everywhere, 
    \begin{equation}
        \norm{S(\hat\Sigma + \gamma)^{-1}\hat S^\top (I - \Pi_{\hat{\cal F}}) f^*}_{L^2(\rho_\X)} \leq \min\brace{\frac{1}{1-t}, 1 + t\cdot\frac{M^2 + \gamma}{\gamma}}\norm{\Sigma_\gamma^{-1/2} \hat S^\top (I - \Pi_{\hat{\cal F}})f^*}.
    \end{equation}
\end{lemma}
\begin{proof}
    Let us set $f = (I - \Pi_{\hat{\cal F}})f^*$ and $A_\gamma = A + \gamma I$ for simplicity.
    Remark that $f$ is orthogonal to the image of $S$, hence $S^\top f = 0$.
    We decompose the last quantity with
    \begin{align*}
        \hat\Sigma_\gamma^{-1}\hat S^\top f 
        &=(\hat\Sigma_\gamma^{-1}\hat - \Sigma_\gamma^{-1}) \hat S^\top f + (\Sigma_\gamma)^{-1} \hat S^\top f
        \\&=\hat\Sigma_\gamma^{-1}\hat (\Sigma_\gamma - \hat\Sigma_\gamma) \Sigma_\gamma^{-1} S^\top f + \Sigma_\gamma^{-1} \hat S^\top f
        \\&=\hat\Sigma_\gamma^{-1}\hat (\Sigma- \hat\Sigma) \Sigma_\gamma^{-1} \hat S^\top f + \Sigma_\gamma^{-1} \hat S^\top f
        \\&=\Sigma_\gamma^{-1/2}\paren{\Sigma_\gamma^{1/2}\hat\Sigma_\gamma^{-1}\Sigma_\gamma^{1/2}\Sigma_\gamma^{-1/2}\hat (\Sigma- \hat\Sigma) \Sigma_\gamma^{-1/2} + I} \Sigma_\gamma^{-1/2} \hat S^\top f
    \end{align*}
    Using the fact that $S$ is isometric to $\Sigma^{1/2}$ which itself if smaller than $\Sigma_\gamma^{1/2}$ (with the Loewner order), we have
    \[
        \norm{S(\hat\Sigma + \gamma)^{-1}\hat S^\top (I - \Pi_{\hat{\cal F}}) f^*}_{L^2(\rho_\X)} \leq \paren{1 + \norm{\Sigma_\gamma^{1/2}\hat\Sigma_\gamma^{-1}\Sigma_\gamma^{1/2}}_{\op}\norm{\Sigma_\gamma^{-1/2}\hat (\Sigma- \hat\Sigma) \Sigma_\gamma^{-1/2}}_{\op}}\norm{\Sigma_\gamma^{-1/2} \hat S^\top f}
    \]
    We know that 
    \[
        \norm{\Sigma_\gamma^{1/2}\hat\Sigma_\gamma^{-1}\Sigma_\gamma^{1/2}}_{\op} \leq \gamma^{-1}(\norm{\Sigma}_{\op} + \gamma) \leq \gamma^{-1} (\sup_{x\in\supp\rho_\X} \norm{\psi(x)}^2 + \gamma)
    \]
    We also have that for $A$ and $\hat A$ self adjoint and any $t > 0$, the sequence of implications
    \begin{align*}
        \norm{A^{-1/2}(A - \hat A) A^{-1/2}}_{\op} \leq t
        &\quad\Leftrightarrow\quad
        -t I \preceq A^{-1/2}(\hat A - A) A^{-1/2} \preceq t I
        \\&\quad\Leftrightarrow\quad
        -t A \preceq \hat A - A \preceq t A
        \\&\quad\Leftrightarrow\quad
        (1-t) A \preceq \hat A \preceq (1+t) A
        \\&\quad\Leftrightarrow\quad
        (1+t)^{-1} A^{-1} \preceq \hat A^{-1} \preceq (1-t)^{-1} A^{-1}
        \\&\quad\Leftrightarrow\quad
        (1+t)^{-1} \preceq A^{1/2}\hat A^{-1}A^{1/2} \preceq (1-t)^{-1}.
    \end{align*}
    Combining the different results leads to the lemma.
\end{proof}

Probabilistic arguments will show that $t$, as well as $\norm{(\Sigma+\gamma^{-1/2}\hat S^\top (I - \Pi_{\hat{\cal F}})f^*}$, vanishes to zero in $n^{-1/2}$.
We will use Bernstein concentration inequality.

\begin{lemma}[Bernstein concentration inequalities]
    \label{rates:thm:bernstein-vector}
    Let denote by ${\cal A}$ a Hilbert space and by $(Z_i)_{i\in[n]}$ a sequence of independent random vectors on ${\cal A}$ such that $\E[Z_i] = 0$, and such that there exists two positive constants $M$ and $\sigma$ such that for all $m > 2$
    \[
        \frac{1}{n}\sum_{i\in[n]}\E[\norm{Z_i}^m] \leq m! \sigma^2 M^{m-2} / 2
    \]
    For any $t>0$,
    \[
      \Pbb(\big\|\frac{1}{n}\sum_{i=1}^{n} Z_{i}\big\| \geq t) \leq 2\exp\paren{\frac{-nt^2}{2\sigma^2 + 2tM}}.
    \]
    In particular when the $(Z_i)$ are bounded by $3M$, and $\sigma^2 = n^{-1}\sum_{i\in n} \E[Z_i^2]$, the condition holds.
    When, instead, $Z_i$ are symmetric matrices in $\R^{k\times k}$ and $\norm{\cdot}$ is the operator norm, the same bound holds with $k\exp(\cdots)$ instead of $2\exp(\cdots)$ on the right-hand side, where $\sigma^2 = \norm{n^{-1}\sum_{i\in[n]}\E[Z_i^2]}$.
\end{lemma}
\begin{proof}
    See Corollary 1 in \citet{Pinelis1986} for the first part, and \citet{tropp2015} for the matrix version.
\end{proof}

\begin{lemma}
    For any $t > 0$, the vector part in last term of the bias decomposition \eqref{eq:to_work_out} can be controlled with
    \begin{equation}
       \Pbb\paren{\norm{\Sigma_\gamma^{-1/2} \hat S^\top (I - \Pi_{\hat{\cal F}})f^*} \geq t}
       \leq 2\exp\paren{\frac{-nt^2}{a(b + 2M\gamma^{-1/2}t/3)}}
    \end{equation}
    where $b = 2\trace{(\Sigma+\gamma)^{-1}\Sigma}$, $M = \sup \norm{\psi(X)}$ and $a=\norm{f^*}_{L^\infty} + M\norm{f^*}_{L^2}$.
    Moreover, this vector part is bounded by $\gamma^{-1}a^2M^2$.
    The matrix part in the last term of \eqref{eq:to_work_out} is controlled with
    \begin{equation}
       \Pbb\paren{\norm{\Sigma_\gamma^{-1/2}( \hat\Sigma - \Sigma )\Sigma_\gamma^{-1/2}}_{\op} \geq t}
       \leq k\exp\paren{\frac{-nt^2}{2M^2\gamma^{-1}(1 + t/3)}}
    \end{equation}
    Moreover, this matrix part is bounded by $\gamma^{-2} M^4$.
\end{lemma}
\begin{proof}
    Let us introduce
    \begin{equation}
        Z_i = (I-\Pi_{\hat{\cal F}})f^*(X_i) (\Sigma+\gamma)^{-1/2}\psi(X_i) \in \R^k.
    \end{equation}
    One can check that
    \[
        \frac{1}{n} \sum_{i\in[n]} Z_i = (\Sigma+\gamma)^{-1/2}\frac{1}{n}\sum_{i\in[n]} (I-\Pi_{\hat{\cal F}})f^*(X_i) \psi(X_i) = (\Sigma+\gamma)^{-1/2} \hat S^\top (I-\Pi_{\hat{\cal F}})f^*,
    \]
    as well as, since $\ima S = \hat{\cal F}$
    \[
        \E[Z_i] = S^\top (I - \Pi_{\hat{\cal F}})f^* = 0.
    \]
    Moreover,
    \[
        \norm{Z_i} = \norm{(\Sigma+\gamma)^{-1/2}\psi(X_i)} \abs{(I - \Pi_{\hat{\cal F}})f^*(X_i)} \leq \gamma^{-1/2} M (\norm{f^*}_{L^\infty} + M\norm{f^*}_{L^2}). 
    \]
    where $R = \sup_X \norm{\psi(X)}$ and we have used the fact that
    \begin{align*}
        \abs{(I - \Pi_{\hat{\cal F}})f^*(X_i)} 
        &\leq \abs{f^*(X_i)} + \abs{\Pi_{\hat{\cal F}}f^*(X_i)}
        = \abs{f^*(X_i)} + \abs{\scap{SS^{-1}\Pi_{\hat{\cal F}}f^*}{\phi(X)}} 
        \\&\leq \abs{f^*(X_i)} + \norm{SS^{-1}}_{\op}\norm{\Pi_{\hat{\cal F}}f^*}_{L^2}\norm{\phi(X_i)}
        \leq \norm{f^*}_{L^\infty} + M\norm{f^*}_{L^2}.
    \end{align*}
    Finally, we have
    \begin{align*}
        \E[\norm{Z_i}^2] 
        &= \E\bracket{\norm{(\Sigma+\gamma)^{-1/2}\psi(X_i)}^2 \abs{(I - \Pi_{\hat{\cal F}})f^*(X_i)}^2} 
        \\&\leq \E\bracket{\norm{(\Sigma+\gamma)^{-1/2}\psi(X_i)}^2} (\norm{f^*}_{L^\infty} + M\norm{f^*}_{L^2}) 
        \\& = \trace\paren{(\Sigma+\gamma)^{-1}\Sigma} (\norm{f^*}_{L^\infty} + M\norm{f^*}_{L^2}). 
    \end{align*}
    Using Bernstein inequality leads to the control on the vector term.
    
    For the matrix term, let us introduce
    \[
        Z_i = U_i U_i^\top - \E[U_i U_i^\top], \qquad
        U_i = (\Sigma + \gamma)^{-1/2} \phi(X_i).
    \]
    We have $(\Sigma + \gamma)^{-1/2}(\hat\Sigma - \Sigma)(\Sigma + \gamma)^{-1/2} = \frac{1}{n} \sum_{i\in[n]} Z_i$, and
    \[
        \sup\norm{Z} \leq \sup\norm{U}^2 \leq \gamma^{-1} M^2.
    \]
    Finally, using the fact that $\norm{U_i}^2 \preceq U_i \sup\norm{U_i}$, with the variational definition of the mean, with the infimum taken with respect to the Loewner order
    \[
       \E[Z_i^2] = \inf_{a} \E[(Z_i - a)^2]  \preceq \E[(U_i U_i^\top)^2]
       \preceq \sup\norm{U}^2\E[U_i^\top U_i] 
       = \sup\norm{U}^2(\Sigma+\gamma)^{-1}\Sigma
       \preceq \sup\norm{U}^2 I
    \]
    Applying the matrix version of Bernstein inequality leads to the lemma.
\end{proof}

We now turn the deviation inequalities of the last lemma into a bound on the average.
\begin{lemma}
    Retaking the notation of the previous lemma.
    \begin{align*}
        \E_{(X_i)}\norm{S(\hat\Sigma + \gamma)^{-1}\hat S^\top (I - \Pi_{\hat{\cal F}}) f^*}^2_{L^2(\rho_\X)}
        &\leq k\exp\paren{\frac{-3n\gamma}{(3+\sqrt{2})M^2}} (\gamma^{-4}M^6 a^2(M^2 + 2\gamma))^2
        \\&+ \frac{16ab}{n} + \frac{512 a^2M^2}{9\gamma n^{2}}.
    \end{align*}
\end{lemma}
\begin{proof}
    In essence, we have two random variables, $X=\norm{\Sigma_\gamma^{-1/2}( \hat\Sigma - \Sigma )\Sigma_\gamma^{-1/2}}^2_{\op}$, and $Y = \norm{\Sigma_\gamma^{-1/2} \hat S^\top (I - \Pi_{\hat{\cal F}})f^*}^2$ the vector one.
    We proceed with computation using the fact that for $X$ positive $\E[X] = \int_{t>0} \Pbb(X>t)\diff t$ and that $ab > t$, implies for any $s$ that $a > 1+s$ or $b > t/(1+s)$,
    \begin{align*}
        &\E[\min\brace{\frac{1}{1-X}, 1 + \frac{(M^2 + \gamma)X}{\gamma}}^2 Y^2]
        \\&= \int_{t\in(0,\sup (1+\gamma^{-1}(M^2 + \gamma)X)^2Y^2)} \Pbb\paren{\min\brace{\frac{1}{1-X}, 1 + \frac{(M^2 + \gamma)X}{\gamma}}^2 Y^2 > t} \diff t
        \\&\leq \int \inf_{s} \Pbb\paren{\min\brace{\frac{1}{1-X}, 1 + \frac{(M^2 + \gamma)X}{\gamma}}^2 > 1+s} + \Pbb(Y^2 > t/(1+s)) \diff t.
    \end{align*}
    Rather than solving this in closed form, we will proceed with a much simpler bound that consists in taking $s = 1$ without any optimization.
    It gives the much simpler formula
    \[
        \E[\min\brace{\frac{1}{1-X}, 1 + \frac{(M^2 + \gamma)X}{\gamma}}^2 Y^2]
        \leq \Pbb(\frac{1}{(1-X)^2} > 2) \sup(1+\gamma^{-1}(M^2 + \gamma)X)^2 Y^2 + 2\E[Y^2].
    \]
    For $Y$ we can use the same technique as before, using that $\exp(-(a+b)^{-1}) \leq \exp(-\max(2a, 2b)^{-1}) \leq \exp(-(2a)^{-1}) + \exp(-(2b)^{-1})$, we get
    \begin{align*}
        \E[Y^2] 
        &= \int_{t>0} \Pbb(Y^2> t)\diff t
        \leq \int_{t>0} 2\exp\paren{-\frac{-nt}{a(b + 2M\gamma^{-1/2}t^{1/2}/3)}}\diff t
        \\&\leq 4\int_{t>0} \exp\paren{-\frac{-nt}{2ab}}
        + \exp\paren{-\frac{-nt^{1/2}}{4 aM\gamma^{-1/2}/3)}}\diff t
        \\&= 8 ab n^{-1} + 256 a^2M^2\gamma^{-1} n^{-2} / 9
    \end{align*}
    We conclude the lemma with the previous one.
\end{proof}

Let us now simplify the constant that appear in the bound derived so far.
\begin{lemma}[Simplifying constants]
    The constant is the previous bound can be worked out as
    \begin{equation*}
        \trace\paren{\Sigma(\Sigma+\gamma)^{-1}} \leq k,\qquad
        M \leq \lambda^{-1}k \sup \norm{\phi},\qquad
        \scap{\Pi_{\hat{\cal F}}f^*}{\Sigma(\Sigma + \gamma)^{-1}\Pi_{\hat{\cal F}}f^*}_{L^2(\rho_\X)} \leq \norm{f^*}_{L^2(\rho_\X)}.
    \end{equation*}
    We also have
    \begin{equation*}
        \norm{f^*}_{L^2(\rho_\X)} \leq \norm{f^*}_{L^\infty(\rho_\X)} \leq \sigma, \qquad
        \epsilon^2 \leq \sigma^2,\quad \sigma^2 = \sup_x\E\bracket{Y^2 \midvert X=x}
    \end{equation*}
    As a consequence, the constant $a$ appearing earlier is smaller than $(1+M)\sigma$.
\end{lemma}
\begin{proof}
    The first bound is a direct application of the fact that $\Sigma \preceq \Sigma + \lambda$, hence $\trace((\Sigma+\gamma)^{-1}\gamma) \leq \trace(I) = k$.
    The second bound is due to the fact that $\psi = \hat\Theta\phi$, hence $\norm{\psi} \leq \norm{\hat\Theta}_{\op} \norm{\phi} \leq \norm{\hat\Theta}_F \norm{\phi}$.
    In the meantime, if $\hat\Theta$ was regularized
    \[
        \lambda \norm{\hat\Theta}_F^2 \leq \hat{\cal L}(\Theta) + \lambda\norm{\hat\Theta}^2
        \leq \hat{\cal L}(0) = k.
    \] 
    For the part in $f^*$, we have that
    \[
        \norm{\Sigma^{1/2}(\Sigma + \gamma)^{-1/2}\Pi_{\hat{\cal F}}f^*}
        \leq \norm{\Pi_{\hat{\cal F}}f^*}
        \leq \norm{f^*}.
    \]
    Finally, the last equality is due to the fact that $f^*(x)$ is the mean of $Y$ conditionally to $X=x$,
    \[
        \norm{f^*(X)} = \norm{\E\bracket{Y\midvert X}} \leq \E\bracket{Y^2\midvert X}^{1/2} \leq \sigma.
    \]
    This ends the lemma
\end{proof}

\begin{lemma}
  Under Assumption \ref{ass:noise}, when $\gamma = M^2 \log(n)^{1+\delta} n^{-1}$, with $\delta > 0$, there exists a $N>0$ such that for any $n > N$, the excess of risk of the regularized empirical risk \eqref{eq:erm} minimizer reads
    \begin{equation}
        \E_{(X_i, Y_i)}[{\cal R}(f_n) - {\cal R}(f^*)]
        \leq  \frac{2k_e\epsilon^2}{n}
		+ \frac{8M^2\log(n)^{1+\delta}}{n}\norm{f^*}_{L^2(\rho_\X)}
        + \frac{64 k a}{n}
        + 2\norm{I-\Pi_{\hat{\cal F}}\Pi_{\cal F}f^*}^2
        + \norm{I-\Pi_{\cal F}f^*}^2
    \end{equation}
    where $k_e = \trace\paren{\Sigma(\Sigma+\gamma I)^{-1}} \leq k$ is the effective dimension, $a = \norm{I-\Pi_{\hat{\cal F}}f^*}_{L^\infty} \leq \norm{f^*}_{L^\infty} + M\norm{f}_{L^2}$, and $M = \sup\norm{\psi} \leq k\lambda^{-1}\sup\norm{\phi}$.
\end{lemma}
\begin{proof}
  When $\gamma = c \log(n)^{1+\delta} n^{-1}$ the excess of risk reads
    \begin{align*}
        \E_{(X_i, Y_i)}[{\cal R}(f_n) - {\cal R}(f^*)]
        \nonumber
		&\leq  \frac{k_e\epsilon^2}{n} \paren{1+\frac{M^2}{c\log(n)}}
		+ \frac{2c\log(n)^{1+\delta}}{n}\paren{1+\frac{M^2}{c\log(n)}}^2 \norm{f^*}_{L^2(\rho_\X)}
        + \frac{64 k a}{n}
        \nonumber
	  \\&+ \frac{114 a^2 M^2}{9 c n \log(n)}
	    + O(\exp(-\log(n)^{1+\delta/2}))
        + 2\norm{I-\Pi_{\hat{\cal F}}\Pi_{\cal F}f^*}^2
        + \norm{I-\Pi_{\cal F}f^*}^2
    \end{align*}
    Taking $c = M^2$ leads to the lemma.
\end{proof}

\subsection{Controlling \texorpdfstring{\eqref{eq:opera}}{Operators}}

An ideal control of \eqref{eq:opera} would leverage closed form solutions to both the population and empirical risk and use concentration inequalities on integral operators, as in \citet{Cabannes2022rates,Loucas2023}.
Yet, those proof proceed with the estimation of the smallest eigenfunctions of $(\hat\Sigma_X + \lambda)^{-1/2} \Sigma (\hat\Sigma_X + \lambda)^{-1/2}$, rather than the biggest of $\Sigma^{-1/2}(\Sigma_X-\lambda)\Sigma^{-1/2}$.
In this proof, we will rather utilize derivations based on empirical processes concentration, together with the following ``transfer bound''.

\begin{lemma}[Transfer bound]
    \label{lem:transfer}
    For $\hat{\Theta}\in\R^k\otimes{\cal H}$, and $\hat{\cal F} = \brace{x\to w^\top\hat\Theta\phi(x)\midvert w\in\R^k}$,
    \begin{equation}
        \label{eq:transfer}
        \sum_{i\in[k]} \lambda_i^2 \norm{(\Pi^{(\mu_\Xi)}_{\cal F} - \Pi^{(\mu_\Xi)}_{\hat{\cal F}})f_i}_{L^2(\mu_\Xi)}^2
        -\sum_{k< i \leq k_\lambda} \lambda_i^2 \norm{\Pi^{(\mu_\Xi)}_{\hat{\cal F}}f_i}_{L^2(\mu_\Xi)}^2
        \leq {\cal L}(\hat\Theta;\lambda) - {\cal L}(\Theta_*;\lambda),
    \end{equation}
    where $\Pi_{\cal F}^{(\tau)}$ is the projection orthogonal on ${\cal F}$ in $L^2(\tau)$.
\end{lemma}
\begin{proof}
    For simplicity, let us remove the dependency to $\mu_\Xi$ in the proof.
    Let us introduce $C = S\hat\Theta\hat\Theta^\top S^\top$, $C$ is a positive operator of rank $k$ in $L^2$, let us write it as $C = \sum_{i\in[k]} \mu_i g_ig_i^\top$ with $\mu_i \geq 0$.
    \[
        {\cal L}(\hat\Theta; \lambda) - k
        = \trace\paren{(C - T_\lambda)^2 - T_\lambda^2}
        = \trace\paren{C^2 - 2C^{1/2}T_\lambda C^{1/2}}.
    \]
    Let us decompose $T_\lambda = T_+ - T_-$ where $T_+$ and $T_-$ are positive.
    Since $T_\lambda \preceq T_+$, $-C^{1/2}T_\lambda C^{1/2} \succeq C^{1/2}T_+C^{1/2}$, hence
    \[
        {\cal L}(\hat\Theta; \lambda) - k
        \geq \trace\paren{C^2 - 2C^{1/2}T_+ C^{1/2}}
        = \sum_{i\leq k} \mu_i^2 - 2\mu_i \norm{T_+^{1/2}g_i}^2.
    \]
    Minimizing this quantity with respect to $\mu_i$, leads to 
    \[
        {\cal L}(\hat\Theta; \lambda) - {\cal L}(\Theta;\lambda)
        \geq \sum_{i\leq k} \lambda_i^2 - \sum_{i\leq k} \norm{T_+^{1/2}g_i}^4.
    \]
    Let us know introduce $(f_i)$ the eigenfunctions of $T_\lambda$.
    With $U = (\scap{g_i}{f_j}^2)_{ij} \in \R^{k\times k_\lambda}$ and $\lambda = (\lambda_i)\in\R^{k_\lambda}$, we have
    \[
        \sum_{i\leq k} \norm{T_+^{1/2}g_i}^4
        = \sum_{i\leq k} (g_i^\top T_+ g_i)^2
        = \sum_{i\leq k} \paren{\sum_{j\leq k_\lambda} \lambda_j \scap{g_i}{f_j}^2}^2
        = \sum_{j, m\leq k_\lambda} \lambda_j\lambda_m \sum_{i\leq k}\scap{g_i}{f_j}^2\scap{g_i}{f_m}^2
        = \lambda^\top U^\top U \lambda.
    \]
    Note that $U$ is at most doubly stochastic since both $(g_i)$ and $(f_i)$ are orthonormal families, thus $\norm{U} \leq 1$, and $U^\top U \preceq I$.
    If one replace the $f_i$ by $f_i / \norm{\Pi_{\hat{\cal F}}}$ in the definition of $U$ that would become $\tilde{U} = \diag((\norm{\Pi_{\hat{\cal F}}f_i}^2)_{i\leq k_\lambda})^{-1} U$, $\tilde{U}$ is still right stochastic. Hence
    \[
        U^\top U \preceq \diag(\norm{\Pi_{\hat{\cal F}}f_i}_{i\leq k_\lambda}^2)^{2}
        \diag(\norm{\Pi_{\hat{\cal F}}f_i}_{i\leq k_\lambda}^2).
    \]
    It follows that
    \[
        \sum_{i\leq k} \norm{T_+^{1/2}g_i}^4
        \leq \lambda^\top U^\top U \lambda
        = \sum_{i\leq k_\lambda} \lambda_i^2 \norm{\Pi_{\hat{\cal F}}f_i}^2.
    \]
    This allows to simplify the lower bound as
    \begin{align*}
        {\cal L}(\hat\Theta; \lambda) - {\cal L}(\Theta;\lambda)
        &\geq \sum_{i\leq k} \lambda_i^2 \paren{f_i^\top f_i - f_i^\top \Pi_{\hat{\cal F}}f_i}
        - \sum_{k<i\leq k_\lambda} \lambda_i^2 f_i^\top \Pi_{\hat{\cal F}}f_i
        \\&= \sum_{i\leq k} \lambda_i^2 \scap{f_i}{(I - \Pi_{\hat{\cal F}})f_i}
        - \sum_{k<i\leq k_\lambda} \lambda_i^2 \norm{\Pi_{\hat{\cal F}}f_i}^2
        \\&= \sum_{i\leq k} \lambda_i^2 \norm{(\Pi_{\cal F} - \Pi_{\hat{\cal F}})f_i}^2
        - \sum_{k<i\leq k_\lambda} \lambda_i^2 \norm{\Pi_{\hat{\cal F}}f_i}^2.
    \end{align*}
    This ends the proof of our transfer bound.
\end{proof}

The left-hand side in Lemma \ref{lem:transfer} is to be linked with the desired control of \eqref{eq:opera}.
In order to deal more finely with distribution-shift, we introduce the following generic variant of Assumptions \ref{ass:interpolation_simple} and \ref{ass:robust_simple}.

\begin{assumption}[Low expansion]
    \label{ass:interpolation}
    Assume that for any function of the original space of functions $f\in\Psi$ \eqref{eq:rkhs},
    \[
        \norm{f}_{L^2(\rho_\X)} \leq \zeta\paren{\norm{f}_{L^2(\mu_\X)}},
    \]
    with $\zeta:\R\to\R$ continuous, increasing and $\zeta(0) = 0$.
\end{assumption}

\begin{definition}[Distribution $\epsilon$-robustness]
    \label{def:proj_eq}
    A close convex set of functions ${\cal F}$ will be said to be $\epsilon$-robust to distribution shift conditionally to the function $f$ if
    \[
        \norm{\Pi_{\cal F}^{(\rho_\X)}f - \Pi_{\cal F}^{(\mu_\Xi)}f}_{L^2(\rho_\X)} \leq \epsilon \norm{f}_{L^2(\rho_\X)},
    \]
    where $\Pi_{\cal F}^{(\tau)}$ is the projection orthogonal on ${\cal F}$ in $L^2(\tau)$.
\end{definition}

\begin{assumption}
    \label{ass:robust}
    There exists a profile $\sigma:\R^2\to\R$ increasing and bounded such that for any $k\in\N$, $\Span\brace{f_i}_{i\in[k]}$ is $\sigma(k) $-robust to $f^*$.
\end{assumption}

\begin{lemma}[Decomposition]
    \label{lem:hook_for_proof}
    Under Assumptions \ref{ass:interpolation} and \ref{ass:robust}, with ${\cal F}_l$ the span of the $(f_i)_{i\in[l]}$
    \begin{equation}
        \norm{(I - \Pi_{\hat{\cal F}}^{(\rho_\X)})\Pi_{{\cal F}_l}^{(\rho_\X)} f^*}_{L^2(\rho_\X)} 
        \leq \sigma(l)
        + \zeta\paren{\sum_{i\leq l} \abs{\scap{f^*}{f_i}_{L^2(\mu_\Xi)}} \norm{(\Pi_{{\cal F}_l}^{(\mu_\Xi)} - \Pi_{\hat{\cal F}}^{(\mu_\Xi)})f_i}_{L^2(\mu_\Xi)}}.
    \end{equation}
\end{lemma}
\begin{proof}
    Using the fact that $I - \Pi$ is a projection when $\Pi$ is a projection, and that projections contract distance, we get
    \begin{align*}
        \norm{(I - \Pi_{\hat{\cal F}}^{(\rho_\X)})\Pi_{{\cal F}_l}^{(\rho_\X)} f^*}_{L^2(\rho_\X)} 
        &\leq \norm{(I - \Pi_{\hat{\cal F}}^{(\rho_\X)})(\Pi_{{\cal F}_l}^{(\rho_\X)} - \Pi_{{\cal F}_l}^{(\mu_\Xi)}) f^*}_{L^2(\rho_\X)} 
        + \norm{(I - \Pi_{\hat{\cal F}}^{(\rho_\X)})\Pi_{{\cal F}_l}^{(\mu_\Xi)}f^*}_{L^2(\rho_\X)} 
        \\&\leq \norm{(\Pi_{{\cal F}_l}^{(\rho_\X)} - \Pi_{{\cal F}_l}^{(\mu_\Xi)}) f^*}_{L^2(\rho_\X)} 
        + \norm{(I - \Pi_{\hat{\cal F}}^{(\rho_\X)})\Pi_{{\cal F}_l}^{(\mu_\Xi)}f^*}_{L^2(\rho_\X)}.
    \end{align*}
    Under Assumption \ref{ass:robust}, the first term in the right-hand side of the previous equation is bounded by $\sigma(l)$.
    Regarding the second term, under Assumption \ref{ass:interpolation}, for $f\in \Psi$ and $f'\in\hat{\cal F}\subset \Psi$, we have
    \[
        \norm{(I - \Pi_{{\cal F}_l}^{(\rho_\X)})f}_{L^2(\rho_\X)} \leq \norm{f - f'}_{L^2(\rho_\X)}
        \leq \zeta\paren{\norm{(f - f'}_{L^2(\mu_\Xi)}}.
    \]
    Taking the minimum on the right-hand side and using the fact that $\zeta$ is increasing leads to
    \[
        \norm{(I - \Pi_{{\cal F}_l}^{(\rho_\X)}) f}_{L^2(\rho_\X)} 
        \leq \zeta\paren{\norm{(I - \Pi_{{\cal F}_l}^{(\mu_\Xi)}) f}_{L^2(\mu_\Xi)}}.
    \]
    Applied to $\Pi_{{\cal F}_l}^{(\mu_\Xi)}f^*$, this leads to 
    \[
        \norm{(I - \Pi_{\hat{\cal F}}^{(\mu_\Xi)})\Pi_{{\cal F}_l}^{(\rho_\X)}f^*}_{L^2(\rho_\X)}
        \leq \zeta\paren{\norm{(I - \Pi_{\hat{\cal F}}^{(\mu_\Xi)})\Pi_{{\cal F}_l}^{(\mu_\Xi)}f^*}_{L^2(\mu_\Xi)}}.
    \]
    We are done with all the quantities that relate to the distribution shift. 
    Under Assumption \ref{ass:source}, we have
    \begin{align*}
        \norm{(I - \Pi_{\hat{\cal F}}^{(\mu_\Xi)})\Pi_{{\cal F}_l}^{(\mu_\Xi)} f^*}_{L^2(\mu_\Xi)} 
        &= \norm{\sum_{i\leq l} \scap{f^*}{f_i}_{\mu_\Xi} (I - \Pi_{\hat{\cal F}}^{(\mu_\Xi)})f_i}_{L^2(\mu_\Xi)} 
        \\&\leq \sum_{i\leq l} \abs{\scap{f^*}{f_i}_{\mu_\Xi}} \norm{(I - \Pi_{\hat{\cal F}}^{(\mu_\Xi)})f_i}_{L^2(\mu_\Xi)}.
    \end{align*}
    Collecting the previous equations leads to the lemma.
\end{proof}

To get a finer control of \eqref{eq:opera}, remark that the left-hand side of \eqref{eq:transfer} has some additional constraints that can help us to tighten our bound.
For simplicity, we will remove all the dependency to $\mu_\Xi$ in the following.
In essence, we want to lower bound the $\lambda_i^2$ and to upper bound the $\norm{(\Pi_{\cal F} - \Pi_{\hat{\cal F}})f_i}$.
The next lemma adds a constraint the maximal error one can make on \eqref{eq:opera} under a constraint on ${\cal L}(\Theta;\lambda)$.

\begin{lemma}
    \label{lem:proj}
    When ${\cal F}$ is of dimension $k$ and $\hat{\cal F}$ is of dimension $k'$ we have
    \begin{equation}
        \sum_{i\leq k} \norm{(\Pi_{\cal F} - \Pi_{\hat{\cal F}})f_i}^2
        = k-k'+\sum_{i > k}\norm{\Pi_{\hat{\cal F}}f_i}^2 \leq k.
    \end{equation}
\end{lemma}
\begin{proof}
    Let us consider two projection $U$ and $V$ onto the span of $(u_i)_{i\in[k]}$ and $(v_i)_{i\in[k]}$ with $(u_i)_{i\in\N}$ and $(v_i)_{i\in\N}$ two orthonormal basis of the ambient space.
    We have, with Hilbert-Schmidt norm everywhere,
    \begin{align*}
        \norm{U(I-V)}^2 &= \norm{U}^2 - \norm{UV}^2 = k - \norm{UV}^2 
        = k - \norm{(UV)^\top}^2 
        = k - k' + k' - \norm{VU}^2 
        = k - k' + \norm{V(I-U)}^2.
    \end{align*}
    Based on invariant of the Hilbert-Schmidt norm to adjoint, and the fact that projection are self-adjoint, we have
    \[
        \norm{U(I-V)}^2 = \norm{(I-V)U}^2 = k-k' + \norm{V(I-U)}^2 = k-k' + \norm{(I-U)V}^2.
    \]
    Finally, we also know that since projection contracts distances $\norm{(I-V)U}^2 \leq \norm{U}^2 = k$.
    The claim of the lemma consists in writing explicitly
    \begin{align*}
        \norm{(I - \Pi_{\hat{\cal F}})\Pi_{\cal F}}^2
        &= \norm{(\Pi_{\cal F} - \Pi_{\hat{\cal F}})\Pi_{\cal F}}^2 
        = \sum_{i\leq k}\norm{(\Pi_{\cal F} - \Pi_{\hat{\cal F}})f_i}^2 
        \\&= k-k' + \norm{\Pi_{\hat{\cal F}}(I-\Pi_{\cal F})}^2
        = k-k'+\sum_{i > k}\norm{\Pi_{\hat{\cal F}}f_i}^2 \leq k.
    \end{align*}
    This is lead to the statement of the lemma.
\end{proof}

Given a control on \eqref{eq:vic}, finding an upper bound on \eqref{eq:opera} reduces to a purely algebraic one. 
In order to find the worse value that $\sum_{i\leq k} \abs{\scap{f^*}{f_i}}\norm{(\Pi_{\cal F}^{(\mu_\Xi)} - \Pi_{\hat{\cal F}})f_i}_{L^2(\mu_\Xi)}$ can take, let us introduce
\begin{equation}
    \label{eq:algebraic}
    x_i = \norm{(\Pi_{\cal F} - \Pi_{\hat{\cal F}})f_i},\quad
    c_i = \abs{\scap{f^*}{f_i}}.
\end{equation}
The previous results lead to the following maximization problem in order to find the worse value of \eqref{eq:opera},
\begin{align}
    \label{eq:max_pb}
    &\max_x \sum_{i\leq k} c_i x_i
    \\\text{subject to}\quad &
    \sum_{i\leq k} \lambda_i^2 x_i^2 - \sum_{k< i\leq k_\lambda} \lambda_i^2 x_i^2 \leq \epsilon \tag{Lemma \ref{lem:transfer}}
    \\&\sum_{i\leq k} x_i^2 = k-k' + \sum_{k< i\leq k_\lambda}x_i^2 \leq k \tag{Lemma \ref{lem:proj}}
\end{align}

\subsection{Keeping it simple and concluding after controlling \texorpdfstring{\eqref{eq:source}}{Source}}

Solving smartly the algebraic problem above to get the best bound on \eqref{eq:opera} requires distinguishing between many cases.
While it might be relevant to distinguish those different cases and show different convergence regimes, this subsection proceed in a simpler way, although less tight.
In particular, we can simplify the problem with respect to the $(x_i)_{i k}$, using the fact that $k' \leq k$ (it is minimum between the number of positive eigenvalues of $\hat T_\lambda$ based on samples and $k$), it leads to $x_{k+1}^2 = \sum_{i\leq k} x_k^2$ and $x_{k+1+j}^2 = 0$, \eqref{eq:max_pb} becomes 
\begin{align}
    \tag{\ref{eq:max_pb}}
    &\max_x \sum_{i\leq k} c_i x_i
    \\\text{subject to}\quad &
    \sum_{i\leq k} (\lambda_i^2 - \lambda_{k+1}^2)x_i^2 \leq \epsilon\nonumber
\end{align}

In general, one could refine this formulation by introducing a probability argument that tells us how much one can expect the error between $\Pi_{\hat{\cal F}}$ and $\Pi_{\cal F}$ to concentrates on the eigenspace linked to the smallest eigenvalue of $T_\lambda^2$. 
The problem shows two behaviors, if the $c_i$ decrease faster than the $\lambda_i$ than we want to charge the energy of $(x_i)_{i\leq k}$ on the smallest indices.
Otherwise, we want to charge the $(x_i)_{i\leq k}$ on the biggest indices. 

To keep it simple, we will optimize ${\cal L}$ without any rank restriction first, which allow considering $\lambda_{k_\lambda+1} = 0$, before thresholding the rank to get to a space of dimension $k$.

\begin{lemma}
    Under Assumptions \ref{ass:interpolation} and \ref{ass:robust}, with ${\cal F}_{l}$ the span of the first $l$ eigenfunctions of $T_\lambda$, 
    \begin{align}
        \nonumber
        &\norm{(I - \Pi_{\hat{\cal F}}^{(\rho_\X)}) f^*}_{L^2(\rho_\X)}^2
        \\&\qquad\leq \inf_{l\leq k} \norm{(I - \Pi_{{\cal F}_l}^{(\rho_\X)})f^*}_{L^2(\rho_\X)}^2 + 4\sigma(l)^2 + 4\zeta^2\paren{\norm{\tilde{T_\lambda}^{-1}\Pi_{{\cal F}_l}^{(\mu_\Xi)} f^*}_{L^2(\mu_\Xi)} \paren{{\cal L}(\hat\Theta;\lambda) - {\cal L}(\Theta;\lambda)}^{1/2}}.
    \end{align}
    where $\tilde T_{\lambda} = \sum_{i\in[k]} (\lambda_i^2 - \lambda_{k+1}^2)^{1/2} f_i f_i^\top$.
    Moreover, when the search for $\hat{\cal F}$ is done without rank restriction on $\Theta$, before thresholding to get reduce $\hat{\cal F}$ to a space of dimension $k$, under the strong Assumptions \ref{ass:interpolation_simple} and \ref{ass:robust_simple}, as well as Assumption \ref{ass:source} 
    \begin{equation}
        \norm{(I - \Pi_{\hat{\cal F}_k})f^*}^2
        \leq \abs{k-k_\lambda}\norm{f^*}_{L^2(\rho_\X)}^2
        + 2c_r \norm{T_\lambda^{-1}f^*}_{L^2(\mu_\Xi)}^2\brace{{\cal L}(\hat\Theta;\lambda) - {\cal L}(\Theta_*;\lambda)}.
    \end{equation}
\end{lemma}
\begin{proof}
    Keeping the algebraic notation above, this comes from a simple application of Cauchy-Schwarz, for $(a_i)\in\R^k$
    \[
        \sum_{i\leq l} c_i x_i = \sum_{i\in[l]} \frac{c_i}{a_i} a_i x_i \leq \paren{\sum_{i\leq [l]} \frac{c_i^2}{a_i^2}}^{1/2} \paren{\sum_{i\in[l]} a_i^2 x_i^2}^{1/2}.
    \]
    When applies to the quantities in \eqref{eq:algebraic} and $a_i = \lambda_i^2 - \lambda_{k+1}^2$ and $l\leq k$, the previous lemma leads to
    \begin{align*}
        \norm{(I - \Pi_{\hat{\cal F}}^{(\rho_\X)})\Pi_{{\cal F}_l}^{(\rho_\X)} f^*}_{L^2(\rho_\X)} 
        &\leq \sigma(l)
        + \zeta\paren{\sum_{i\leq l} \abs{\scap{f^*}{f_i}_{L^2(\mu_\Xi)}} \norm{(\Pi_{{\cal F}_l}^{(\mu_\Xi)} - \Pi_{\hat{\cal F}}^{(\mu_\Xi)})f_i}_{L^2(\mu_\Xi)}}
        \\&\leq \sigma(l) + \zeta\paren{\sum_{i\leq l} c_ix_i}
        \leq \sigma(l) + \zeta\paren{\paren{\sum_{i\leq l} \frac{c_i^2}{a_i^2}}^{1/2} \paren{\sum_{i\in[l]} a_i^2 x_i^2}^{1/2}}
        \\&\leq \sigma(l) + \zeta\paren{\paren{\sum_{i\leq l} \frac{c_i^2}{a_i^2}}^{1/2} \paren{{\cal L}(\hat\Theta;\lambda) - {\cal L}(\Theta;\lambda)}^{1/2}}.
    \end{align*}
    We conclude by remarking that $\sum_{i\leq l} \frac{c_i^2}{a_i^2} = \norm{\tilde{T}_\lambda^{-1}\Pi_{{\cal F}_l^{(\mu_\Xi)}}f^*}_{L^2(\mu_\Xi)}$. 

    For the second part, set $\hat{\cal F}_k$ the $k$ first eigenfunctions to the all the one retrieve with the empirical minimization of ${\cal L}$, and ${\cal F}$ to be the span of all the eigenfunctions linked with positive eigenvalues of $T_\lambda$.
    Let us rework the decomposition of the excess of risk, we have
    \begin{align*} 
        \norm{(I - \Pi_{\hat{\cal F}_k})f^*}^2
        &= \norm{\Pi_{\hat{\cal F}_{k_\lambda}}(I - \Pi_{\hat{\cal F}_k})f^*}^2
        + \norm{(I - \Pi_{\hat{\cal F}_{k_\lambda}})(I - \Pi_{\hat{\cal F}_k})f^*}^2
        \\&= \norm{(\Pi_{\hat{\cal F}_{k_\lambda}} - \Pi_{\hat{\cal F}_k})f^*}^2
        + \norm{(I - \Pi_{\hat{\cal F}_{k_\lambda}})f^*}^2
        \\&\leq \norm{(\Pi_{\hat{\cal F}_{k_\lambda}} - \Pi_{\hat{\cal F}_k})f^*}^2
        + 2\norm{(I - \Pi_{\hat{\cal F}_{k_\lambda}})\Pi_{\cal F}f^*}^2
        + \norm{(I - \Pi_{\cal F})f^*}^2
        \\&\leq \abs{k-k_\lambda}\norm{f^*}^2
        + 2\norm{(I - \Pi_{\hat{\cal F}_{k_\lambda}})\Pi_{\cal F}f^*}^2. 
    \end{align*}
    The last bound begin due to Assumption \ref{ass:source}, as well as the lax bounding that on the operator norm of two projections.
    When one could remove the $k-k_\lambda$ we let it as we expect the quantity to behave it this way, with a constant similar to $\norm{f^*}^2 / k_\lambda$ instead of $\norm{f^*}^2$.
\end{proof}

We can now state the master theorem.

\begin{theorem}
    \label{thm:conv_master}
    Under Assumptions \ref{ass:source}, \ref{ass:noise}, \ref{ass:interpolation} and \ref{ass:robust}, there exists a regularizer $\gamma$ such that the regularized empirical risk minimizer verifies that: for any $\delta>0$, there exists an $N_\delta >0$ such that for any $n > N_\delta$, the excess of risk of the regularized empirical risk \eqref{eq:erm} minimizer reads
    \begin{align}
        \nonumber
        &{\cal R}(f) - {\cal R}(f^*)
        \leq \frac{2k_e\epsilon^2}{n}
		+ \frac{8M^2\log(n)^{1+\delta}}{n}\norm{f^*}_{L^2(\rho_\X)}
        + \frac{64 k a}{n}
        \\&\qquad + \inf_{l\leq k} \norm{(\Pi_{{\cal F}_{k_\lambda}} - \Pi_{{\cal F}_l}^{(\rho_\X)})f^*}_{L^2(\rho_\X)}^2 
        + 4\sigma(l)^2 + 4\zeta^2\paren{\norm{\tilde{T_\lambda}^{-1}\Pi_{{\cal F}_l}^{(\mu_\Xi)} f^*}_{L^2(\mu_\Xi)} \paren{{\cal L}_k(\hat\Theta;\lambda) - {\cal L}_k(\Theta;\lambda)}^{1/2}}.
    \end{align}
    where ${\cal F}_l$ the span of $l$-th first eigenfunction of $T_\lambda$, $k_\lambda$ the number of strictly positive eigenfunctions of $T_\lambda$, $k_e\leq k$ is the effective dimension of $\psi$ in $L^2(\rho_\X)$, $a = \norm{I-\Pi_{\hat{\cal F}}f^*}_{L^\infty} \leq \norm{f^*}_{L^\infty} + M\norm{f}_{L^2}$, $M = \sup\norm{\psi} \leq k\lambda^{-1}\sup\norm{\phi}$, and $\tilde T_{\lambda} = \sum_{i\in[k]} (\lambda_i^2 - \lambda_{k+1}^2)^{1/2} f_i f_i^\top$.
    Moreover, under the sole Assumptions \ref{ass:interpolation_simple} and \ref{ass:robust_simple}, we have the simpler bound
    \begin{align*}
        \nonumber
        {\cal R}(f) - {\cal R}(f^*)
        &\leq \frac{2k_e\epsilon^2}{n}
		+ \frac{8M^2\log(n)^{1+\delta}}{n}\norm{f^*}_{L^2(\rho_\X)}
        + \frac{64 k a}{n}
        + \max(k-k_\lambda, 0)\norm{f^*}_{L^2(\rho_\X)}^2
	  \\&\qquad\qquad+ 2c_r \norm{T_\lambda^{-1} \Pi_{{\cal F}_\lambda} f^*}_{L^2(\mu_\Xi)}^2\brace{{\cal L}_{k_\lambda}(\hat\Theta;\lambda) - {\cal L}_{k_\lambda}(\Theta_*;\lambda)}
		+ \norm{(I - \Pi_{{\cal F}_{\lambda}}) f^*}_{L^2(\mu_\Xi)}
    \end{align*}
	Where $\hat\Theta$ is understood as belonging to $\R^{k_\lambda}\otimes {\cal H}$ in this last expression and ${\cal F}_\lambda$ the eigenspace linked with positive eigenvalues of $T_\lambda$.
\end{theorem}

\subsection{Discussion}

\subsubsection{Finite number of positive eigenvalues}

The following result relates the eigenvalues of $T_\lambda$ with those of $K$.
It notably proves that $k_\lambda$ is finite when $K$ is trace-class, which is one claim of Theorem \ref{thm:conv_simple}.

\begin{lemma}[Relating capacity between $K$ and $T_\lambda$]
    \label{lem:capacity}
    If $(\mu_i)$ are the eigenvalues of $K$, then the number of eigenvalues of $T_\lambda$ that are bigger than $t\in\R$ is smaller than the cardinality of  
    \(
        \brace{i\midvert \mu_i > \lambda / (1-t)}.
    \)
    Moreover, if there exists $q>0$ such that $\trace\paren{K^{1/q}} < +\infty$, then there exists a $c_q$ such that if $(\mu_i)$ are the eigenvalues of $K$, we have
    \(
        \mu_i \leq c_q i^{-q}.  
    \)
    As a consequence, in this setting, for any $t\in\R$, the number of eigenvalues of $T_\lambda$ that is bigger than $t$ is smaller than $(c_q(1-t)/\lambda)^{1/q}$.
\end{lemma}
\begin{proof}
    Let us consider the set of eigenvectors $(f_i)$ whose eigenvalues are bigger than $t$.
    Consider the span of this space, we want to quantify its dimension.
    We know that all unitary vectors in this span satisfies
    \[
        t \leq x^\top T_\lambda x \leq x^\top Tx - \lambda x^\top K^{-1}x \leq 1 - \lambda x^\top K^{-1}x
    \]
    Hence
    \[
        x^\top K^{-1} x \leq \frac{1 - t}{\lambda} 
    \]
    This means that this span does not intersect the span of $\phi_i$ for $\phi_i$ the eigenvectors of $K^{-1}$ such that the eigenvalues are bigger than  $\lambda / (1 - t)$.
    In other terms, this linear space does not intersect a linear space of co-dimension $d$ where $d$ is the cardinality mentioned in the lemma statement.
    Let us denote by $U$ the space we are interested in and by $V$ the space it does not intersect beside in the origin, and by $E$ the ambient space
    Since $U\cap V = \brace{0}$, the quotient $(U+V)/V$ is isomorphic to $U$, hence
    \[
        \dim(U) = \dim\paren{ \frac{U+V}{V} } \leq \dim\paren{\frac{E}{V}} = \operatorname{codim}(V) = d.
    \]
    This concludes the proof of the first part of the lemma.

    The second claim follows from the fact that $\mu_i^{1/q}$ are summable and decreasing, hence the sequence $S_n = \sum_{i\leq n} \mu_i^{1/q}$ is a Cauchy sequence.
    As a consequence, there exists $N\in\N$, such that for any $s > N/2$, we have
    \[
        s\mu_{2s}^{1/q} \leq S_{2s} - S_s \leq 1/2.
    \]
    Hence, for all $s\geq N$, we have $\mu_{s} \leq s^{-1/q}$, hence $\mu_s / s^{-1/q}$ is bounded.
    Denoting by $c_q$ the maximum, leads to the first result. 
    The final statement is a consequence of the fact that $c_q i^{-q} > \lambda / (1-t)$ implies $i < (c_q(1-t) / \lambda)^{1/q}$.
\end{proof}

\begin{example}
    \label{ex:rbf}
    When considering the radial basis function kernel $\phi(x)^\top\phi(x') = \exp(-\norm{x-x'}^2)$, $\Psi$ is the space of analytical functions \citep{Sun2008}, which is known to be small compared to $L^2$ spaces \citep{Kolmogorov1959}.
    As a consequence, one can think as $q=+\infty$ in the previous lemma.
    More in general, when $\phi$ is bounded, $K$ is trace-class and one can take $q = 1$.
\end{example}
\begin{proof}
    The capacity of $K$ is relates to the capacity of $K(\brace{f\midvert \norm{f}_{L^2(\mu_\Xi)} \leq 1})$, which itself relates to the capacity of $\Psi = \ima K^{1/2}$.
    This explains why $q$ can be taken, in essence, as arbitrarily big \citep{Bach2023}.

    When $\phi$ is bounded, the following
    \begin{align*}
        \trace\paren{K} 
        &= \trace\paren{SS^\top} = \trace\paren{S^\top S} = \trace\paren{\E[\phi(X)\phi(X)^\top]}
        = \E[\trace\paren{\phi(X)\phi(X)^\top}]
        \\&= \E[\phi(X)^\top\phi(X)]
        = \E[\norm{\phi(X)}^2] < +\infty,
    \end{align*}
    proves that $K$ is trace class.
\end{proof}

\subsubsection{Derivation for vanishing bias}
In the main text, we have assumed that $T_\lambda$ was the right operator to define the solution of the representation learning (which explains Assumption \ref{ass:source}).
This might offend the purist as it would be nicer to define a principled solution that does not depend on the choice of the architecture (yet that might be easier to approximate with some architecture than others).
This suggests studying the behavior of the last expression in Theorem \ref{thm:conv_master} when $\lambda$ goes to zero.

We let for future work a more precise study of the inductive bias in this vanishing setting: 
in essence, the choice of architecture $\Psi$ perturbs $T$ by $\lambda K^{-1}$ to make it $T_\lambda$, and ideally, we would like to quantify the speed at which $T_\lambda$ converges to $T$ when seen through the eyes of $f^*$ as we decrease the regularization parameter.
In the kernel regime, it could be characterized by perturbation theory \citep{kato_perturbation_1995}, and refinement of Davis-Kahan theorem \citep{davis_rotation_1970} taking into account Assumption~\ref{ass:source}.
Moreover, when $K$ and $T$ commute, the interplay can be studied in a more direct fashion thanks to Proposition \ref{prop:commute}.

%% file: appendix/upstream.tex
\section{Control of the upstream excess of risk}
\label{app:upstream}
 
In order to control the excess of risk, one can use technique steaming from optimization as well as technique steaming from classical statistical learning.

\subsection{Rademacher complexity}
First, let us remark that ${\cal L}$ is a quadratic function when parameterized with $\Lambda = \Theta^\top \Theta \in {\cal H}\otimes{\cal H}$.

\begin{lemma}
    \label{lem:quadra}
    Let $\Theta\in\R^k\otimes {\cal H}$, denote $\Lambda = \Theta^\top \Theta \in {\cal H}\otimes{\cal H}$
    \begin{equation}
        {\cal L}(S\Theta) = 
        2(\beta - 1)\E_{\xi}[\scap{\Lambda}{\phi(\xi)\phi(\xi)^\top}] - 2\beta\E_X\E_{\xi, \xi'}\bracket{\scap{\Lambda}{\phi(\xi')\phi(\xi)^\top}\midvert X} 
        + \E_{\xi, \xi'}\bracket{\scap{\Lambda}{\phi(\xi)\phi(\xi')^\top}^2} + k.
    \end{equation}
    Moreover, the regularization reads $\lambda \norm{\Theta}^2 = \lambda \trace{\Lambda} = \lambda \scap{\Lambda}{I}$.
\end{lemma}
\begin{proof}
    Consider $\psi = \Theta\phi$, we have
    \begin{align*}
    &{\cal L}(\psi; \beta) 
    = 2(\beta - 1)\E_{\xi}[\psi(\xi)^\top\psi(\xi)] - 2\beta\E_X\E_{\xi, \xi'}\bracket{\psi(\xi)^\top\psi(\xi')\midvert X} 
    + \E_{\xi, \xi'}\bracket{(\psi(\xi')^\top\psi(\xi))^2} + k.
    \\&\quad= 2(\beta - 1)\E_{\xi}[\phi(\xi)^\top \Lambda \phi(\xi)] - 2\beta\E_X\E_{\xi, \xi'}\bracket{\phi(\xi)^\top\Lambda\phi(\xi')\midvert X} 
    + \E_{\xi, \xi'}\bracket{(\phi(\xi')^\top\Lambda\phi(\xi))^2} + k.
    \\&\quad= 2(\beta - 1)\E_{\xi}[\trace\paren{\Lambda \phi(\xi)\phi(\xi)^\top}] - 2\beta\E_X\E_{\xi, \xi'}\bracket{\trace\paren{\Lambda \phi(\xi')\phi(\xi)^\top}\midvert X} 
    + \E_{\xi, \xi'}\bracket{\trace\paren{\Lambda\phi(\xi)\phi(\xi')^\top}^2} + k.
    \end{align*}
    The lemma follows from the characterization of the Hilbert-Schmidt geometry with the trace, the fact that $\Lambda$ is self-adjoint, and that the regularization reads $\norm{\Theta}^2 = \trace{\Theta^\top\Theta}$.
\end{proof}

Let us recall three useful facts from the statistical learning literature.

\begin{lemma}
    Let ${\cal R}(\zeta) = \E_Z[\ell(\zeta, Z)]$, $\zeta^*$ be the minimizer of ${\cal L}$ inside a domain for $\zeta$, and $\zeta_n$ be the minimizer of ${\cal R}_{(Z_i)}(\zeta) = \frac{1}{n} \sum_{i\in[n]} \ell(\zeta, Z_i)$ based on exchangeable data $Z_i$ such that $\E_{(Z_i)}[{\cal R}_{(Z_i)}] = {\cal R}$.
    The average excess of risk of $\zeta_n$ is bounded by Rademacher complexity as
    \begin{equation}
        {\cal R}(\zeta_n) - {\cal R}(\zeta_*) \leq 4 \E_{(Z_i), (\sigma_i)}\bracket{\sup_{\zeta}\frac{1}{n}\sum_{i=1}^n \sigma_i\ell(\zeta, Z_i)}
    \end{equation}
    where $\sigma_i$ are i.i.d variables taking values one and minus one with probability one half.
\end{lemma}
\begin{proof}
    The proof is a classical result from learning theory \citep{bartlett_rademacher_2002}, its proof consists in introducing both the empirical risk of $\zeta_n$ and $\zeta$, and bounding the difference between the empirical and population of $\zeta$ by the supremum of this deviation over the entire domain of $\zeta$.
    This is followed by the replacement of the population risk by the average empirical one, and a symmetrization trick that introduce the variable $(\sigma_i)$ based on exchangeability of the $(Z_i)$.
\end{proof}

\begin{lemma}
    For linear model, the Rademacher complexity can be bounded as
    \begin{equation}
        \E_{(Z_i),(\sigma_i)}\bracket{\sup_{\norm{\zeta}\leq M} \frac{1}{n}\sum_{i=1}^n \sigma_i \scap{Z_i}{\zeta}} \leq \frac{M}{\sqrt{n}} \sqrt{\E[\norm{Z}^2]}.
    \end{equation}
\end{lemma}
\begin{proof}
    This is a classical result on Rademacher complexity of ball constraints predictors \citep{bartlett_rademacher_2002}.
\end{proof}

\begin{lemma}
Moreover, when $h:\R\to\R$ is Lipschitz, the following contraction principle holds
\[
   \E[\sup_f \frac{1}{n}\sum_{i=1}^n \sigma_i h(f(Z_i))] \leq \sup \norm{\diff h(x)} \E[\sup_f \frac{1}{n}\sum_{i=1}^n \sigma_i f(Z_i)]
\]
\end{lemma} 
\begin{proof}
    This follows from contraction of space capacity by Lipschitz functions \citep{Vitushkin1954}, see \citet{meir_contraction_2003} for a proof in the context of machine learning.
\end{proof}

We can now state the convergence property based on Rademacher complexity.
\begin{lemma}
    Let $\Theta_n\in\R^k\otimes {\cal H}$ be the minimizer of the unbiased regularized empirical version of ${\cal L}$ based on a dataset ${\cal D}_n$. 
    Assume that ${\cal D}_n$ is built from $n$ input samples $(X_i)$ and $m$ augmentation per samples $(\xi_{ij})$, then the average excess of risk is bounded by
    \begin{equation}
        \E_{{\cal D}_n}[{\cal L}(S\Theta_n)] - {\cal L}(S\Theta) 
        \leq \frac{8\kappa^2 \sup\norm{\Lambda}_{HS}}{\sqrt{n}} \paren{\frac{m+1+\beta}{m} + \frac{\sqrt{2}\kappa^2 \sup\norm{\Lambda}_{HS}(m^2 + 1)}{m^2}},
    \end{equation}
    where $\kappa$ is a bound on $\norm{\phi(X)}$.
\end{lemma}
\begin{proof}
    Following the previous lemmas on Rademacher complexity we have
    \begin{align*}
        &\E_{{\cal D}_n}[{\cal L}(S\Theta_n);\lambda] - {\cal L}(S\Theta;\lambda) 
        \leq 8 \E_{{\cal D}_n, \sigma}\bracket{\sup_{\Lambda} \frac{1-\beta}{n} \sum_{i\in[n]}\sigma_i\scap{\Lambda}{\frac{1}{m}\sum_{j\in[m]}\phi(\xi_{ij})\phi(\xi_{ij})^\top}} 
        \\&\qquad\qquad+ 8 \E_{{\cal D}_n, \sigma}\bracket{\sup_{\Lambda} \frac{\beta}{n}\sum_{i\in[n]}\sigma_{i}\scap{\Lambda}{\frac{2}{m}\sum_{j\in [m/2]; j+k-1=m}\phi(\xi_{ij})\phi(\xi_{ik})^\top}}
        \\&\qquad\qquad+ 4 \E_{{\cal D}_n,\sigma}\bracket{\frac{2}{n}\sum_{i \in [n/2]; i+j-1=n}\sigma_{i} \frac{1}{m^2} \sum_{k,l\in[m]}\scap{\Lambda}{\phi(\xi_{ik})\phi(\xi_{jk})^\top}^2}
        \\&\leq \frac{8\sup \norm{\Lambda}_{HS}}{\sqrt{n}} \paren{(1-\beta)\E_X\bracket{\E\bracket{\big\|\frac{1}{m}\sum_{i=1}^m \phi(\xi_i)\phi(\xi_i)^\top\big\|_{HS}^2 \midvert X}}^{1/2}} 
        \\&\qquad\qquad + \frac{8\sup \norm{\Lambda}_{HS}}{\sqrt{n}} \paren{\beta\E_X\bracket{\E\bracket{\big\|\frac{2}{m}\sum_{i,j=1}^{m/2} \phi(\xi_i)\phi(\xi_j)^\top\big\|_{HS}^2 \midvert X}}^{1/2}}
        \\&\qquad\qquad+ \frac{8\sup \norm{\Lambda}_{HS}}{\sqrt{n}} \paren{\sqrt{2}\sup \abs{\scap{\Lambda}{\phi(\xi)\phi(\xi')^\top}} \E\bracket{\big\|\frac{1}{m^2}\sum_{i,j=1}^{m} \phi(\xi_i)\phi(\xi_j)^\top\big\|_{HS}^2}^{1/2} }.
    \end{align*}
    To work out those terms, remark that if $(Z_i)$ are i.i.d. variables,
    \[
        \E[\norm{\frac{1}{p} \sum_{i\in[p]}Z_i}^2] 
        = \E[\norm{\frac{1}{p}\sum_{i\in[p]} Z_i - \E[Z]}^2] + \norm{\E[Z]}^2
        = \frac{1}{p}\E[\norm{Z - \E[Z]}^2] + \norm{\E[Z]}^2.
    \]
    While one could work out each term, the lemma consists in simply bounding $\phi$ by $\kappa$, hence all the mean and standard deviation one can obtain with expression of $\phi$ by $\kappa$.
\end{proof}

The expression in the main text is due to the following lemma.
\begin{lemma}
    \label{lem:domain}
    When minimizing a regularized risk, one can reduce the search of $\Theta$ under the constraint
    \(
        \norm{\Lambda}_{HS} \leq \lambda^{-1} k.
    \)
\end{lemma}
\begin{proof}
    When regularizing we have
    \[
        \norm{\Lambda}_{HS} = \norm{\Theta^\top\Theta}_{HS} \leq \norm{\Theta}_{\op} \norm{\Theta}_{HS} \leq \norm{\Theta}_{HS}^2,
    \]
    and for minimizer of the empirical or population risk
    \[
        \lambda\norm{\Theta}^2 \leq {\cal L}(S\Theta) + \lambda\norm{\Theta}^2 \leq {\cal L}(0) = k,
    \]
    which explains the statement of the lemma.
\end{proof}

The attentive reader would remark that compared to the bound of \citet{haochen_provable_2021} we gain a factor $k^{-1/2}$. Indeed, this factor could be recovered in \citet{haochen_provable_2021} by using the techniques of \citet{Maurer2016} rather than a trivial bound on Rademacher complexity of vector-valued function spaces in $k\max_{i\in[k]}\hat{\cal R}({\cal F}_i)$ with \citet{haochen_provable_2021} notations.

\subsection{Convex optimization}
When a least-square problem benefits from additional structure, such as smoothness or strong convexity, results from convex optimization could lead to improvement over the usual convergence rates in $n^{-1/2}$.
Recall basic results from convex optimization.

\begin{lemma}
    \label{lem:optim}
    Let ${\cal L}(\Theta) = \E_Z[\ell(\Theta, Z)]$ be a convex function optimized over a convex domain.
    Given $n$ samples $(Z_i)$, (unbiased) stochastic gradient descent with final averaging can achieve an excess of risk
    \begin{equation}
        \E_{(Z_i)}{\cal L}(\hat\Theta) - {\cal L}(\Theta_*) \leq \sqrt{2} M V n^{-1/2}
    \end{equation}
    with $M^2 = \norm{\Theta_* - \Theta_0}^2$ and $V^2 = \E[\norm{\nabla_\Theta\ell(\Theta, Z_i)}^2]$.
    Moreover, if ${\cal L}$ is $\alpha$-smooth, then it can achieve
    \begin{equation}
        \E_{(Z_i)}{\cal L}(\hat\Theta) - {\cal L}(\Theta_*) \leq \sqrt{2} M\sigma n^{-1/2} + \alpha M^2 n^{-1}
    \end{equation}
    where $\sigma^2 = \E[\norm{\nabla {\cal L} - \nabla \ell}^2]$.
    Finally, when ${\cal L}$ is $\alpha$-strongly convex, it achieves
    \begin{equation}
        \E_{(Z_i)}{\cal L}(\hat\Theta) - {\cal L}(\Theta_*) \leq \frac{2V^2}{\alpha (n+1)}.
    \end{equation}
    As a consequence, given $n$ data samples, there exists an empirical estimate of $\hat\Theta$ that guarantee those generalization bounds.
\end{lemma}
\begin{proof}
    This lemma is a direct consequence of Theorems 6.1, 6.2 and 6.3 of \citet{Bubeck2015}.
\end{proof}

It should be noted that when parameterized with $\Lambda = \Theta^\top \Theta$, ${\cal L}$ is a quadratic form as stated by Lemma \ref{lem:quadra}, yet it is minimized over a non-convex domain, the domain of symmetric operator of rank $k$.
We will relax this constraint and consider the harder problem of optimizing over $\Lambda$ in the set of self-adjoint positive operators.
This is justified by the fact that Theorem \ref{thm:conv_master} provides guarantee on the downstream task, even when one relaxes the rank constraint on $\Lambda$.

To benefit from Lemma \ref{lem:optim}, one should consider an unbiased expression of ${\cal L}$.
Consider the minibatch scheme that consist in sampling two inputs $X_1$, $X_2$, and $m$ augmentations $\xi_{ij}$ for each $X_i$, formally
\begin{equation}
    \label{eq:sampling}
    X_i \sim \mu_\X^{\otimes 2},\qquad
    \xi_{ij} \sim \mu\vert_{X_i}^{\otimes m}.
\end{equation}
Here $\mu_\X$ denotes the marginal of $\mu$ with respect to $X$, which is likely to match $\rho_\X$, and $\mu\vert_X$ denotes the distribution of $\Xi$ conditionally to $X$.

\begin{lemma}
    An unbiased formulation of ${\cal L}$ is based on $\ell$ defined as
    \begin{align}
        \label{eq:sgd_sample}
        \nonumber
        \nabla_{\Lambda}\ell(S\Theta;\lambda) 
        &= \frac{2(\beta - 1)}{m}\sum_{j\in[m]} \phi(\xi_{1j})\phi(\xi_{1j})^\top - \frac{2\beta}{m(m-1)} \sum_{1\leq j \neq k \leq m} \phi(\xi_{1j})\phi(\xi_{1k})^\top
        \\&\qquad\qquad+ \frac{1}{m^2} \sum_{i,i'=1}^2 \sum_{j, k=1}^m \scap{\Lambda}{\phi(\xi_{ij})\phi(\xi_{i'k})^\top} \phi(\xi_{ij})\phi(\xi_{i'k})^\top.
    \end{align}
    Moreover, when ${\cal L}$ is regularized, one has to add $+\lambda I$ to get a gradient on the regularized risk.
\end{lemma}
\begin{proof}
    This formula follows from Lemma \ref{lem:quadra}.
\end{proof}

In order to bound the norm squared of the gradient, one can use the following lemma.
\begin{lemma}
    For $\ell$ given in \eqref{eq:sgd_sample}, bounds on the gradient norm and its variance are
    \begin{equation}
        \norm{\nabla_\Lambda \ell} \leq 2\kappa^2 + \kappa^4 \sup\norm{\Lambda},
        \qquad\text{and}\qquad
        \E[\norm{\nabla_\Lambda \ell-\nabla {\cal L}}^2] \leq (\sigma_X^2 + m^{-1}\sigma_\xi^2)(1+\sup\norm{\Lambda}^2),
    \end{equation}
    where $\sigma_X$ relates to the variance of $\E\bracket{\psi(\xi)\midvert X}$ and $\sigma_\xi$ relates to the average variance of $\paren{\xi\midvert X}$.
\end{lemma}
\begin{proof}
    Let us decompose $\nabla\ell$ into three terms $\nabla\ell = a+b+c$ as appearing in \eqref{eq:sgd_sample}, we have
    \begin{align*}
        &\norm{a} \leq 2(1-\beta) \norm{\phi(\xi)\phi(\xi)^\top} \leq 2(1-\beta) \kappa^2
        \\&\norm{b} \leq 2 \beta \norm{\phi(\xi)\phi(\xi')^\top}^2 \leq \beta \kappa^2
        \\&\norm{c} \leq \norm{\scap{\Lambda}{\phi(\xi)\phi(\xi')^\top}\phi(\xi)\phi(\xi')^\top} \leq \sup \norm{\Lambda}\kappa^4.
    \end{align*} 
    To bound the variance, one can proceed with
    \[
       \E\norm{\nabla \ell - \nabla{\cal L}}^2\leq 3\E\norm{a - \E[a]}^2 + 3\E\norm{b-\E[b]}^2 + 3\E\norm{c - \E[c]}^2. 
    \]
    Let us begin with the part in $a$,
    \begin{align*}
        &\E\bracket{\norm{\frac{1}{m}\sum_{i\in[m]}\phi(\xi_{1i})\phi(\xi_{1i})^\top - \E[\phi(\xi)\phi(\xi)^\top]}^2}
        = \E\bracket{\norm{\frac{1}{m}\sum_{i\in[m]}\phi(\xi_{1i})\phi(\xi_{1i})^\top - \E[\phi(\xi)\phi(\xi)^\top\midvert X=X_1]}^2}
        \\&\qquad\qquad\qquad\qquad+ \E\bracket{\norm{\E[\phi(\xi)\phi(\xi)^\top\midvert X=X_1] - \E[\phi(\xi)\phi(\xi)^\top]}^2}
        \\&\qquad\qquad= \frac{1}{m} \E_X \E_\xi\bracket{\norm{\phi(\xi)\phi(\xi)^\top - \E[\phi(\xi)\phi(\xi)^\top\midvert X]}^2\midvert X}
        + \E\bracket{\norm{\E[\phi(\xi)\phi(\xi)^\top\midvert X] - \E[\phi(\xi)\phi(\xi)^\top]}^2}.
    \end{align*}
    Similarly, the part in $b$ can be expressed as
    \begin{align*}
        \E[\norm{b - \E[b]}^2]
        &= 2\beta\paren{\frac{2}{m} \E_X \E_\xi\bracket{\norm{\phi(\xi)\phi(\xi')^\top - \E[\phi(\xi)\phi(\xi')^\top\midvert X]}^2\midvert X}}
        \\&\qquad\qquad\qquad\qquad+ 2\beta\E\bracket{\norm{\E[\phi(\xi)\phi(\xi')^\top\midvert X] - \E[\phi(\xi)\phi(\xi')^\top]}^2}.
    \end{align*}
    Finally, 
    \begin{align*}
        \E[\norm{c - \E[c]}^2]
        &= \frac{1}{m^2} \E_X \E_\xi\bracket{\norm{\scap{\Lambda}{\phi(\xi)\phi(\xi')^\top} \phi(\xi)\phi(\xi')^\top - \E[\scap{\Lambda}{\phi(\xi)\phi(\xi')^\top}\phi(\xi)\phi(\xi')^\top\midvert X, X']}^2\midvert X, X'}
        \\&\qquad\qquad+ \E\bracket{\norm{\E\bracket{\scap{\Lambda}{\phi(\xi)\phi(\xi')^\top}\phi(\xi)\phi(\xi')^\top\midvert X, X'} - \E[\scap{\Lambda}{\phi(\xi)\phi(\xi')^\top}\phi(\xi)\phi(\xi')^\top]}^2}
        \\&= \frac{1}{m^2} \E_X \E_\xi\bracket{\norm{\scap{\Lambda}{\phi(\xi)\phi(\xi')^\top \otimes \phi(\xi)\phi(\xi')^\top - \E[\phi(\xi)\phi(\xi')^\top \otimes \phi(\xi)\phi(\xi')^\top\midvert X, X']}}^2\midvert X, X'}
        \\&\qquad\qquad+ \E\bracket{\norm{\scap{\Lambda}{\E\bracket{\phi(\xi)\phi(\xi')^\top\otimes \phi(\xi)\phi(\xi')^\top\midvert X, X'} - \E[\phi(\xi)\phi(\xi')^\top\otimes \phi(\xi)\phi(\xi')^\top]}}^2}.
        \\&\leq \frac{1}{m^2} \norm{\Lambda}^2\E_X \E_\xi\bracket{\norm{\phi(\xi)\phi(\xi')^\top \otimes \phi(\xi)\phi(\xi')^\top - \E[\phi(\xi)\phi(\xi')^\top \otimes \phi(\xi)\phi(\xi')^\top\midvert X, X']}^2\midvert X, X'}
        \\&\qquad\qquad+\norm{\Lambda}^2 \E\bracket{\norm{\E\bracket{\phi(\xi)\phi(\xi')^\top\otimes \phi(\xi)\phi(\xi')^\top\midvert X, X'} - \E[\phi(\xi)\phi(\xi')^\top\otimes \phi(\xi)\phi(\xi')^\top]}^2}.
    \end{align*}
	As a consequence, we get
	\[
	\E\norm{\nabla \ell - \nabla{\cal L}}^2\leq 3\paren{2(1-\beta)\paren{\frac{\sigma_{\xi, 1}^2}{m} + \sigma_{X, 1}^2} + 2\beta\paren{\frac{2\sigma_{\xi, 2}^2}{m} + \sigma_{X, 2}^2} + \sup\norm{\Lambda}^2 \paren{\frac{\sigma_{\xi, 3}^2}{m^2} + \sigma_{X, 3}^2}}
	\]
	where
	\begin{align*}
	  	&\sigma_{\xi, 1}^2 =  \E_X \E_\xi\bracket{\norm{\phi(\xi)\phi(\xi)^\top - \E[\phi(\xi)\phi(\xi)^\top\midvert X]}^2\midvert X}
	  \\& \sigma_{X, 1}^2 = \E\bracket{\norm{\E[\phi(\xi)\phi(\xi)^\top\midvert X] - \E[\phi(\xi)\phi(\xi)^\top]}^2}
	  \\&\sigma_{\xi, 2}^2 = \E_X \E_\xi\bracket{\norm{\phi(\xi)\phi(\xi')^\top - \E[\phi(\xi)\phi(\xi')^\top\midvert X]}^2\midvert X}
      \\&\sigma_{X, 2}^2 = \E\bracket{\norm{\E[\phi(\xi)\phi(\xi')^\top\midvert X] - \E[\phi(\xi)\phi(\xi')^\top]}^2}
	  \\&\sigma_{\xi,3}^2 = \E_X \E_\xi\bracket{\norm{\phi(\xi)\phi(\xi')^\top \otimes \phi(\xi)\phi(\xi')^\top - \E[\phi(\xi)\phi(\xi')^\top \otimes \phi(\xi)\phi(\xi')^\top\midvert X, X']}^2\midvert X, X'}
      \\&\sigma_{X,3}^2 = \E\bracket{\norm{\E\bracket{\phi(\xi)\phi(\xi')^\top\otimes \phi(\xi)\phi(\xi')^\top\midvert X, X'} - \E[\phi(\xi)\phi(\xi')^\top\otimes \phi(\xi)\phi(\xi')^\top]}^2}.
	\end{align*}
    Using the fact that $m^2 \geq m$ and choosing the right $\sigma_X$ and $\sigma_\xi$ leads to the lemma. 
\end{proof}

The following lemma states the convexity properties of ${\cal L}$
\begin{lemma}
    As a function of $\Lambda$, the objective ${\cal L}$ is $\alpha$-smooth with $\alpha = \kappa^4$, where $\kappa$ is a bound on $\norm{\phi}$.
    Moreover, when $\X$ is finite, it is $\alpha'$-strongly, with $\alpha'$ being the square of eigen gap of $K = SS^\top$. 
\end{lemma}
\begin{proof}
    This is a consequence of Lemma \ref{lem:quadra}, ${\cal L}$ is a quadratic function, with the quadratic part being 
    \[
        \E[\scap{\Lambda}{\phi(\xi)\phi(\xi')^\top}^2] = 
        \scap{\Lambda}{\E[\phi(\xi')\phi(\xi')^\top] \otimes \E[\phi(\xi)\phi(\xi)^\top] \Lambda} = 
        \scap{\Lambda}{\Sigma \otimes \Sigma \Lambda}. 
    \]
    In other terms, the hessian of ${\cal L}$ is $\Sigma\otimes \Sigma \in {\cal H}^{\otimes 2}\otimes {\cal H}^{\otimes 2}$. As a consequence,
    \(
        \Sigma\otimes \Sigma \preceq \norm{\Sigma\otimes\Sigma}_{\op} I = \norm{\Sigma}^2_{\op} I \preceq \kappa^4 I.
    \)
    Similarly,
    \(
        \Sigma\otimes \Sigma \succeq \norm{\Sigma^{-1}}^{-2}_{\op} I = \gamma_\xi^2 I,
    \)
    where $\gamma_\xi$ is the eigen gap of $\Sigma$, hence of $K$.
\end{proof}

There are few remaining difficulties that must be addressed before concluding. 
First, although the identity is not Hilbert-Schmidt, it should be noted that the term in $\lambda$ will only contract distances in the stochastic gradient descent. As a consequence, optimizing the regularized risk will only contract the descent trajectory (to prove it formally one could go back to the proofs of \citet{Bubeck2015}).
Finally, we have described a descent in the space of self-adjoint positive operators, without incorporating any constraints on the rank of $\Lambda$.
Notice that based on Lemma \ref{lem:domain}, on can restrict the search of $\Lambda$ to inside the domain $\norm{\Lambda} \leq k_\lambda / \lambda$.
Finally, if $\Lambda$ minimizes the loss ${\cal L}$, then one can show that thresholding its eigenvalues to make it of dimension at most $k$ can only increase the loss ${\cal L}$ by a bounded multiplicative factor.
We note that without explicit regularization, the previously described stochastic gradient descent algorithm with early stopping has a regularization effect that could be studied in the spectral filtering framework of \citet{Lin2020}.

%% file: appendix/examples.tex
\section{Examples}
\label{app:examples}

This section is devoted to illustrate what $T$ and $K$ are under simple distributions thanks to harmonic analysis techniques.

\subsection{Harmonics analysis on the sequence of bits, a.k.a. the Boolean hypercube}

A fine-grained analysis of the role of classical augmentations can be derived in settings that allow pr\'ecise derivations.
We shall focus on invariant data distribution such as the uniform distribution, and augmentations consisting of permutations or perturbations of coordinates that left this distribution invariant.
While such distributions may lack structure present in real data, they allow for a precise study of the effect of certain architectures and augmentations, which may also partly apply to more realistic data.
The study involves the construction of appropriate~$L^2$ bases that ease the study of the effect of both the kernel operator~$K$ and the smoothing operator~$T$ defined from augmentations. 
These are closely related to the study of invariant kernels~\citep[see, e.g.,][]{bietti2021sample,bietti2022approximation,mei2021learning,misiakiewicz2021learning}.

Will focus here on the data that are n-bit inputs on the Boolean cube~$\X = \{-1,+1\}^d$ with uniform distribution.
To be able to use the harmonic analysis tools to their fullest, we assume that inputs are sampled from the uniform distribution on $\X$.
In this setting, the space of function $L^2(\X) = L^2(\X, \R, \mu_\X)$ is defined through the usual scalar product, for $f,g:\X\to\R$,
\[
  \scap{f}{g} = \E_{x\sim\tau}[f(x) g(x)] = \frac{1}{2^d}\sum_{x\in\X} f(x) g(x).
\]

\subsection{The role of augmentation}

Let us know analyze the role of augmentation in the definition of $T$ on the Boolean cube.
For simplicity and ease of notation, we assume indexing of the bits is taken $\mathrm{mod} \; d$, e.g., $x_{-1}=x_d$. 

\subsubsection{Study through parity functions}

\paragraph{Parity functions.}
A useful basis in this space are the parity functions, which can be seen as Fourier functions in this $L^2$-space~\citep{o2014analysis}.
They are defined for each subset $S\subseteq [d]$ as counting the parity of $x$ within this set
\begin{equation}
    \label{eq:parity}
    \chi_S( x) = \prod_{i \in S} x_i.
\end{equation}

\begin{lemma}
  The parity functions $\chi_S$ form an orthonormal basis of $L^2(\X)$.
\end{lemma}
\begin{proof}
  It is straightforwards to check that $\scap{\chi_S}{\chi_S} = 1$.
  If $S \neq S'$, then w.l.o.g.~there is an $i\in S\setminus S'$, and we have
  \[
	\scap{\chi_S}{\chi_{S'}}
	= \E_x[x_{i} \chi_{S\backslash \brace{i}}(x) \chi_{S'}(x)]
	= \E_{x_{i}}[x_{i}] \E_{x_{-i}}[\chi_{S\backslash \brace{i}}(x) \chi_{S'}(x)] = 0.
  \]
  This proves orthogonality of this basis.
\end{proof}

Let us begin with augmentations that are easily to study with the parity basis.

\begin{proposition}[Random noise]
    Consider the flip of each bit of $x$ with probability equal to $p$ formally via the operation
    \begin{equation}
        B^p_y(x) = x \odot y, \; \; \; \; y \sim \operatorname{Ber}( \{-1,+1\}, p)^{\otimes d},
    \end{equation}
    where the operation $x \odot y$ applies pointwise multiplication and the distribution $\operatorname{Ber}( \{-1,+1\}, p)$ returns the value $-1$ with probability $p$ and $+1$ with probability $1-p$.
	Under the augmentations $\xi = X\otimes y$, $T$ is diagonalized in the parity basis with
	\begin{equation}
		T\chi_S = \abs{1 - 2p}^{\card{S}} \chi_S.
	\end{equation}
	In other terms, $T$ applies a factor $|1-2p|^{|S|}$ to reduce the effect of higher order Fourier functions.
\end{proposition}
\begin{proof}
  Recall the formula $g^\top T f = \E_X\E_{\xi, \xi'}\bracket{\scap{f(\xi)}{g(\xi')}\midvert X}$.
  As a consequence, with $y,y'$ denoting the noise strings (each bit equal to $-1$ with probability $p$) and $S\vartriangle S' = (S \cup S')  / (S \cap S')$,
\begin{align*}
\label{eq:effect_of_bitflip_noise}
	\chi_S^\top T \chi_{S'} 
	&= \E_X[ \E_{y,y'} [ \chi_S (X \odot y) \chi_{S'} (X \odot y') ] ]
   = \E_X \bracket{  \E_{y,y'}\bracket{\prod_{i \in S }  X_i y_i \prod_{j \in S'} X_j y_j'}} 
 \\ &= \E_X \left[ \E_y \left[\prod_{i \in S\vartriangle S'} X_i y_i \right]\right] \E_{y,y'}\left[\prod_{i \in S \cap S'}  y_i y_i' \right]
   \\ &=\E[\chi_{S\vartriangle S'}(X)] \cdot |1-2p|^{|S\vartriangle S'|} |1-2p|^{2 | S \cap S'|} 
   = |1-2p|^{|S|} \delta_{S,S'}.
\end{align*}
Therefore, in the case of bit-flip augmentations, $T$ is diagonalized in the parity basis. 
\end{proof}

\begin{proposition}[Cropping/Masking]
    Consider the cropping operation within a window of size $w$, formally defined as
    \begin{equation}
        [M^w_{a}(x)]_i = 
        \begin{cases} 
        x_i & \text{if } i \in [a, a+w) \\ \operatorname{Ber}( \{-1,+1\}, 0.5) & \text{otherwise} 
	\end{cases}, \qquad a \sim \uniform{[d]},
    \end{equation}
    where $[a, a+w) = \{a, a+1, \dots, a+w-1\}$, $a$ is drawn from the uniform distribution over $[d]$, and the distribution $\operatorname{Ber}( \{-1,+1\}, 0.5)$ returns a random bit with equal probability for $+1$ and $-1$ thus effectively masking the values outside of the window in $[a, a+w)$.
	Under the augmentations $\xi = M^w_a(X)$, $T$ is diagonalized in the parity basis with
	\begin{equation}
	  T\chi_S = \frac{\max\brace{1+w-\diam(S), 0}^2}{d^2}\cdot \chi_S \qquad\text{with}\qquad \diam(S) = \min \brace{v\midvert v, a \in [d]; S \subseteq [a, a + v)}.
	\end{equation}
  	In other terms, the action of cropping effectively removes any dependence on the kernel with parity functions of high order whose support falls outside the windows of size $w$.
\end{proposition}
\begin{proof}
  In this setting,
  \begin{align*}
	\chi_S^\top T \chi_{S'} 
	&= \E_X[ \E_{a,b} [ \chi_S (M_a^w(X)) \chi_{S'} (M_b^w(X)) ] ]
  \\&= \frac{1}{d^2}\sum_{a,b=1}^d \E_{X, \nu, \nu'} \left[ \prod_{i \in S \cap [a,a+w)} x_i \prod_{i' \in S \backslash [a,a+w)} \nu_{i'} \prod_{j \in S' \cap [b,b+w)} x_j  \prod_{j' \in S'\backslash [b,b+w)} \nu_{j'}'  \right]
	\\&= \frac{1}{d^2}\sum_{a,b=1}^d \ind{S \subseteq [a,a+w)}\;\ind{S' \subseteq [b,b+w) } \;\mathbb{E}_X[\chi_S(X)\chi_{S'} (X)] 
	= \frac{1}{d^2}\sum_{a,b=1}^d \ind{S \subseteq [a,a+w) }\;\ind{S' \subseteq [b,b+w)} \;\delta_{S,S'}
	\\&= \paren{\frac{1}{d}\sum_{a=1}^d \ind{S \subseteq [a,a+w) }}^2 \;\delta_{S, S'}.
  \end{align*}
  The count of the sum relates to the diameter of $S$.
\end{proof}

\begin{proposition}[2D Cropping]
	Consider that 2D setting $\X = \brace{-1,+1}^{m \times d}$ where inputs are organized into an $m \times d$ grid. 
	Consider the cropping operation to a window of size $v \times w$, formally 
    \begin{equation}
        [M^{v \times w}_{a,b}(x)]_{i+jm} = 
        \begin{cases} 
        x_{i+jm} & \text{if } i \in [a, a+v), j \in [b, b+w) \\ \operatorname{Ber}( \{-1,+1\}, 0.5) & \text{otherwise} 
	  \end{cases}, \qquad (a, b) \sim \uniform{[m]\times [d]}.
    \end{equation}
	Under the augmentation $\xi = M^{v\times w}_{a,b}(X)$, $T$ is diagonalizable in the parity basis and
	\begin{equation}
	  T\chi_S = \frac{1}{m^2d^2}\paren{1 + v - \diam_{e_1}{S}}_+^2\cdot \paren{1 + w - \diam_{e_2}{S}}_+^2 \chi_S,
	\end{equation}
	where $\diam_{e_1}{S}$ is the diameter of $S$ projected onto the first dimension.
\end{proposition}
\begin{proof}
	This follows from the proof of the 1D case.
\end{proof}

\begin{proposition}[Flipping]
    Consider the operator which, with probability $p$, flip the indices into reverse order, formally
    \begin{equation}
        [R(x)]_i = x_{-i}.
    \end{equation}
	Under the augmentation $\xi = R(X)$, 
	\begin{equation}
	  T = (1-2p + 2p^2) I + 2p(1-p) J,
	 \end{equation}
	 where $J$ is the involution that matches any set $S$ to its mirror $\tilde{S} = \brace{-i\midvert i\in S}$.
      In this setting, $T$ is diagonalized by the $\sqrt{2}(\chi_S + \chi_{\tilde{S}})$ and $\sqrt{2})(\chi_S - \chi_{\tilde S})$ for $S\subseteq [d]$.
\end{proposition}
\begin{proof}
  In this setting,
  \begin{align*}
	\chi_S^\top T\chi_{S'}
	&= ((1-p)^2 + p^2) \E_X\left[  \chi_S (X) \chi_{S'} (X) \right] + 2p(1-p) \E_X\left[  \chi_{\tilde{S}}(X) \chi_{S'} (X) \right] 
	\\& = (1-2p+2p^2) \delta_{S,S'} + 2p(1-p)\delta_{\tilde{S},S'},
   \end{align*}
  which explain the lemma.
\end{proof}

\begin{remark}
  Up to now, we have studied all the operators in the space $L^2(\X,\R,\mu_X)$ while the main text considered those operators in $L^2(\X, \R,\mu_\Xi)$, this is justified by the fact that all transformations studied earlier let invariant the uniform distribution, hence 
  \begin{equation}
	L^2(\mu_\X) = L^2(\mu_\Xi).
  \end{equation}
\end{remark}

\subsubsection{Study of translations through cyclic parities}
In order to study augmentations that consist of permutations, and more specifically translations, the parity basis is not adapted to diagonalize~$T$.
Instead we define below a different basis that incorporates cyclic symmetries~\citep{misiakiewicz2021learning}.
We note that a similar study may be carried on other distributions, e.g., uniform on the sphere, product of spheres, or torus~\citep{bietti2021sample,bietti2022approximation,favero2021locality,mei2021learning}.

\paragraph{Cyclic parity functions.}
The functions~$\chi_S$ are polynomials that can be grouped by their degree~$\ell = \card{S}$ into spaces~$V_{d,\ell}$, whose direct sum yields the full~$L^2(\X)$ space, with
\[
\dim V_{d,\ell} = \card{\{S \subseteq [d], \card{S} = \ell\}} = {d \choose \ell}.
\]
Those different spaces can be further decomposed into orbits under the action of a group.
In particular for the group of permutations $G = \Sfrak_d$, we define the action ${\cal A}:G\times\X\to\X$ denoted ${\cal A}(a, x) = a\cdot x$ as
\[
    (a \cdot x)_i = x_{a^{-1}(i)}.
\]
To give a concrete example of the study of augmentations through harmonic analysis, let us focus more specifically on the action of translation, which form a sub-group of permutations. 
For simplicity, we will denote this group $[d]$ which is understood as $\Z / d\Z$, acting on $\X$ as 
\[
    (a\cdot x)_i = x_{i-a}
\]
where $i-a$ being understood modulo $d$.
Define the orbits of this action as $\brace{S + a\midvert a\in[d]}$ for $S\subseteq [d]$.
On those different orbits, one can define the following ``cyclic parities'' $\psi_{m,S}: \X \to \C$:
\begin{equation}
    \label{eq:spherical}
    \psi_{m,S} = \frac{1}{\sqrt{k_S}}\sum_{k\in[k_S]} e^{2i\pi k m / k_S} \chi_{S+k}
    = \frac{\sqrt{k_S}}{d}\sum_{k\in[d]} e^{2i\pi k \frac{m d}{k_S} / d} \chi_{S+k}
    \qquad\text{where}\qquad
    k_S = \card{\orb(S)},
\end{equation}
where $m \in[k_S]$ and $S$ is taken as a representant of an orbit.

\begin{lemma}
    \label{lem:diag_trans}
    The cyclic parities $(\psi_{m, S})$, for $m \in [k_S]$ and $S$ is a set of representers of each orbit of the translations action, form an orthogonal basis of $L^2(\X, \C, \mu)$ where $\mu$ is the uniform measure on $\X$.
    Moreover, they diagonalize the operators $A:L^2\to L^2$ defined as $Af(x) = f(a\cdot x)$ for any $a\in [d]$.
\end{lemma}
\begin{proof}
    The first part follows from the fact that $L^2(\X)$ can be decomposed into the direct sum linked with the $V_{d, \ell}$ for $\ell \in [0, d]$, that each subspace can be decomposed into the orbits of the action translation $\orb(S) = \brace{S + a\midvert a\in[d]}$ (note that translation do not change the cardinals of the sets $S$).
    Those latter spaces can be parameterized through the discrete Fourier transform, yielding the $\psi_{m, S}$.
    
    A natural way to ``find'' those basis is when trying to diagonalize an operator $T$ such that $(\chi_S^\top T \chi_{S'})_{S,S'\subseteq [d]}$ that is block diagonal, where each block corresponds to a circulant matrix on an orbit, which can be diagonalized with the discrete Fourier transform.
    This is especially the case for operator of the lemma
  \begin{align*}
        A\chi_{S} = \chi_{S+a}
  \end{align*}
  The above is only nonzero when $[d]\cdot S$ intersects $[d]\cdot S'$, which implies $\orb(S) = \orb(S')$ thereby constructing a block diagonal structure. 
  Indexing the elements of the $i$-th block by $S_{i,k} = S_i + k$ for $k \in [d]$, we have
  \[
      \chi_{S_{i,k}}^\top A\chi_{S_{i,k'}} = \ind{S_{i,k} = S_{i,k'} + a},
      = \ind{S_{i} + k = S_{i} + k' + a},
      = \ind{k - k' = a},
  \]
   which only depends on the value of $(k-k')$. 
   Therefore, each block above is a circulant matrix which is diagonalized by the discrete Fourier transform. 
   The eigenvectors of this matrix are
    \[
        v_m = \frac{1}{\sqrt{k_S}}\sum_{k\in[k_S]} e^{2i\pi k m / k_S} e_k,
        \qquad\text{where}\qquad
        k_S = \card{\orb(S)},
    \]
    for $m\in[k_S]$ and the corresponding eigenvalues read
    \[
        \mu_m = \sum_{k\in[k_S]}  c_{k_S-k} \exp\paren{\frac{2i\pi k m}{k_S}},
    \]
    where
    \[
        c_i = \ind{i = a},
    \]
    Using the fact that we wrote those matrices for $e_i\simeq \chi_{S+i}$ yields the lemma.
\end{proof}

The study of the operator~$T$ can be simplified thanks to its square root $A:L^2(\mu_\Xi)\mapsto L^2(\mu_\X)$ formally defined by
\begin{equation}
    Af(x) = \E_{\xi}\bracket{f(\xi)\midvert X=x}
\end{equation}
and verifying
\begin{equation}
    \scap{f}{T g}_{L^2(\mu_\Xi)} = \E_X\E_{\xi, \xi'}[ f(\xi) g(\xi') | X] = \langle Af, Ag \rangle_{L^2(\mu_{\mathcal X})}.
\end{equation}
This decomposition will be particularly useful, when $\mu_\X$ is invariant under the action of permutations, which implies $\mu_\X = \mu_\Xi =: \mu$.

\begin{lemma}
    In the uniform Boolean setting, when augmentations are defined as $\xi = a\cdot X$ where $a$ is a permutation sampled from the probability distribution $p\in\prob{\Sfrak_d}$,
    \[
        Tf(x) = \sum_{a,b \in \Sfrak_d} p(a) p(b) f((a^{-1}b) \cdot x).
    \]
\end{lemma}
\begin{proof}
    The square root of $T$ is defined as $Af(x) = \sum_{a\in\Sfrak_d} p(a)f(a\cdot x)$.
    Let us focus on the case where $p(b) = \delta_{a=b}$, using the fact that $\mu_\X$ is the uniform measure, hence is left invariant by translation, we compute the adjoint of $A$ with
    \begin{align*}
        \scap{Af}{g}_{L^2(\mu_\X)} 
        &= \frac{1}{2^d}\sum_{x\in\X} Af(x) g(x) 
        = \frac{1}{2^d}\sum_{x\in\X} p(a) f(a\cdot x) g(x)    
        \\&= \frac{1}{2^d}\sum_{x\in\X} p(a) f(a\cdot x) g(a^{-1} \cdot a\cdot x)    
        = \frac{1}{2^d}\sum_{x\in\X} p(a) f(x) g(a^{-1} \cdot x)    
       \\& = \scap{f}{x\mapsto p(a)g(a^{-1} \cdot x)}_{L^2(\mu_\X)}.    
       \\& = \scap{f}{A^\top g}_{L^2(\mu_\Xi)}.    
    \end{align*}
    In the general case, we get by linearity,
    \[
        A^\top f(x) = \sum_{a\in \Sfrak_d} p(a) f(a^{-1}\cdot x).
    \]
    Computing $T = A^\top A$ leads to the result.
    Remark that if we further assume that~$p$ is symmetric (i.e., $p(a) = p(a^{-1})$), then we have~$A^\top = A$, so that~$T = A^2$.
\end{proof}

This allows us to characterize more finely the effect of translation on the operator $T$.

\begin{proposition}[Translations]
    Consider the translation operator defined formally as 
    \begin{equation}
	[T_a(x)]_i = x_{i - a}, \qquad a \sim p \in \prob{[d]}
    \end{equation}
    Under the augmentation $\xi = T_a(X)$, $T$ is diagonalized in $\C$ by the cyclic parity functions~\eqref{eq:spherical}.
    \begin{equation}
        T\psi_{m, S} = \frac{d^2}{k_S^2} \abs{\hat{p}\paren{\frac{m d}{k_S}}}^2 \psi_{m, S},
    \end{equation}
    where $\hat{p}$ is the Fourier transform of $p$, defined for~$\omega \in [d]$ by
    \begin{equation}
        \hat{p}(\omega) = \sum_{a\in[d]} p(a) \exp\paren{\frac{-2i\pi a\omega }{d}}
    \end{equation}
\end{proposition}
\begin{proof}
    In the case of translation, we have
    \[
        Af(x) = \sum_{a\in[d]} p(a)f(a\cdot x)
        = \sum_{a\in[d]} p(a) A_a f(x),
    \]
    were $A_k$ be the operator that associate $f$ to $x\to f(k\cdot x)$, it is a translation operator and retaking the proof of Lemma \ref{lem:diag_trans}, $A_k \psi_{m, S} = e^{-2i\pi km / k_S}\psi_{m,S}$.
    This leads to 
    \[
        A \psi_{m, S}
        = \sum_{a\in[k_S]} p(a) \exp\paren{\frac{-2i\pi a m}{k_S}} \psi_{m, S}
        = \sum_{a\in[d]} \frac{d}{k_S} \cdot p(a) \exp\paren{\frac{-2i\pi a m}{k_S}} \psi_{m, S}
        = \frac{d}{k_S} \cdot \hat{p}\paren{\frac{m d}{k_S}} \psi_{m, S},
    \]
    and
    \begin{align*}
        T = A^*A = \sum_{m, S} \frac{d^2}{k_S^2} \abs{\hat{p}\paren{\frac{m d}{k_S}}}^2 \psi_{m, S} \psi_{m, S}^*.
    \end{align*}
    This proves the lemma.
\end{proof}

We now show how different sampling distributions over translations induce varying smoothing effects in the operator~$T$.

\begin{example}[Smoothing effect of translations]
\label{ex:translation_smoothing}
    To see the effect of augmentation strength, consider a distribution~$p$ over translations that takes the form~$p(a) = \omega p_0(\omega a)$, where~$p_0$ is a localized window shape (e.g., uniform or Gaussian) that sums to~$1$. Here~$\omega \approx 1/\Delta$ is inversely related to the window size~$\Delta$, which controls the ``strength'' or range of augmentations.
    Then we have
    \[|\hat p(m)|^2 = |\hat p_0(m / \omega)|^2 \approx |\hat p_0(\Delta m)|^2. \]
    Here, the squared Fourier coefficients~$|\hat p_0(m)|^2$ typically decay with the frequency~$m$, which shows that~$T$ has a smoothing effect that penalizes eigenfunctions~$\psi_{m,S}$ with larger~$m$, i.e., those which oscillate more quickly.
    The above formula also highlights that the increasing the augmentation strength~$\Delta$ will lead to faster decay with~$m$, while leaving the translation-invariant eigenfunctions ($m = 0$) unaffected.
\end{example}

\subsection{The role of architectures}

A particularly useful feature space $\phi$ to define the linear class of functions $\Psi$ is the set:
\begin{equation}
  \label{eq:bool_features}
  \phi:\X\mapsto\R^{2^d}; x\mapsto(e_S\chi_S(x))_{S\subseteq[d]}.
\end{equation}
for any sequences $(e_S)\in\R^{2^d}$.
Linear model of this form can be diagonalized in the parity basis, which allows one to effectively study the interplay between the role of augmentation and the role of the architecture.

\begin{lemma}
    \label{lem:dot-prod-eigval}
  For any linear model defined through the features $\phi$ in \eqref{eq:bool_features}, the integral operator $K:L^2(\X)\mapsto L^2(\X)$ is diagonalized in the parity basis,
  \begin{equation}
	K\chi_S = e_S^2 \chi_S.
  \end{equation}
\end{lemma}
\begin{proof} 
  This follows from the fact that $Kf(x) = d^{-1}\sum_{x'\in[d]} k(x,x') f(x')$ where $k(x,x') = \phi(x)^\top \phi(x')$.
\end{proof}

Among those classes of functions are dot-product kernel that verifies $k(x, y) := \phi(x)^\top \phi(y) = \tilde{h}(\norm{x - y}^2) = h(x^\top y)$.
Once again, those kernels are particularly well adapted to the Fourier geometry of the Boolean hypercube.

\begin{lemma}[Spectral decomposition of dot-product kernel]
    \label{lem:strong_diag}
    Any dot-product kernel is diagonalizable in the parity basis. 
    Specifically, there exists $(\nu_i)_{i\in[0,d]} \in \R^{d+1}$ such that, when $\mu_\X$ is the uniform distribution on the hypercube,
    \begin{equation}
        K\chi_S = \nu_{\card{S}} \chi_S.
    \end{equation}
\end{lemma}
\begin{proof}
    One can check that $x^\top y = d - 2k$ for $k$ the number of bits that differs in $x$ and $y$.
    Define $Q_\ell$ the degree-$\ell$ averaged polynomials of degree~$\ell$ as
    \begin{equation}\label{eq:gegenbauer_boolean}
        \sum_{S \subseteq [d], |S| = \ell} \chi_S(x) \chi_S(y) = {d \choose \ell} Q_{\ell, d}(\scap{x}{y}),
    \end{equation}
    for any Boolean strings $x$ and $y$. 
    The $Q_{\ell, d}$ are well defined since the left-hand side is translation invariant.
    Moreover, leveraging the orthogonality of the $\chi_S$, one can show that the $(Q_{\ell, d})_{\ell\in[0, d]}$ form a basis of functions on $\brace{d-2k\midvert k\in[0, d]}$.
    More exactly, the $m\to {d \choose \ell}^{-1/2} Q_{\ell,d}(m)$ are orthonormal basis of the $L^2$ space endowed with $\tau$ the pushforwards measure of the uniform distribution on $\X$ through the mapping $x\to\scap{x}{y}$ for any fixed $y$, and the dimensions match.
    As a consequence, there exists $\nu_\ell$ such that
    \[
        h(\scap{x}{y}) = \sum_{\ell\in[0,d]} \nu_\ell {d \choose \ell} Q_{\ell, d}(\scap{x}{y})
    \]
    where $\nu_\ell$ can be found by computing the scalar product between $h$ and $Q_\ell$ in $L^2(\tau)$.
    \begin{equation} \label{eq:gg_coeff}
        \nu_\ell = \scap{h}{Q_\ell}_{L^2(\tau)}.
    \end{equation}
    Finally, using the fact that, in the uniform setting, $Kf(x) = d^{-1}\sum_{x'\in[d]} k(x,x') f(x')$ where $k(x,x') = \phi(x)^\top \phi(x')$, we have
    \[
        K\chi_S(x) = \E_Y[h(\scap{x}{Y})\chi_S(Y)] = \sum_{\ell} \nu_\ell \sum_{S'\subset[d], \card{S'} = \ell} \chi_{S'}(x) \E[\chi_{S'}(Y) \chi_S(Y)] = \nu_\ell \chi_S(x).
    \]
    This ends the proof of this lemma.
\end{proof}

Lemma \ref{lem:dot-prod-eigval} can also be shown on the sphere.
Its proof showcase the $Q_\ell$ which act as normalized Legendre (or Gegenbauer) polynomials. See, \emph{e.g.}~\citet{smola2000regularization,bietti2021sample,mei2021learning} for details.
Note that for common kernel functions on the sphere, such as the ones appearing in the NTK, the~$\nu_k$ decay polynomially with~$k$~\citep{bach2017breaking,bietti2019inductive}. 

The features \eqref{eq:bool_features} are rich enough to describe the neural tangent kernels of simple architectures with fully connected or convolutional layers. First, we describe the general form of such NTKs as below.

\begin{proposition}[Linearization of simple network]
    Define a simple neural architecture as
    \begin{equation} \label{eq:simple_architecture_eqn}
        f(x) = \sqrt{\frac{\Delta}{N\omega d}} \sum_{i\in[N]} \sum_{k\in[d/\Delta]} a_{ik} \sum_{s\in[\omega]} \sigma\paren{\scap{w_i}{x_{(k\Delta + s)}^{(q)}}},
    \end{equation}
    where $x_{(k)}^{(q)} = (x_k, x_{k+1}, \cdots, x_{k+q-1})$ is a local patch of size~$q$ (with indices being defined modulo $d$), $w_i$ the weights initialized from a rotation-invariant distribution~$\cal W$, $\sigma:\R\mapsto\R$ is an activation function, $\omega\in\N$ is the size of the average pooling window, $\Delta\in\N$ is the pooling window, and $N$ is the channel number.
    The linearization of this network near initialization yields the kernel
    \begin{equation}
        k(x, x') = \phi(x)^\top \phi(x') = \frac{\Delta}{d\omega} \sum_{k\in[d/\Delta]} \sum_{s,s'\in [\omega]} h\paren{\scap{x_{(k\Delta + s)}^{(q)}}{y_{(k\Delta + s')}^{(q)}} / q}
    \end{equation}
    where 
    \begin{equation}
        \label{eq:boolean_kernel}
        h(\scap{u}{v} / q) = \E_{w\sim\cal W}\bracket{\sigma(\scap{u}{w} / \sqrt{q}) \sigma(\scap{v}{w} / \sqrt{q}) + \sigma'(\scap{u}{v} / \sqrt{q})\sigma'(\scap{u}{v} / \sqrt{q})\cdot\scap{u}{v} / q}.
    \end{equation}
\end{proposition}
\begin{proof}
  Such a linearization can be found, e.g., in Proposition 3 of \citet{misiakiewicz2021learning}.
\end{proof}

\begin{proposition}[Linearization of a fully connected network]
 \label{prop:eigbasis_FC}
    A one hidden layer fully connected layer
    \[
        f_{FC}(x) = \frac{1}{\sqrt{N}}\sum_{i\in[N]} a_{ik} \sigma(w_i^\top x),
    \]
    can be linearized as a dot-product kernel with $k_{FC}(x, y) = h(x^\top y / d)$ for $h$ defined in \eqref{eq:boolean_kernel}.
    Moreover, the resulting integral operator $K_{FC}$ is diagonalized in the parity basis as
    \[
        K_{FC}\chi_S = \nu_{h}(d,\card{S}) \chi_S,
    \]
    where the coefficients are given by $\nu_h(d,\ell) = \scap{h}{Q_\ell}_{L^2(\tau)}$ as in \eqref{eq:gg_coeff}.
    
    Note that eigenvalues~$\nu_h(d,\ell)$ are non-increasing with~$\ell$, and for fixed~$\ell$ and large~$d$ they satisfy~$\nu_h(d, \ell) = \Theta_d(d^{-\ell})$. More generally, it can be shown that $\lim_{d \to \infty} d^k \nu_h(d,\ell) = \frac{d^k}{dt^k}h(t) \bigr\rvert_{t=0}$.
\end{proposition}
\begin{proof}
    The first part is a direct consequence of the prior proposition with $\omega = 1$ and $q=\Delta=d$.
    The second part is due to Lemma \ref{lem:strong_diag}, and \eqref{eq:gegenbauer_boolean}.
    For the statements on eigenvalues, see~\citep{yang2019fine}.
\end{proof}

\begin{proposition}[Linearization of a convolutional network]
    A convolutional layer followed by a fully connected layer
    \[
        f_{CNN}(x) = \frac{1}{\sqrt{N d}} \sum_{i\in[N]} \sum_{k\in[d]} a_{ik} \sigma\paren{w_i^\top x_{(k)}^{(q)}},
    \]
    can be linearized with the $h$ of \eqref{eq:boolean_kernel} as
    \[
        k_{CNN}(x, y) = \frac{1}{d} \sum_{k\in[d]} h\paren{\scap{x_{(k)}^{(q)}}{y_{(k)}^{(q)}}/ q}.
    \]
    In the Boolean setting, the resulting integral operator $K_{CNN}$ is diagonalized in both the parity and the cyclic basis as
    \[
        K_{CNN}\psi_{m, S} = \begin{cases}
            \nu_h(q,\card{S}) \frac{(q+1-\diam(S))_+}{q} \psi_{m, S},  &\text{if }\diam(S) \leq q, \\
            0 &\text{otherwise.}
        \end{cases}
    \]
    where $\nu_h(q,\ell)$ are defined by Proposition \ref{prop:eigbasis_FC}.
\end{proposition}
\begin{proof}
    The first part corresponds to the case $\omega=\Delta=1$.
    The second part is due to the expansion of $h$ over the $\Q_\ell$ basis, which leads to (see Eq. (30) in \citet{misiakiewicz2021learning} for details)
    \[
        k_{CNN}(x,y) = \sum_{S \subseteq [d], \diam(S) \leq q} q^{-1} \nu_h(q,\card{S}) (q+1-\diam(S))_+ \chi_S(x) \chi_S(y).
    \]
    The fact that $K$ let the $\brace{S\midvert \card{S}=a, \diam(S)=b}$ invariant, since the eigenvalues only depends on $\card{S}$ and $\diam(S)$, allows to change from the parity basis to the cyclic basis.
\end{proof}

When pooling is included in the kernel and $\omega > 1$ in \eqref{eq:simple_architecture_eqn}, then the architecture enforces local translation invariance. 
As a simple example, consider the setting of global average pooling $\omega=d$ where strict invariance to translations is enforced and parity functions are projected onto their sum of elements of the orbit to form the eigenbasis. In this case, $K$ is no longer diagonalized in the parity basis, but it is diagonal in the basis of cyclic parities.

\subsection{Interplay between augmentations and architecture} 

In the uniform Boolean setting, the interplay between augmentations and architecture is made easy by the fact that many operators $K$ and $T$ commutes.

\begin{lemma}
    The operator $K$ associated with a dot-product kernel in the uniform Boolean setting commutes with all the operators $T$ that can be built from bitwise noise, cropping, translations or index flip.
\end{lemma}
\begin{proof}
    In the case of a dot-product kernel in the uniform setting, the spaces $V_{d,\ell}$ are eigenspaces of $K$.
    Those spaces are left invariant by all the $T$ defined through usual augmentations, since translations and index-flip operations preserve the cardinality of subsets.
    As a consequence, $K$ and $T$ can be diagonalized in the same basis, hence they commute.
\end{proof}

As a consequence of the previous lemma, the integral operator $K$ associated with the linear model of fully connected layer commute with all the operators $T$ defined for usual augmentations.
It is also the case for the convolutional layer with $T$ deriving from random noise, cropping, or translation.\footnote{Since it lets invariant the orbit of translation.}
As a consequence, the interplay between the architecture and the augmentations can be studied easily thanks to Proposition \ref{prop:commute}.

\begin{example}[Interplay between FC kernel and translation augmentations] \label{ex:interplay_FC_translation}
Recall from Example~\ref{ex:translation_smoothing} that when sampling translations from a localized window, the eigenvalues of~$T$ are of the form~$|\hat p(m)|^2$ and typically decay
with the frequency index $m$ in $\psi_{m,S} = \frac{1}{\sqrt{k_S}}\sum_{k\in[k_S]} e^{2i\pi k m / k_S} \chi_{S+k}$ for any set~$S$ with no periodicity. 
In contrast, the eigenvalues $\nu_h(d,|S|)$ of K for eigenfunctions $\psi_{m,S}$ decay as $\Theta_d(d^{-|S|})$, independently of $m$. Regularization with parameter $\lambda$ thus shrinks the eigenvalues to $|\hat{p}(m)|^2 - \lambda \nu_h(d,|S|)^{-1}$ after pre-training.
This most notably eliminates contributions from eigenfunctions $\psi_{m,S}$ where $m$ is small (i.e., near-invariant) but $|S|$ is large.
See Figures~\ref{fig:degree_vs_invariance} and~\ref{fig:downstream_error} for an illustration.
\end{example}

\begin{example}[Interplay between kernel for CNN and translation augmentations]
    Consider the setting as before in Example \ref{ex:interplay_FC_translation} with translations sampled from a localized window. For a single layer CNN with patch width $q$, eigenfunctions correspond to parity functions $\chi_S$, or cyclic parities~$\psi_{m,S}$ where $\diam(S) \leq q$ with corresponding eigenvalue $\nu_h(q,\ell)\frac{ q+1-\diam(S)}{d}$. Here, the eigenfunctions $\psi_{m,S}$ of $T$ for $S$ with diameter larger than $q$ are completely eliminated, regardless of the regularization strength~$\lambda$, . For eigenfunctions $\psi_{m,S}$ where $\diam(S) \leq q$, the CNN shrinks the contribution to $|\hat{p}(m)|^2 - \lambda (\nu_h(q,\ell)\frac{ q+1-\diam(S)}{d})^{-1}$, which shrinks more when $\diam(S)$ is larger.
\end{example}



\begin{figure}[ht]
  \centering
  \includegraphics{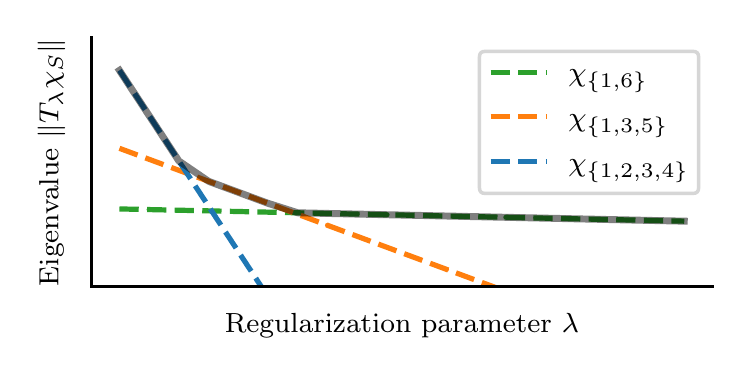}
  \caption{\emph{Illustration of the interplay} between $T$ and $K$ as a function of $\lambda$ where $K$ is the NTK of a 2-layer ReLU network and $T$ performs crops of window size $8$ on $12$-bit inputs. 
  Here we plot eigenvalues of three different parity functions in the eigenbasis of both operators. Parity functions which large diameters have smaller eigenvalues for $T$ (here, the parity function with largest diameter is $\chi_{\{1,6\}}(X) = X_1 X_6$). Eigenvalues of $K$, in contrast, bias towards parities supported over fewer bits. Therefore, small regularization biases towards parities with small diameter whereas added regularization penalizes parities with high cardinality.}
  \label{fig:interplay_crop}
\end{figure}

\begin{figure}
    \centering
    \includegraphics{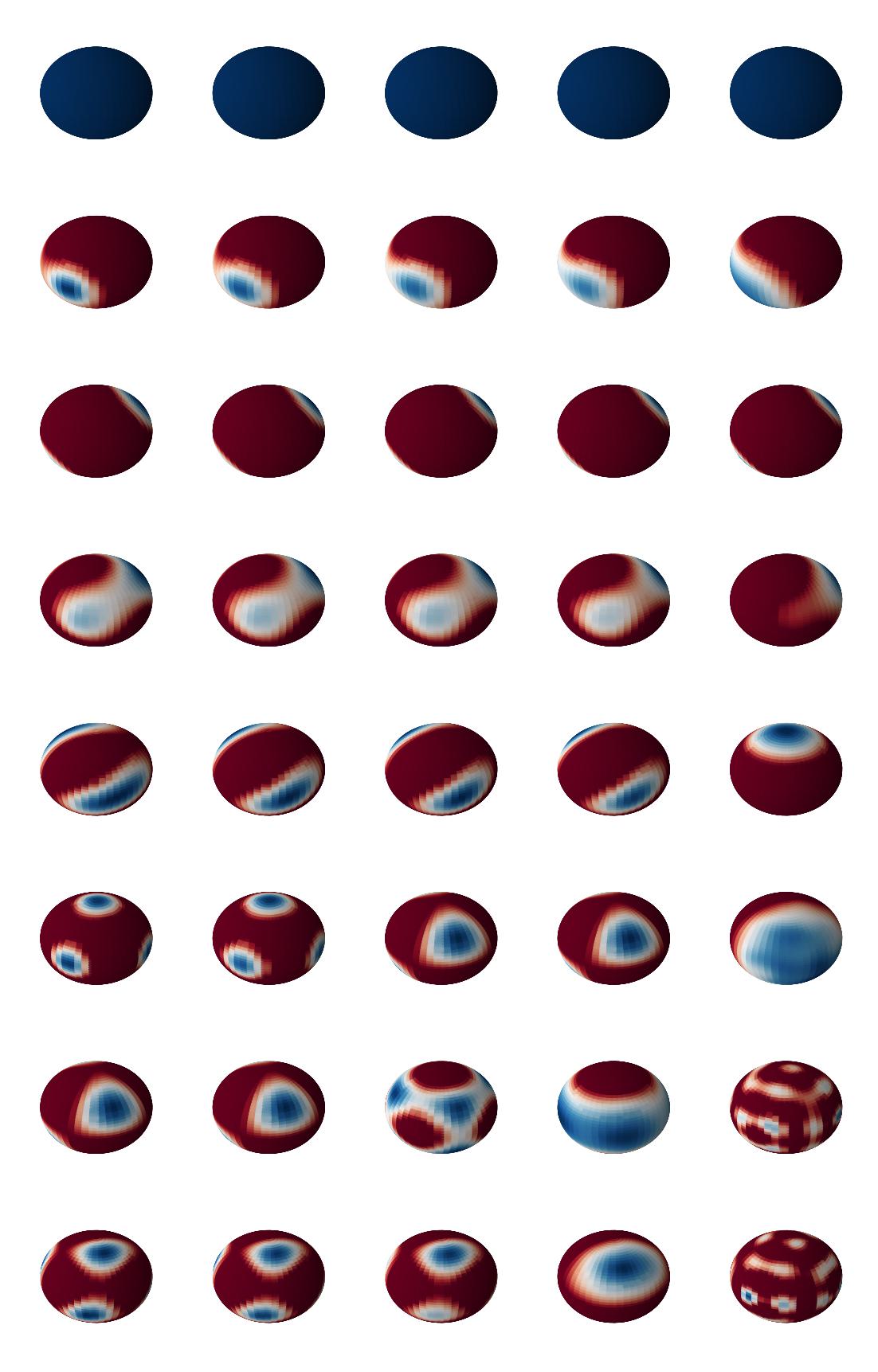}
    \caption{
    {\em Extending Figure \ref{fig:sphere_min}.} 
    The $i$-th row representing the $i$-th eigenfunctions of $T_\lambda$ (ordered by decreasing eigenvalues). Regularization $\lambda$ increases over the columns as $\lambda\in\brace{0, .1, 1, 10, 100}$.
    Small $\lambda$ biases towards functions invariant to the translation augmentation chosen here whereas large $\lambda$ biases towards smoother functions on the sphere corresponding to low order spherical harmonics in this setting.
    The last two on the right are artifacts of the instability of the pseudo-inverse for $K$ (leading to the implementation $\phi K^{-1} \phi = 0$ while we have defined $\phi^\top K^{-1} \phi = +\infty$ when $K\phi = 0$). 
    }
    \label{fig:spherical_interplay}
\end{figure}

\subsection{Remark on the Sphere setup}
\label{app:spherical_setup}
In experiments, we also consider a setup with uniform data on the sphere~$\X = \mathbb S^{d-1}$, with augmentations consisting of permutations, and a dot-product kernel~$\phi(x)^\top \phi(y) = h(x^\top y)$.
A natural choice of basis functions for~$L^2(\X)$ in this case are \emph{spherical harmonics}~\citep{efthimiou2014spherical}.
These consist of homogeneous harmonic polynomials, and similar to the parity case, these can be grouped by degree, leading to orthogonal spaces~$V_{d,\ell}$ of spherical harmonics of any degree~$\ell \geq 0$, with
\[N(d,\ell) := \dim V_{d,\ell} = \frac{2\ell + d - 2}{\ell}{\ell + d -3 \choose d - 2}.\]
It is well-known that for dot-product kernels,~$K$ is diagonal in such a basis~\citep{smola2000regularization,bach2017breaking}, with decaying eigenvalues that only depend on the degree~$\ell$.
These are given analogously to the hypercube setting by
\[
\nu_h(d,\ell) = \E_{t \sim \tau}[h(t) Q_{\ell,d}(t)],
\]
where~$Q_{\ell,d}$ are now Legendre (Gegenbauer) polynomials of degree~$\ell$ orthogonal w.r.t.~a different measure~$d \tau(t) = (1 - t^2)^{\frac{d-3}{2}} dt$ over~$[-1, 1]$.

Since the spaces~$V_{d,\ell}$ are left stable by the operator~$T = A^\top A$, it is possible to show that there exists a choice of spherical harmonics that also diagonalizes~$T$~\citep[see, e.g.,][Lemma 12]{bietti2021sample}.
We may then see the eigenvalues~$\lambda_{\ell,j}$ of~$T$ in this basis as capturing the invariance of the corresponding harmonic~$Y_{\ell,j}$, in particular~$Y_{\ell,j}$ is invariant to all augmentations when~$\lambda_{\ell,j} = 1$, and non-invariant or only partially invariant when~$\lambda_{\ell,j} < 1$.

Ordering~$\lambda_{k,j}$ at fixed~$k$ by decreasing~$j$, the interplay between~$T$ and~$K$ then resembles the one described, e.g., in Figure~\ref{fig:degree_vs_invariance}.

%% file: appendix/experiments.tex
\section{Experiments}
\label{app:experiments}

\subsection{Implementation details}
Previously, we extensively studied the embedding of ${\cal H}$ in $L^2$ defined as 
$S:{\cal H}\mapsto L^2; \theta \to \phi(\cdot)^\top \theta$.
Given samples $(\xi_{ij})_{i\leq n, j\leq m}$, all the action on ${\cal H}$ can be reduced to the span of the $\phi(\xi_{ij})$ (which is known as the representer theorem), and $S$ can be reduced to the embedding $\hat{S}:{\cal H}\mapsto\R^{nm}; \theta\to (\frac{1}{nm}\phi(X_{ij})^\top \theta)_{ij}$.
This leads to the implementation
\[
    \hat{T_\lambda} = \beta + (1-\beta) \hat{T} - \lambda \hat{K}.
\]
Where, $\hat T \in \R^{nm}$ is the matrix equal to the following where we index elements in $\R^{nm}$ by $ij$ with $i\in[n]$ and $j\in[m]$,
\[
    \hat{T} = I + \sum_{ijk} e_{ij} e_{ik}^\top,
\]
and $K$ is the Gram matrix defined as
\[
    nm \cdot e_{ij}^\top K e_{kl} = k(\xi_{ij}, \xi_{kl}) = \phi(\xi_{ij})^\top\phi(\xi_{kl}).
\]
Note the the matrix $\hat{T} - I$ can be seen as the adjacency matrix of the graph that connects augmentations if and only if they come from the same input. Equivalently, $\hat{T}$ can be seen as a Laplacian matrix.
An eigenvector of $\hat{T}_\lambda$ in $\R^{nm}$ is projected back onto $L^2$ thanks to $S\hat{S}^{-1} = S\hat{S} (\hat{S}\hat{S}^\top)^{-1} = K_x^\top K^{-1}$ where
\[
    nm K_x = \paren{\phi(x)^\top\phi(\xi_{ij})}_{ij} \in \R^{nm}.
\]

\begin{figure}[ht]
  \centering
  \includegraphics{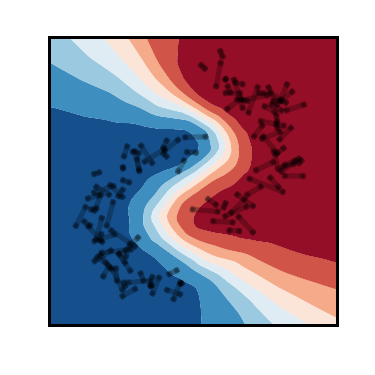}
  \includegraphics{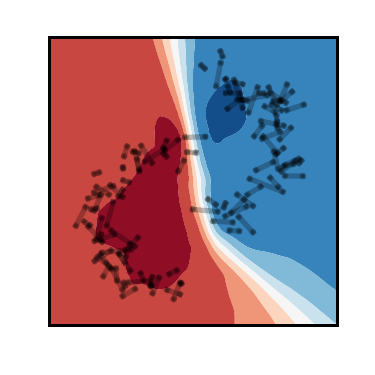}
  \vspace{-2em}
  \caption{\emph{VCReg with Neural networks}. 
  Contour plots of the minimizer $\psi:\X\to\R$ of ${\cal L}$ for $\beta=1$ (left) and $\beta=0$ (right) with a two layer fully connected neural network when $k=1$, $\X=\R^2$, $X$ is distributed according to a half-moon structure and $\xi = X + \epsilon$ for a small noise $\epsilon$.
  Augmentations are represented as black dots, connected by a line when they provide from the same input $X$.
  }
  \label{fig:var}
  \vspace{-1em}
\end{figure}

\subsection{Experiment details for Figure~\ref{fig:downstream_error}}
\label{app:downstream_error_sphere}

We consider data uniformly distributed on the sphere~$\mathbb S^{d-1}$ with~$d = 8$, augmentations consisting of cyclic shifts of~$\{-1, 0, 1\}$, and a dot-product kernel of the form~$k(x, y) = (1 + x^\top y) \kappa(x^\top y)$, with~$\kappa(u) = 1 - \arccos(u)/\pi$.

The target functions~$f_\ell^*$ are given by:
\begin{align*}
    f_1^*(x) &= \frac{1}{3} \sum_{j=1}^3 Q_{1,d}(x_j) \\
    f_3^*(x) &= \frac{1}{d} \sum_{j=1}^d Q_{3,d}(x_j),
\end{align*}
where~$Q_{\ell,d}$ are the Gegenbauer polynomials introduced in Appendix~\ref{app:spherical_setup}.
Note that $f_3^*$ is a cyclic-invariant spherical harmonic of degree~$3$, while~$f_1^*$ is a non-invariant spherical harmonic of degree~$1$ (though is has some local shift stability).
Labels on the downstream tasks are generated from the~$f_\ell^*$ without noise.

Figure~\ref{fig:downstream_error} shows the downstream relative excess risk~$\|\hat f_n - f_\ell^*\|_{L^2}^2 / \|f_\ell^*\|_{L^2}^2$, approximated over 1500 test datapoints, as a function of the regularization parameter~$\lambda$ used in pretraining.
We use the same~$n=300$ samples for pretraining and downstream linear prediction.
Pretraining uses all 3 augmentations for each sample, with a representation dimension~$k=20$. The downstream problem is solved with kernel ridge regression using the induced kernel from pretraining, and the ridge parameter is tuned on test samples to avoid dealing with model selection issues.

\begin{figure}[ht]
  \centering
  \includegraphics{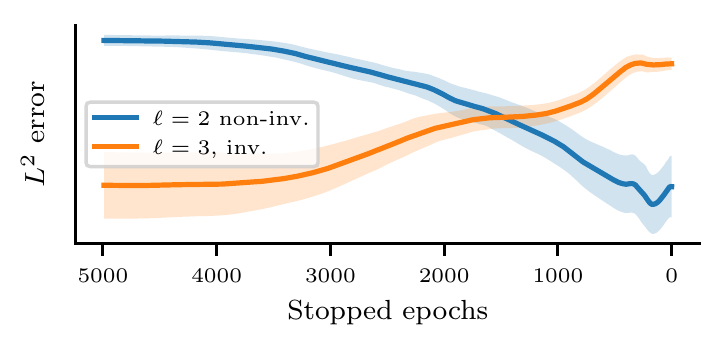}
  \vspace{-1em}
  \caption{\emph{Behavior of Figure \ref{fig:downstream_error} with a neural network}.
  The regularization parameter $\lambda$ is replaced by early stopping of SGD.
  We consider a neural network with two hidden layers, both made of 200 neurons.
  Optimization was performed with gradient descent with a constant step-size.
  Randomness due to weights initialization is averaged over 100 trials, the standard deviation being shown on the Figure.
  }
  \label{fig:nn_sphere}
  \vspace{-1em}
\end{figure}

\subsection{Experiment details for Figure~\ref{fig:error_cl}}

Figure \ref{fig:error_cl} considers a classification problem involving four classes with a pretraining task specifically constructed to design a representation $\psi:\X\to\R^k$ for $k=4$ that solves this particular classification problem.
The dataset we consider is the halfmoon dataset, where $X = Z + \ind{\scap{Z}{e_1} > 0} e_2 + U$, $Z\sim\uniform{\mathbb{S}^2}$, and $U\sim {\cal N}(0, \sigma^2 I)$ for $\sigma = 0.1$.
Augmentations apply Gaussian noise, $\xi = X + V$ for $V\sim {\cal N}(0, \sigma^2 I)$ with $\sigma=0.1$.
This setting corresponds to that with a Laplacian where ${\cal L}(\psi) \simeq \norm{\nabla \psi}^2_{L^2(\rho_\X)}$.
As a consequence, the ideal $\psi$ will correspond to the top eigenvalues of the Laplacian. I.e., the first two span the constant functions on both moons, the next two are waves with a single oscillation on a given moon, etc.
In essence, one can view the harmonics on $L^2([0,1])$ as $x\to \cos(2i \pi\omega x + \chi)$ for $\chi \in \brace{0, \pi / 2}$ and $\omega \in \N$, deforming the segment $[0, 1]$ to match one moon, and duplicating this basis on the other moons.
In this setting, eigenfunctions are not analytic, since analytic functions cannot be dissociated on two different manifold (e.g., a locally constant analytic function is globally constant).
As a consequence, searching for the eigenfunction with the radial basis function kernel ($\Psi$ only contains analytical function in this case \citep{Sun2008}) requires proper tuning of the regularization parameter as a function of the number of samples.
This explains our choice of the exponential kernel in this experiment, which corresponds to $\phi(x)^\top \phi(y) = \exp(-\norm{x-y} / \sigma)$ and is associated with the looser Sobolev space that is still a Reproducing kernel Hilbert space (in $\R^2$, this is $H^1$).
This improves the learning of the top eigenfunctions of $T$ without varying $\lambda$, better illustrating the convergence rates of Theorems \ref{thm:conv_simple}, \ref{thm:rad} and \ref{thm:sgd}.

\begin{figure}[ht]
  \centering
  \includegraphics{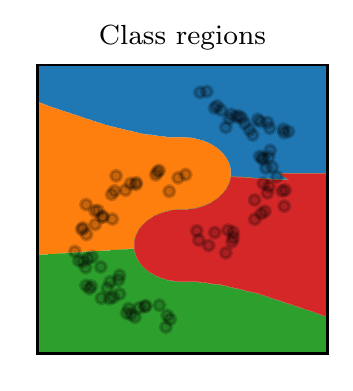}
  \includegraphics{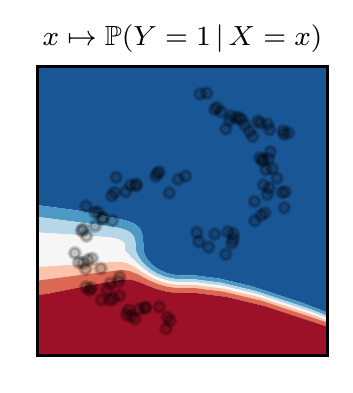}
  \includegraphics{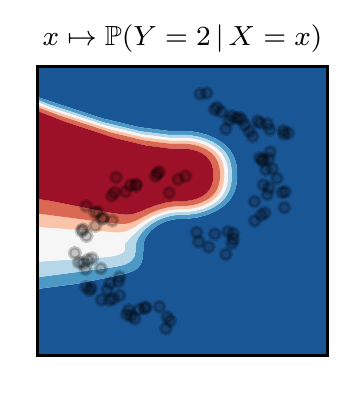}
  \includegraphics{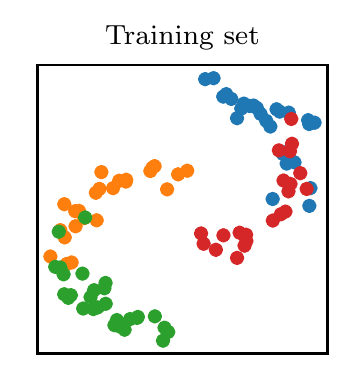}
  \caption{
  {\em Setting of Figure \ref{fig:error_cl}.}
  The downstream task consists in learning four classes in $\X = \R^2$ with are represented on the left.
  Those classes are generated with noise.
  The level lines of the conditional distribution of $Y$ given $X$ are represented on the middle for the left moons; the right moon follows the same structure.
  A training set example is on the right.
  }
  \label{fig:error_cl_setup}
  \vspace{-1em}
\end{figure}

\begin{figure}[ht]
  \centering
  \includegraphics{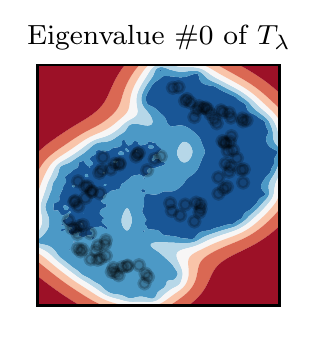}
  \includegraphics{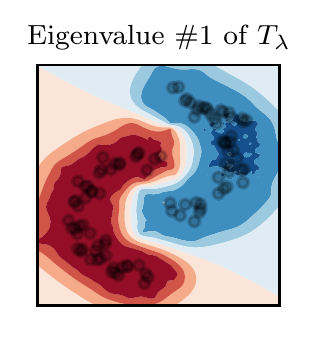}
  \includegraphics{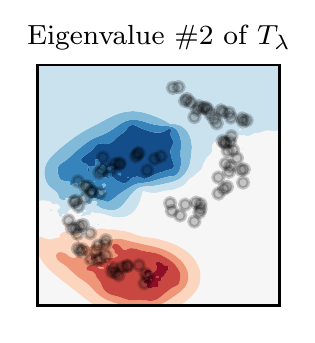}
  \includegraphics{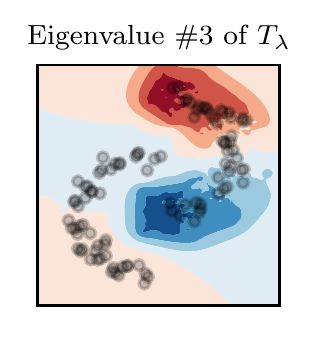}
  \includegraphics{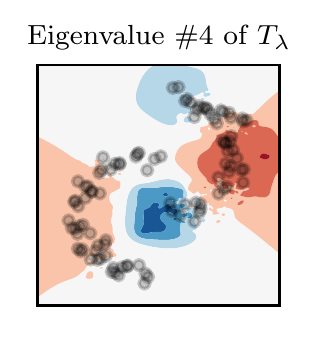}
  \caption{
  {\em Eigenvalues of $T_\lambda$} estimated empirically with 2000 pretraining samples on the problem that yield the empirical rates displayed on Figure \ref{fig:error_cl}.
  }
  \label{fig:cl_reprs}
  \vspace{-1em}
\end{figure}

\begin{figure}[ht]
  \centering
  \includegraphics{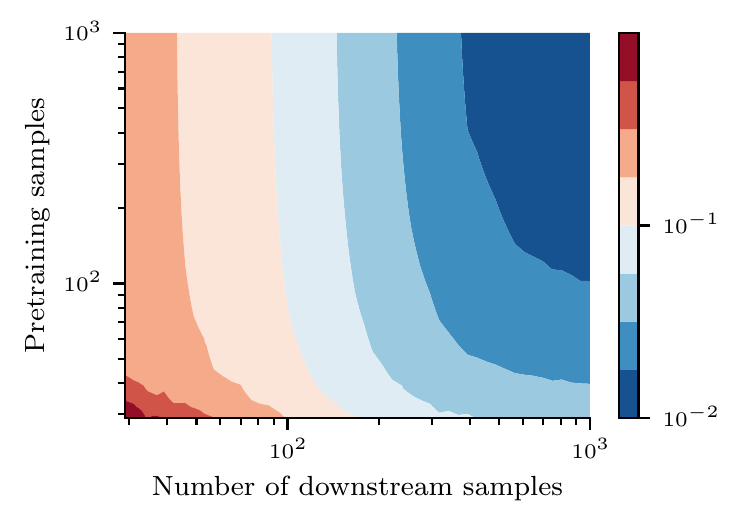}
  \includegraphics{figures/cl.pdf}
  \vspace{-1em}
  \caption{
  {\em Averaged downstream error} computed over 100 trials on the pretraining task and 200 trials on the downstream task, for both the least-squares loss (right) and the 0-1 loss (left).
}
  \label{fig:error_reg}
  \vspace{-1em}
\end{figure}

In our experiments, we fixed $\lambda = 10^{-3}$ and the scale of the exponential kernel $\sigma$ to be about one fifth of the problem diameter.
We plot the eigenfunctions of $T$ derived empirically with $n_{pre} = 2000$ samples in Figure \ref{fig:cl_reprs}.
The classification tasks aims to learn the four classes described on the left of Figure \ref{fig:error_cl_setup}.
Class labels include some noise as indicated by the level lines of the conditional probability of $Y$ as a function of $X$ shown in the middle of Figure \ref{fig:error_cl_setup}. A training set example is shown on the right of this figure with $n_{down} = 100$.
In the experiments we fix $k=5$, which ensures that there is strong correlation in performance between the pretraining and downstream tasks.
The downstream task is optimized with a least-squares surrogate: we learn $g:\X\to\R^4$ that minimizes the least-square error $\E[\norm{g(X) - e_Y}^2]$ before decoding it as $f(X) = \argmax_{i\in[4]} g_i(X)$ to get an estimate of the ideal mapping $f^*:\X\to\Y$.
We report the downstream generalization error on both the least-squares (surrogate) loss and the 0-1 loss on Figure \ref{fig:error_reg}.
This error is computed as the average over 100 trials on the pretraining task and 200 trials on the downstream task.

